\documentclass{article} 
\usepackage{iclr2027_conference,times}

\usepackage{amsmath,amsfonts,bm}

\def\eqref#1{eq.~\ref{#1}}

\def\1{\bm{1}}

\DeclareMathAlphabet{\mathsfit}{\encodingdefault}{\sfdefault}{m}{sl}
\SetMathAlphabet{\mathsfit}{bold}{\encodingdefault}{\sfdefault}{bx}{n}

\usepackage{hyperref}
\usepackage{url}
\usepackage{graphicx}
\usepackage{booktabs}
\usepackage{xcolor}
\usepackage{colortbl}
\usepackage{multirow}
\usepackage{tabularx}
\usepackage{adjustbox}
\usepackage{array}
\usepackage{amsfonts}
\usepackage{subcaption}
\usepackage{ragged2e}
\usepackage{afterpage}
\usepackage{tcolorbox}
\tcbuselibrary{breakable,skins}
\newtcolorbox{promptbox}[1]{
  enhanced,
  breakable,
  colback=gray!6,
  colframe=gray!45,
  boxrule=0.4pt,
  arc=2pt,
  left=3pt,
  right=3pt,
  top=2pt,
  bottom=2pt,
  fontupper={\RaggedRight\sloppy\ttfamily\scriptsize\setlength{\emergencystretch}{2.5em}},
  title={#1},
  coltitle=white,
  colbacktitle=gray!55,
  attach boxed title to top left={yshift=-1.5mm,xshift=2.5mm},
  boxed title style={rounded corners, sharp corners=northeast},
}
\graphicspath{{figures/}}
\newcommand{\ourmodel}{\textsc{PIVOT}}
\newcommand{\armGRPO}{\textsf{Vanilla-GRPO}}
\newcommand{\armSFTGRPO}{\textsf{SFT-GRPO}}
\newcommand{\armP}{\textsf{P}}
\newcommand{\armM}{\textsf{M}}
\newcommand{\armMP}{\textsf{M+P}}
\newcommand{\armMPR}{\textsf{M+P+R}}
\newcommand{\armMPRandR}{\textsf{M+P-Random+R}}
\newcommand{\acttxt}{a^{\mathrm{txt}}}
\newsavebox{\pivottabbox}

\title{PIVOT: Pivot-Aware On Policy Self Distillation for Multi-Turn VLM Agents}

\author{%
  \textbf{Jiazhou Zhou}$^{1,2}$\thanks{Work done during an internship at IDEA Research.}\hspace{1em}%
  \textbf{Hu Zhou}$^{3}$\hspace{0.35em}%
  \textbf{Yucheng Chen}$^{4}$\hspace{0.35em}%
  \textbf{Jinyuan Qu}$^{5,2}$\hspace{0.35em}%
  \textbf{Ying-Cong Chen}$^{1}$\hspace{0.35em}%
  \textbf{Lei Zhang}$^{2}$\thanks{Corresponding Author.} \\
  \\
  $^1$AI Thrust, The Hong Kong University of Science and Technology (Guangzhou) \\
  $^2$International Digital Economy Academy (IDEA) \\
  $^3$The Hong Kong Polytechnic University \\
  $^4$MedVisAI Lab, Lee Kong Chian School of Medicine, \\ Nanyang Technological University, and Centre of AI in Medicine \\
  $^5$Tsinghua University \\
  \\
}

\iclrfinalcopy
\makeatletter
\def\@maketitle{\vbox{\hsize\textwidth
{\LARGE\sc \@title\par}
\vskip 12.2pt
\centerline{\href{https://jiazhou-garland.github.io/PIVOT/}{Project Page}}%
\def\And{\end{tabular}\hfil\linebreak[0]\hfil
        \begin{tabular}[t]{l}\bf\rule{\z@}{16pt}\ignorespaces}%
\def\AND{\end{tabular}\hfil\linebreak[4]\hfil
        \begin{tabular}[t]{l}\bf\rule{\z@}{16pt}\ignorespaces}%
\begin{tabular}[t]{l}\bf\rule{\z@}{16pt}\@author\end{tabular}%
\vskip 0.3in minus 0.1in}}
\makeatother
\hypersetup{%
  pdftitle={PIVOT: Pivot-Aware On Policy Self Distillation for Multi-Turn VLM Agents},
  pdfauthor={Jiazhou Zhou, Hu Zhou, Yucheng Chen, Jinyuan Qu, Ying-Cong Chen, Lei Zhang},
}

\begin{document}

\maketitle
\lhead{Preprint}

\begin{abstract}
    Reinforcement learning with verifiable rewards (RLVR) via Group-Relative Policy Optimization (GRPO) is widely used for multi-turn VLM agent training, yet it suffers from zero-gradient silence on uniform failures and coarse episode-level credit assignment. 
    While On-Policy Distillation (OPD) and On-Policy Self-Distillation (OPSD) mitigates sparse rewards using hindsight information, its underlying mechanisms remain poorly understood. 
    Through controlled counterfactual rollback probes across five multi-turn VLM agent benchmarks, we reveal that performance gains in OPSD/OPD are largely driven by physical state rollback at the \emph{pivot step}, defined as the first unrecoverable action without remaining step budget. 
    However, physical state rollbacks are computationally prohibitive and infeasible in real-world environments. 
    To bridge this gap, we present \textbf{P}ivot-Aware \textbf{I}nternalized \textbf{V}isual \textbf{O}n-Policy \textbf{T}raining (\ourmodel{}), an RL framework that internalizes pivot localization and state restoration directly into token-level parameter updates, eliminating environment rollbacks during RL training and additional skill hints at test time.
    \ourmodel{} unifies three functional roles within a single architecture: a failure Analyzer non-invasively localizes the pivot step and diagnoses failure modes from visual trajectory collages and action logs; a detached Teacher re-scores failed tokens under this privileged diagnostic context; and a Student optimizes joint GRPO and confidence-gated OPD objectives. 
    At test time, both Teacher and Analyzer branches are stripped. 
    Evaluated on five multi-turn VLM agent tasks across cognitive grid puzzles, 3D embodied control and navigation, and generative reasoning, \ourmodel{} achieves $0.90$ overall accuracy on Qwen2.5-VL-3B ($+8\%$ over SFT+GRPO baseline and $+5\%$ over previous SOTA) and scales to $0.92$ on Qwen3-VL-2B ($+12\%$ over SFT+GRPO baseline).
\end{abstract}

\afterpage{%
\begin{figure*}[t]
\centering
\includegraphics[width=\linewidth]{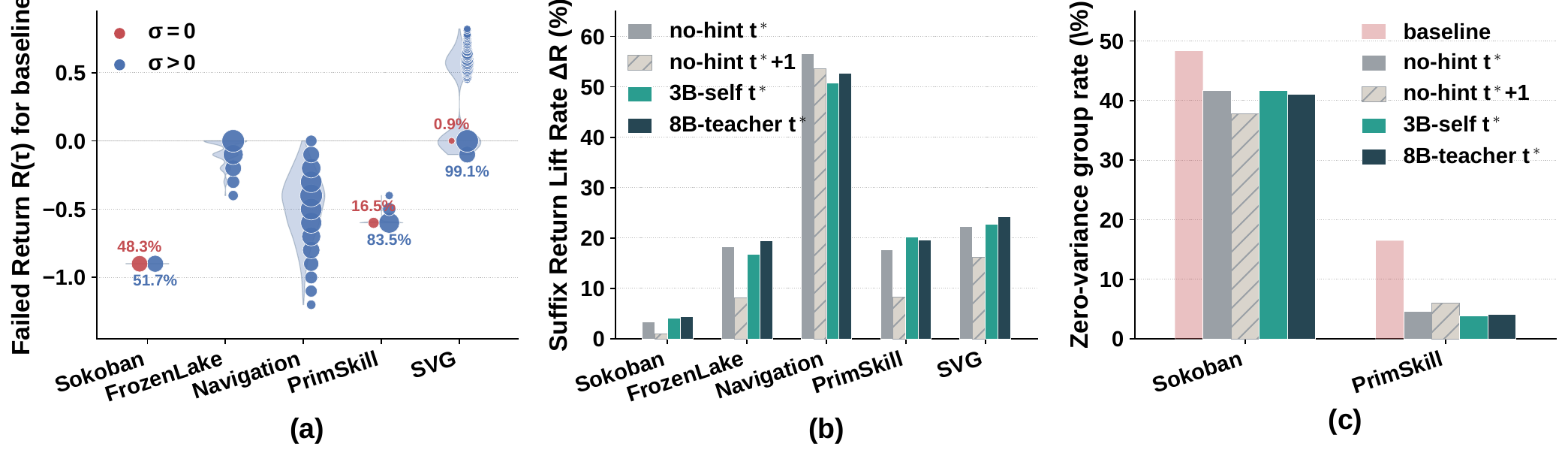}
\caption{%
\textbf{(a)}~Base policy failed-rollout returns $R(\tau)$, distinguishing zero-variance ($\sigma{=}0$, red) from variance-bearing ($\sigma{>}0$, blue) groups (marker size scales with trajectory frequency; percentages denote $\sigma{=}0$ share).
\textbf{(b)}~Suffix return lift $\Delta R$ (\%) across four rollback probes at the pivot step: no-hint resets at $t^\ast$ and $t^\ast{+}1$, alongside 3B-self and 8B-teacher skill hints at $t^\ast$.
\textbf{(c)}~Zero-variance group rate (\%) across four rollback probes on Sokoban and PrimitiveSkill.
}
\label{fig:sec2}
\end{figure*}%
}

\section{Introduction}
\label{sec:intro}
Reinforcement learning (RL) has become a dominant post-training paradigm for vision-language model (VLM) agents~\citep{ouyang2022instructgpt,guo2025deepseekr1,yao2023react,wang2025vagen,shen2025vlm,li2026glance}.
The prevailing RLVR framework, Group-Relative Policy Optimization (GRPO)~\citep{shao2024deepseekmath}, samples multiple rollouts per prompt and computes relative advantages by normalizing episode-level returns within each group.
However, as analyzed in Sec.~\ref{sec:grpo-blindspot} (Fig.~\ref{fig:sec2}(a)) across five multi-turn VLM agent benchmarks, GRPO suffers from two critical failure modes: \emph{zero-gradient silence}, where the policy gradient vanishes as the group returns are identical~\citep{yu2025dapo}, and \emph{coarse episode-level credit assignment}, which ranks entire trajectories by scalar returns without pinpointing which step first renders the task unrecoverable~\citep{lightman2024prm,uesato2022process}.

On-Policy Distillation (OPD)~\citep{agarwal2024gkd,lu2026sdar} mitigates sparse rewards via token-level supervision, but fundamentally requires access to white-box teacher logits and matched vocabularies. 
To bypass these constraints, On-Policy Self-Distillation (OPSD)~\citep{wang2026skillsd,zhao2026opsd,andrychowicz2017hindsight,shinn2023reflexion} performs self-supervision by conditioning policy updates on hindsight information. 
However, the mechanisms driving OPSD/OPD gains in multi-turn VLM agents remain poorly understood. 
To figure out these underlying mechanisms, we conduct counterfactual probes by systematically rolling back the environment to the \emph{pivot step}, defined as the first unrecoverable action within the remaining step budget, and then resampling trajectory suffixes to evaluate the performance gains.

Empirically, as shown in Sec.~\ref{sec:cf-recovery} and Figs.~\ref{fig:sec2}(b)--(c), our probe reveals three key findings:
\textbf{(i)} Restoring the environment state to the pivot step without any skill hints provides the primary share of suffix return lift and substantially reduces zero-advantage GRPO groups;
\textbf{(ii)} Delaying this state reset by even a single step ($t^\ast{+}1$) causes suffix return gains to sharply diminish; and 
\textbf{(iii)} Incorporating explicit skill hints, whether self-generated or teacher-provided, yields only marginal incremental gains over hint-free state restoration (Fig.~\ref{fig:sec2}(b)).  These observations yield a vital insight:  \textit{\textbf{the efficacy of OPD/OPSD in multi-turn VLM agents is largely driven by precise state rollbacks at the pivot step, rather than by additional skill prompting}}.

However, state rollback is computationally prohibitive and infeasible in real-world, non-rewindable environments, while inference cannot tolerate additional skill prompt overhead. Therefore, a practical RL framework for multi-turn VLM agents must: 
\textbf{(i)}~non-invasively and precisely localize the pivot step directly from trajectory histories, and 
\textbf{(ii)}~convert this localized pivot step into fine-grained, token-level policy gradients wihout environment rollbacks during RL training and additional skill hints at test time.

To this end, we present \textbf{P}ivot-Aware \textbf{I}nternalized \textbf{V}isual \textbf{O}n-Policy \textbf{T}raining (\ourmodel{}), an on-policy distillation framework that internalizes pivot step localization and failure diagnostic hints directly into parameter updates without simulator rollbacks (Fig.~\ref{fig:overview}, Sec.~\ref{sec:method}).
In Stage~I, a lightweight SFT phase cold-starts a \textbf{failure Analyzer} to non-invasively predict the pivot step, failure mode, and optional guidance skills directly from visual trajectory collages and action logs (Sec.~\ref{sec:sft}).
In Stage~II, a \textbf{three-role unified architecture} orchestrates training by consolidating Student, Analyzer, and Teacher roles within shared parameters: the Analyzer extracts a localized visual panel around the pivot step, combining it with the predicted failure mode and guidance skills to construct a privileged context.
A detached Teacher then re-scores original failed tokens under this privileged context, while the Student optimizes joint GRPO and Gated OPD objectives (Sec.~\ref{sec:loss}).
At test time, both Analyzer and Teacher are stripped away, deploying the agent strictly as an unprivileged Student with zero computational and skill prompt overhead. Our main contributions are summarized as follows:
\begin{itemize}
    \item \textbf{Mechanistic Insight into OPSD:} We characterize GRPO failure modes in multi-turn VLM agents and empirically analyze via counterfactual rollback probes that the OPSD/OPD performance gains stem mainly from state rollback at the critical pivot step, rather than skill hints as additional prompts.
    \item \textbf{Non-Invasive Pivot Step Localization:} We propose \ourmodel{}, introducing a trajectory Analyzer that non-invasively localizes pivot steps and provide failure diagnoses directly from complete visual collages and action logs.
    \item \textbf{Unified Three-Role Architecture:} We formulate a single-model paradigm consolidating Student, Analyzer, and Teacher roles, internalizing state rollback during training while preserving zero computational and skill prompt overhead at inference.
    \item \textbf{Superior Performance:} Evaluated on five multi-turn VLM agent benchmarks across cognitive grid puzzles, 3D embodied control and navigation, and generative reasoning, \ourmodel{} achieves $0.90$ overall accuracy on Qwen2.5-VL-3B~\citep{bai2025qwen25vl} ($+8\%$ over the SFT+GRPO baseline and $+5\%$ over dense-reward SOTA) and scales to $0.92$ on Qwen3-VL-2B~\citep{bai2025qwen3vl} ($+12\%$ over the SFT+GRPO baseline).
\end{itemize}

\section{Mechanistic Analysis: Physical Restoration Dominates Hindsight Guidance}
\label{sec:pivot-analysis}

\subsection{Limitations of Episode-Level Credit Assignment}
\label{sec:grpo-blindspot}

\paragraph{Preliminaries.}
We formulate a multi-turn vision-language agent task as a partially observable Markov decision process (POMDP)~\citep{aastrom1965optimal} defined by the tuple $\mathcal{M}=(\mathcal{S},\mathcal{O},\mathcal{A},\mathcal{P},\mathcal{R},\Omega,\gamma)$, where $\mathcal{S}$ is the state space, $\mathcal{O}$ the observation space, $\mathcal{A}$ the action space, $\mathcal{P}(s_{t+1}\mid s_t,a_t)$ the transition dynamics, $\mathcal{R}(s_t,a_t)$ the reward function, $\Omega(o_t\mid s_t)$ the observation emission model, and $\gamma \in (0, 1]$ the discount factor.
At turn $t$, the agent receives observation $o_t \sim \Omega(\cdot\mid s_t)$, maintains execution history $h_t=(o_0,a_0,\ldots,o_t)$, and samples action $a_t\sim\pi_\theta(\cdot\mid h_t)$.
Execution yields a rollout trajectory $\tau=(o_0,a_0,r_0,\ldots,o_{T-1},a_{T-1},r_{T-1})$ of length $T$ with cumulative return $R(\tau)=\sum_{t=0}^{T-1} \gamma^t r_t$.
Given prompt $q\sim\mathcal{Q}$, post-training RL optimizes expected trajectory return $J(\theta)=\mathbb{E}_{q,\tau}[R(\tau)]$.

The prevailing estimator for RLVR, Group-Relative Policy Optimization (GRPO)~\citep{shao2024deepseekmath}, samples a group of $N$ rollouts $\{\tau^{(n)}\}_{n=1}^{N}$ per prompt and computes standardized group advantages:
\begin{equation}
A_{n}^{\mathrm{rl}} = \frac{R(\tau^{(n)})-\mu}{\sigma+\epsilon}, \quad 
\mu=\frac{1}{N}\sum_{n=1}^{N}R(\tau^{(n)}), \quad 
\sigma=\sqrt{\frac{1}{N}\sum_{n=1}^{N}\bigl(R(\tau^{(n)})-\mu\bigr)^{2}},
\label{eq:grpo-adv}
\end{equation}
where $\epsilon > 0$ is a numerical stabilizer.
GRPO assigns this scalar $A_{n}^{\mathrm{rl}}$ uniformly to \emph{every} action token in $\tau^{(n)}$. Consequently, it evaluates trajectories solely by episode-level return, ignoring which specific turn rendered the task unrecoverable.

\paragraph{Failed Rollouts vs. Zero-Advantage Groups.}
Let $R_{\mathrm{succ}}$ denote the domain-specific success return threshold. We define the set of failed rollouts within a sampled group as $N^{-} = \{\tau^{(n)} \mid R(\tau^{(n)}) < R_{\mathrm{succ}}\}$.
While failed rollouts can still receive non-zero advantages when group returns vary, a \emph{zero-advantage group} arises when return variance vanishes ($\sigma{=}0$ in Eq.~\eqref{eq:grpo-adv}).
In such cases, GRPO suffers from \emph{zero-gradient silence} ($A_n^{\mathrm{rl}}=0$), rendering policy updates inactive even when execution trajectories exhibit diverse failure behaviors.

\paragraph{Empirical Observation.}
We evaluate untrained \texttt{Qwen2.5-VL-3B-Instruct} across five VAGEN environments~\citep{wang2025vagen} (Sec.~\ref{sec:setup}, $N{=}8$).
As shown in Fig.~\ref{fig:sec2}(a), low baseline success on Sokoban ($17.8\%$) and PrimitiveSkill ($8.9\%$) yields frequent zero-variance groups ($\sigma{=}0$ on $48.3\%$ and $16.5\%$), where identical returns cause all GRPO advantages $A_n^{\mathrm{rl}}$ to vanish despite diverse errors.
In FrozenLake and Navigation ($0.0\%$ $\sigma{=}0$), the $-0.1$ per-step penalty differentiates returns across varying trajectory lengths,forming the $-0.1$ ladder in Figure~\ref{fig:sec2}(a), but $\sigma{>}0$ merely penalizes episode length without isolating the critical failure turn.
Similarly, SVG visual scores yield low $\sigma{=}0$ ($0.9\%$), yet credit assignment remains episode-level rather than step-level.

\paragraph{Takeaway.}
Failed GRPO trajectories suffer from \emph{zero-gradient silence} or \emph{coarse credit}.

\subsection{Disentangling OPD/OPSD: Physical State Rollback Dominates Skill Hints}
\label{sec:cf-recovery}
Recently, OPD has gained traction by supplying token-level targets via negative KL divergence with a teacher~\citep{agarwal2024gkd,lu2026sdar}. On-Policy Self-Distillation (OPSD) relaxes white-box constraints by conditioning policy updates on hindsight information~\citep{wang2026skillsd,shinn2023reflexion,zhao2026opsd}. 
However, the underlying mechanism driving OPSD in multi-turn VLM agents remains poorly understood. 
By systematically restoring environment states at critical steps and resampling trajectory suffixes with or without textual guidance, we isolate the true driver of post-failure performance gains.

\paragraph{Pivot Step \(t^\ast\).}
Given a domain feasibility certificate $\mathsf{Feas}(s, k) \in \{0, 1\}$ indicating whether state $s$ can reach the goal within $k$ steps (App.~\ref{app:certs}), we define the \emph{pivot step} $t^\ast$ as the first action that renders the task unfeasible under the remaining budget:
\begin{equation}
t^\ast = \min\bigl( \{t \in [0,T-1] : \mathsf{Feas}(s_{t+1}, T - t - 1) = 0\} \cup \{T-1\} \bigr).
\label{eq:pivot}
\end{equation}

\paragraph{Suffix Return Lift Rate.}
Let $\tau_{t^\ast:}$ denote the suffix of $\tau$ from pivot step $t^\ast$ onward; its discounted return is $R(\tau_{t^\ast:})=\sum_{t=t^\ast}^{T-1} \gamma^{t-t^\ast} r_t$.
After rolling back the environment state to $s_{t^\ast}$ and resampling a new suffix $\tilde\tau_{t^\ast:}\sim\pi_{\mathrm{base}}$ on faliure trajectories $N^{-}$, we record a successful recovery if $R(\tilde\tau_{t^\ast:}) > R(\tau_{t^\ast:})$.
We quantify recovery via the \emph{suffix return lift rate} $\Delta R$:
\begin{equation}
\Delta R = \bigl|\{\tau \in N^{-} : R(\tilde\tau_{t^\ast:}) > R(\tau_{t^\ast:})\}\bigr|\,/\,|N^{-}|.
\label{eq:lift-rate}
\end{equation}

\paragraph{Counterfactual Rollback Setup.}
On failed rollouts, we roll back the environment state to $s_{t^\ast}$ and resample a single suffix under four controlled conditions:
(1) \textbf{no-hint}: resamples purely from the rolled-back state $s_{t^\ast}$ without skill hints;
(2) \textbf{no-hint ($t^\ast{+}1$)}: rolls back state to $s_{t^\ast{+}1}$ to evaluate temporal localization sensitivity;
(3) \textbf{3B-self}: appends failure mode $m^\ast$ and self-generated skill hints from $\pi_{\mathrm{base}}$ (mirroring OPSD);
(4) \textbf{8B-teacher}: appends skill hints generated by a larger teacher model (\texttt{Qwen3-VL-8B}, mirroring OPD).

\paragraph{Empirical Findings.}
Fig.~\ref{fig:sec2}(b) and~(c) summarize the $n{=}1$ probe (numerical details in App.~\ref{app:cf-recovery}):
\textbf{(i)} Rollback $s_{t^\ast}$ with \textbf{no hints} boosts suffix returns $\Delta R$ across all tasks ($3.3\%$--$56.6\%$) while cutting zero-variance group rates ($48.3\%\rightarrow 41.7\%$ on Sokoban, $16.5\%\rightarrow 3.8\%$ on PrimitiveSkill; Fig.~\ref{fig:sec2}(c)),confirming that state rollback delivers the primary performance gains and resolves zero-gradient silence.
\textbf{(ii)} Resetting just one step later ($t^\ast{+}1$) sharply drops $\Delta R$ on four tasks (e.g., Sokoban $3.3\%\rightarrow 0.9\%$, PrimitiveSkill $17.7\%\rightarrow 8.3\%$; Fig.~\ref{fig:sec2}(b)), demonstrating that recovery strictly hinges on temporal precision.
\textbf{(iii)} Skill hints yield marginal additional gain (3B-self differs from no-hint by $\le 2.6$ points; 8B-teacher advances 3B-self within $0.3$--$3.5$ points).

\paragraph{Takeaway.}
\textbf{(i) State Rollback Drives Recovery:} State rollback at $t^\ast$ provides the predominant performance gain, serving as the core mechanism underlying OPSD/OPD paradigms. 
\textbf{(ii) Skill Hints Offer Marginal Value:} Skill hints, whether self-generated (3B-self) or teacher-provided (8B-teacher), yield negligible gains beyond pure state rollback. 
\textbf{(iii) Temporal Precision is Critical:} Delaying state rollback by even a single step ($t^\ast{+}1$) severely degrades recovery capacity.

\subsection{From Physical Rollbacks to Parameter-Internalized Updates}
\label{sec:motivation}

These mechanistic findings expose the \emph{Rollback Paradox} in post-training VLM agents: while trajectory recovery stems overwhelmingly from state rollback at the pivot turn ($t^\ast$), explicit physical rollbacks are computationally prohibitive during online RL and impossible in real-world, non-rewindable environments. 

To bridge this gap and convert physical state rollback directly into parameter updates without simulator resets, a practical post-training framework must fulfill three core requirements:
\textbf{(i)~Non-Invasive Pivot Localization:} identifying the critical pivot step ($t^\ast$) directly from failure trajectory histories, bypassing environment resets;
\textbf{(ii)~Internalized Credit Assignment:} converting the localized pivot turn into fine-grained, token-level policy gradients directly on original failed rollouts without physical rollbacks; and
\textbf{(iii)~Zero Test-Time Overhead:} internalizing hindsight diagnostic capabilities into policy parameters to enable unprivileged deployment without additional skill prompts or auxiliary models at test time.
 
\section{PIVOT: Pivot-Aware Internalized Visual On-Policy Training}
\label{sec:method}

\begin{figure}[t]
\centering
    \includegraphics[width=0.98\linewidth]{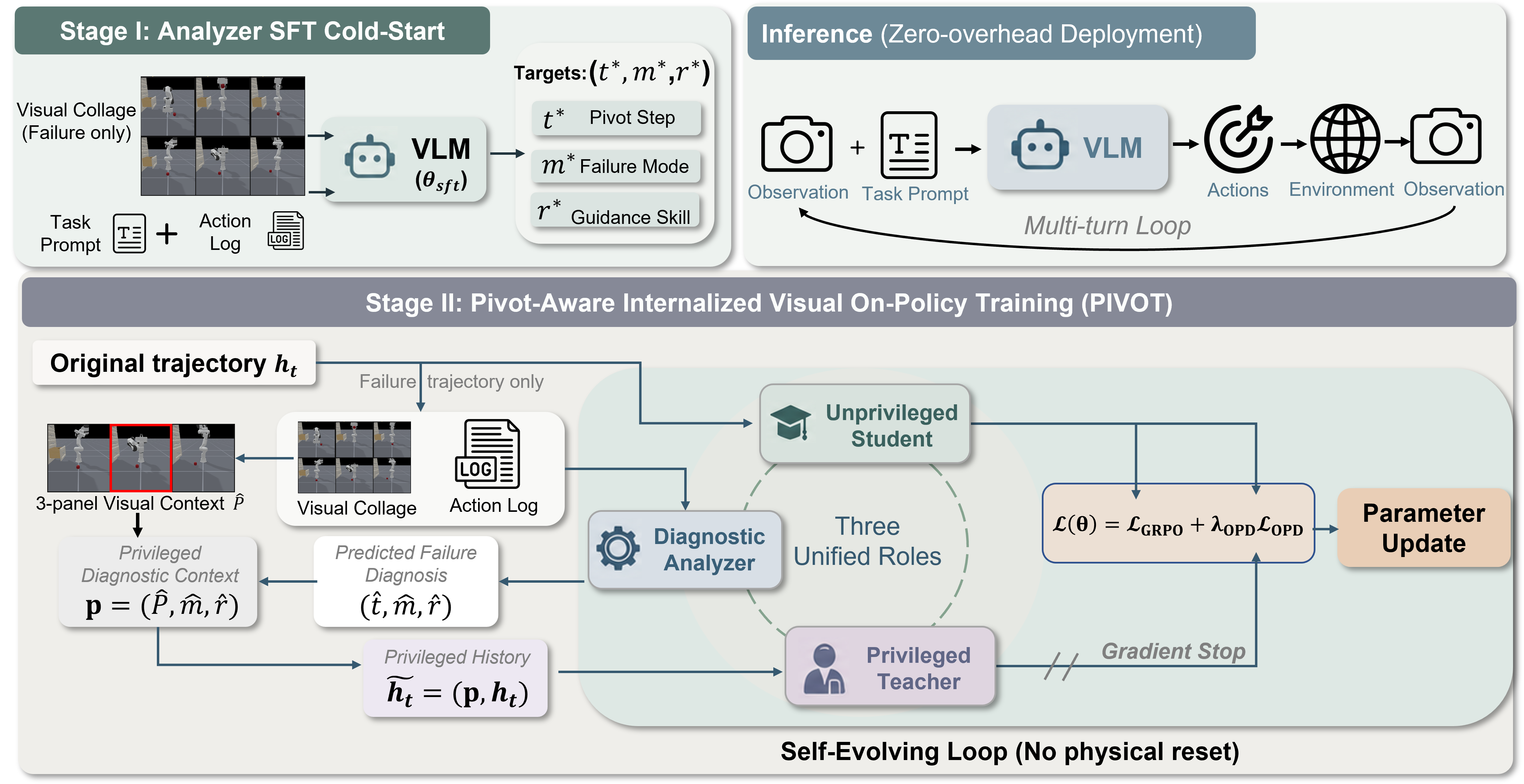}
    \caption{\textbf{\ourmodel{} Architecture Overview.}
    \ourmodel{} unifies Student, Analyzer, and Teacher roles within a single shared policy $\pi_\theta$.
    \textbf{Stage I:} SFT cold-starts the Analyzer to predict diagnostic triplets $(t^\ast, m^\ast, r^\ast)$ from visual trajectory collages $C(\tau)$ and action logs.
    \textbf{Stage II:} For failed rollouts, the Analyzer localizes the pivot step $\hat{t}$ to construct a privileged context $p$, enabling the detached Teacher to re-score failed tokens under $\tilde{h}_t = (p, h_t)$ for joint GRPO and confidence-gated OPD training.
    \textbf{Inference:} Analyzer and Teacher branches are stripped, deploying $\pi_\theta$ strictly as an unprivileged Student with zero computational and prompt overhead.}
\label{fig:overview}
\end{figure}

In this section, we present \ourmodel{}, a post-training framework that consolidates Student, Analyzer, and Teacher roles within a single parameterized VLM policy $\pi_\theta$.
Specifically, $\pi_\theta$ non-invasively localizes the pivot step $t^\ast$ and diagnoses failure via a visual trajectory Analyzer (Sec.~\ref{sec:sft}), and acts as its own Privileged Teacher to distill fine-grained visual hindsight into token-level policy updates on original failed rollouts (Sec.~\ref{sec:loss}).
By internalizing visual state restoration directly into model parameters, \ourmodel{} eliminates environment rollbacks during training while maintaining zero computational and prompt overhead at test time.

\subsection{Stage I: Visual Diagnostic Cold-Start for Pivot Step Localization}
\label{sec:sft}

To bypass state rollbacks during online RL training, the agent first learns to non-invasively localize the recoverable pivot step $t^\ast$ directly from VLM agent's trajectory history. As shown in Fig.~\ref{fig:overview} (Stage~I), we initialize this diagnostic capability via the Supervised Fine-Tuning (SFT) phase.

\paragraph{Diagnostic Target Triplets $(t^\ast,m^\ast,r^\ast)$.}
Each failed trajectory $\tau \in N^{-}$ is annotated with a target diagnostic triplet $z_\tau = (t^\ast,m^\ast,r^\ast)$, where $t^\ast$ is the ground-truth pivot step from Eq.~\eqref{eq:pivot}, $m^\ast \in \mathcal{M}_{\mathrm{fail}}$ denotes a discrete failure mode category (e.g., \texttt{deadlock}, \texttt{timeout}), and $r^\ast$ represents an optional guidance skill prompt generated offline via a teacher (App.~\ref{app:hyper}).

\begin{figure*}[t]
\centering
\begin{subfigure}[t]{0.96\textwidth}
\centering
\includegraphics[width=\linewidth]{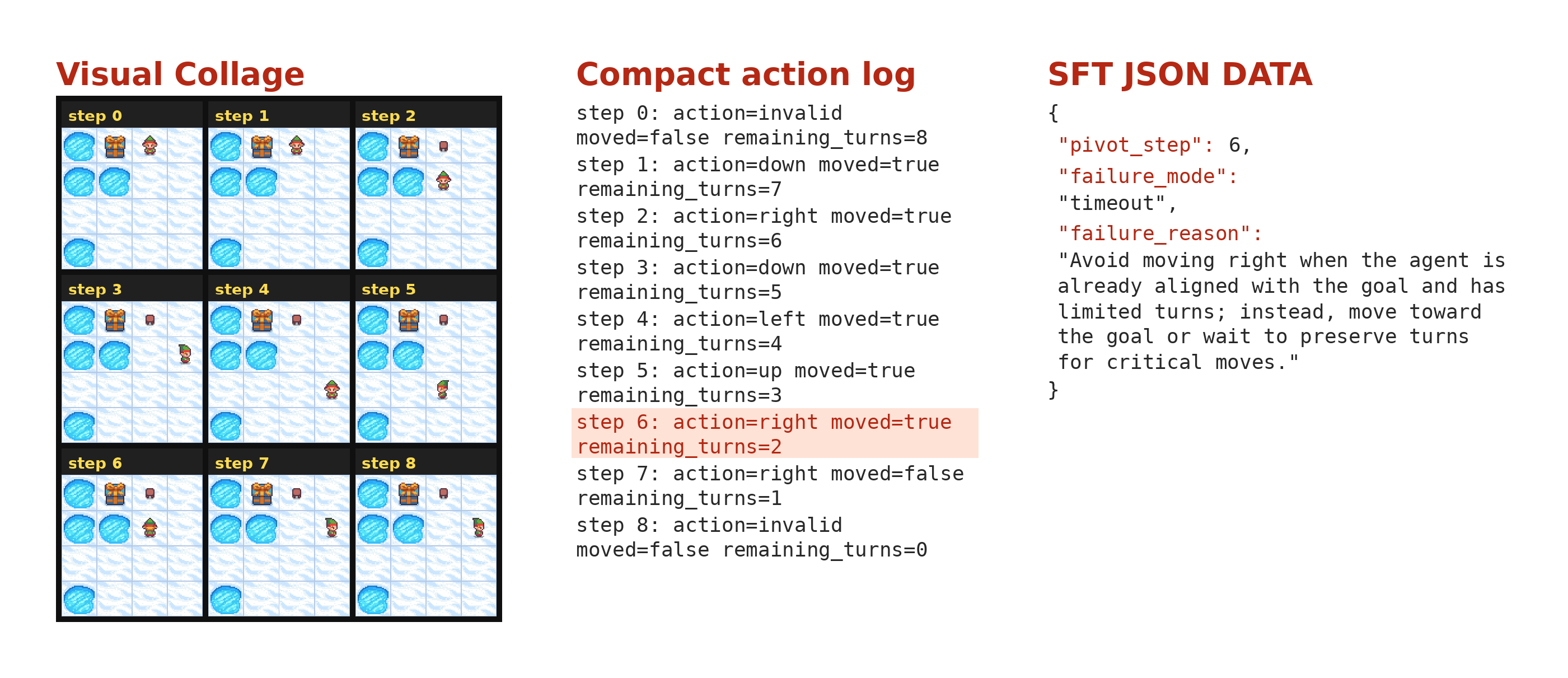}
\caption{FrozenLake ($t^\ast{=}6$, \texttt{timeout}).}
\end{subfigure}\\[-2pt]
\begin{subfigure}[t]{0.96\textwidth}
\centering
\includegraphics[width=\linewidth]{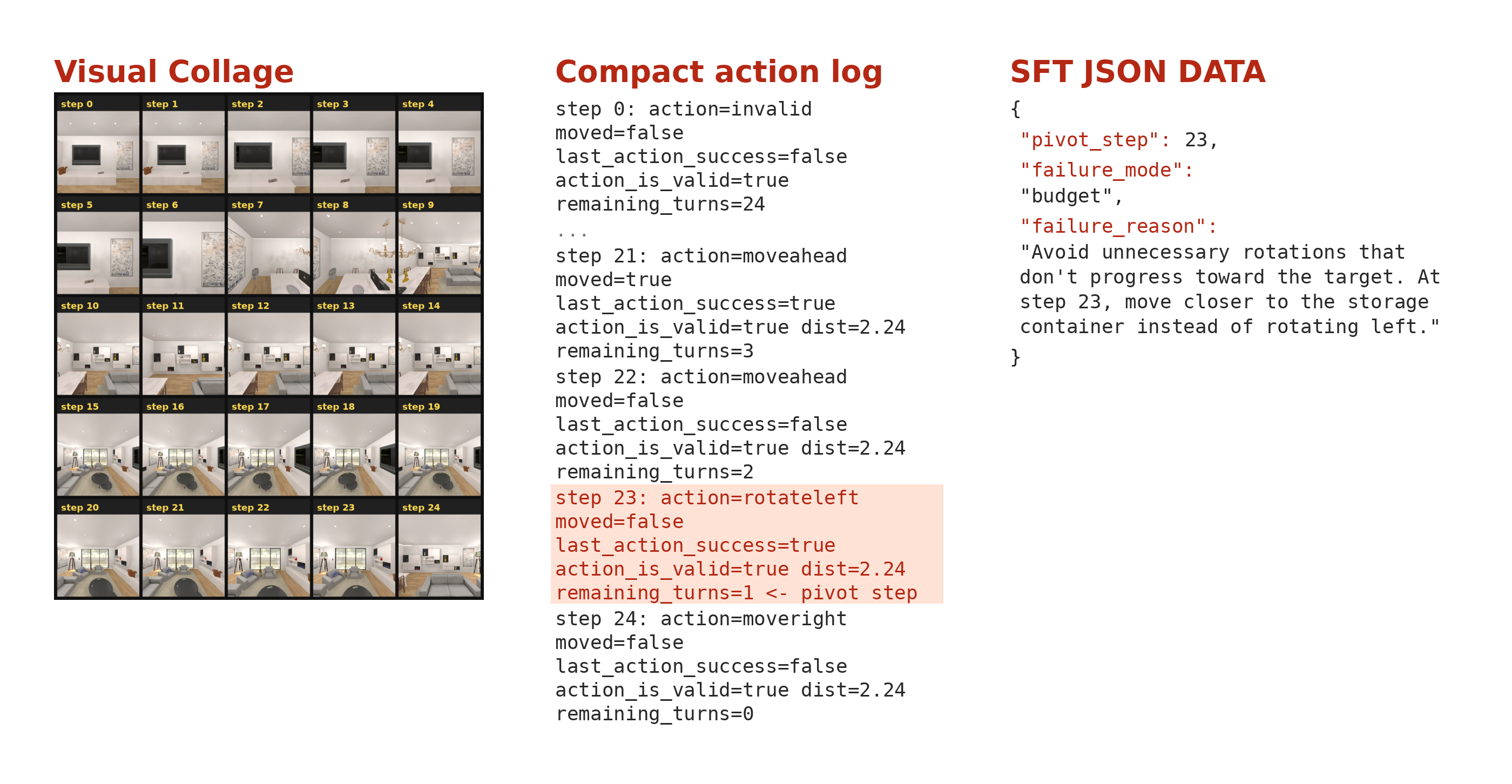}
\caption{Navigation ($t^\ast{=}23$, \texttt{budget}).}
\end{subfigure}\\[-2pt]
\begin{subfigure}[t]{0.96\textwidth}
\centering
\includegraphics[width=\linewidth]{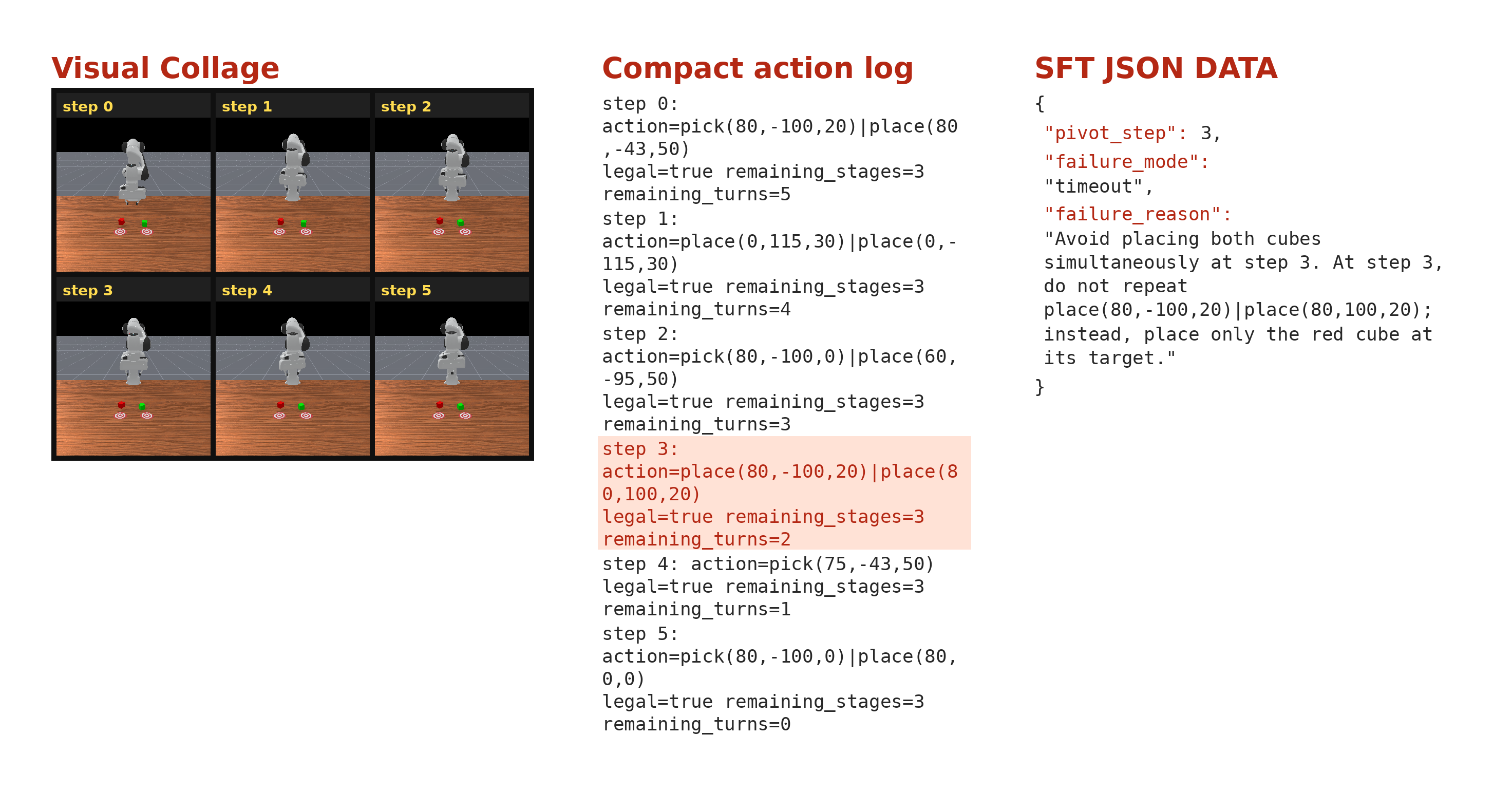}
\caption{PrimitiveSkill ($t^\ast{=}3$, \texttt{timeout}).}
\end{subfigure}
\caption{\textbf{Analyzer SFT input--output examples.} Each row pairs a visual trajectory collage and compact action log with the JSON supervision target. The highlighted log line is the pivot step. Additional environments and samples are in App.~\ref{app:examples}.}
\label{fig:sft-examples}
\end{figure*}

\paragraph{Visual Collage Construction \& Input.}
The Analyzer receives a multi-frame visual collage $C(\tau) = \mathrm{Grid}(o_0,\ldots,o_{T-1})$ augmented with explicit step indices, alongside the full action sequence $a_{0:T-1}=(a_0,\ldots,a_{T-1})$ and task prompt $q$. Crucially, this input relies strictly on observable trajectory histories without requiring access to environment states or internal memories.
Fig.~\ref{fig:sft-examples} shows one accepted training pair on FrozenLake, Navigation, and PrimitiveSkill.

\clearpage

\paragraph{SFT Training.}
We optimize $\pi_\theta$ on the diagnostic dataset $\mathcal{D}_{\mathrm{sft}} = \{ (q, C(\tau), a_{0:T-1}) \to (t^\ast,m^\ast,r^\ast) \}$ using standard autoregressive cross-entropy loss (App.~\ref{app:sft-loss}).
The resulting checkpoint $\theta_{\mathrm{sft}}$ equips $\pi_\theta$ with visual diagnostic capabilities, providing a robust cold-start for Stage~II.

\subsection{Stage II: Internalizing Localized Pivot Step Hindsight into Policy Updates}
\label{sec:loss}

In Stage~II, \ourmodel{} translates localized pivot step hindsight directly into fine-grained, token-level policy gradients on original failed rollouts without physical environment resets (Fig.~\ref{fig:overview}, Stage~II). Each online RL iteration synchronizes on-policy rollout sampling with privileged target scoring, optimizing policy parameters under a unified joint objective.

\begin{figure*}[t]
\centering
\begin{subfigure}[t]{0.32\textwidth}
\centering
\includegraphics[width=\linewidth]{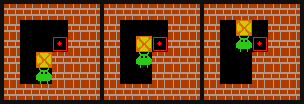}
\caption{Sokoban ($t^\ast{=}4$, \texttt{deadlock}).}
\end{subfigure}\hfill
\begin{subfigure}[t]{0.32\textwidth}
\centering
\includegraphics[width=\linewidth]{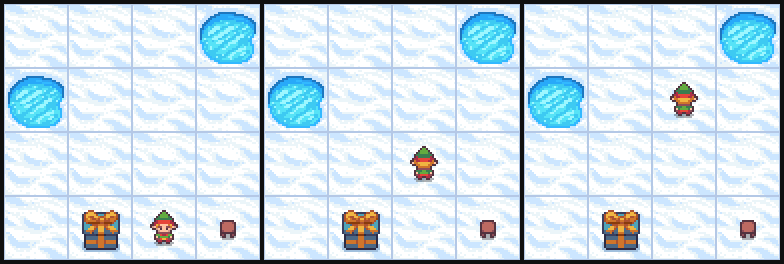}
\caption{FrozenLake ($t^\ast{=}6$, \texttt{timeout}).}
\end{subfigure}\hfill
\begin{subfigure}[t]{0.32\textwidth}
\centering
\includegraphics[width=\linewidth]{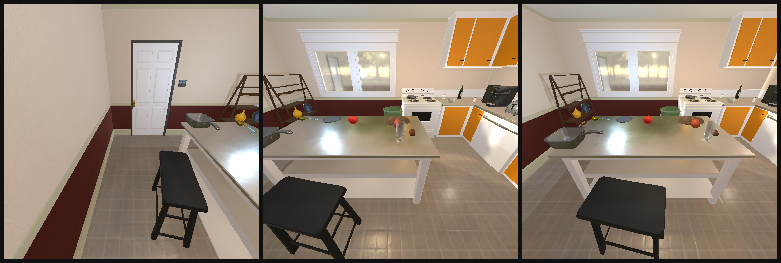}
\caption{Navigation ($t^\ast{=}21$, \texttt{budget}).}
\end{subfigure}\\
\begin{subfigure}[t]{0.32\textwidth}
\centering
\includegraphics[width=\linewidth]{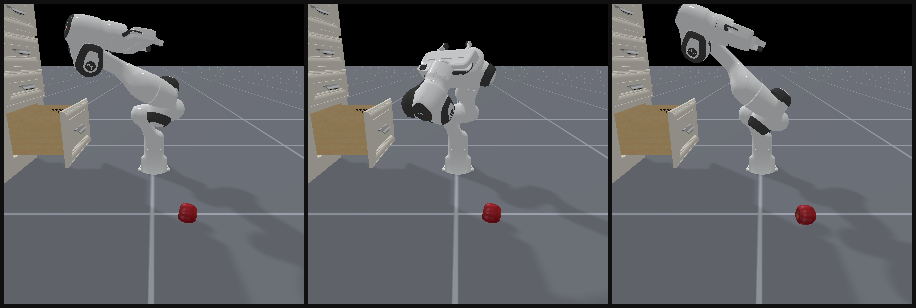}
\caption{PrimSkill ($t^\ast{=}4$, \texttt{timeout}).}
\end{subfigure}\hspace{0.02\textwidth}
\begin{subfigure}[t]{0.32\textwidth}
\centering
\includegraphics[width=\linewidth]{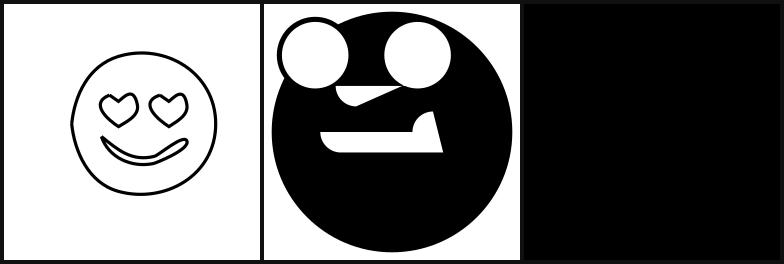}
\caption{SVG ($t^\ast{=}1$, \texttt{timeout}).}
\end{subfigure}
\caption{\textbf{Hindsight Visual Panels $P_{t^\ast}=[o_{t^\ast-1},o_{t^\ast},o_{t^\ast+1}]$.} Localized visual neighborhood extracted around the pivot step $t^\ast$ across five benchmark tasks (boundary frames zero-padded).}
\label{fig:panels}
\end{figure*}

\paragraph{Role 1: Diagnostic Analyzer (Pivot Localization).}
Given a sampled rollout group $\mathcal{G}=\{\tau^{(1)},\ldots,\tau^{(N)}\}$, the Diagnostic Analyzer processes each failed rollout $\tau \in N^{-}$ to predict:
\begin{equation}
(\hat t,\hat m,\hat r) = \pi_\theta\bigl(q, C(\tau), a_{0:T-1}\bigr),
\label{eq:analyzer}
\end{equation}
where $\hat t$ denotes the predicted pivot step, $\hat m \in \mathcal{M}_{\mathrm{fail}}$ the predicted failure category, and $\hat r$ the predicted guidance skill.
Using the predicted pivot step $\hat t$, the Analyzer extracts a 3-panel visual neighborhood crop $P_{\hat t} = [o_{\hat t-1}, o_{\hat t}, o_{\hat t+1}]$ centered around $\hat t$ (with boundary frames zero-padded), illustating in Fig.~\ref{fig:panels}.
The privileged diagnostic context is then constructed as $p = (P_{\hat t}, \hat m, \hat r)$.

\paragraph{Role 2: Privileged Teacher (Hindsight Target Scoring).}
The Privileged Teacher re-evaluates the Student's original action tokens under the privileged history $\tilde h_t$, formed by prepending diagnostic context $p$ to the standard interaction history $h_t=(o_0,a_0,\ldots,o_t)$:
\begin{equation}
\tilde h_t = (p, h_t),\qquad
\ell_t^{\mathrm{tea}} = \operatorname{sg}\bigl[\log \pi_\theta(a_t \mid \tilde h_t)\bigr].
\label{eq:teacher}
\end{equation}
Without executing extra simulation steps or resetting environment states, the Teacher acts as a detached target evaluator: the stop-gradient operator $\operatorname{sg}[\cdot]$ prevents Teacher-side parameter updates.
Conditioning on $\tilde h_t$ enables the Teacher to provide precise token-level credit by implicitly answering how original actions should be re-weighted under privileged context $p$.

\paragraph{Role 3: Acting Student (Token-Level Policy Update).}
Concurrently, the Unprivileged Student branch evaluates action tokens under standard interaction histories $h_t$, producing unprivileged log-probabilities $\ell_t^{\mathrm{stu}} = \log \pi_\theta(a_t \mid h_t)$.
GRPO optimizes policy updates based on group-normalized advantages $A^{\mathrm{rl}}$:
\begin{equation}
\mathcal{L}_{\mathrm{GRPO}}(\theta) = -\mathbb{E}\Bigl[ m_{\mathrm{act}} \min\bigl( \rho_t A^{\mathrm{rl}},\, \mathrm{clip}(\rho_t, 1-\varepsilon, 1+\varepsilon) A^{\mathrm{rl}} \bigr) \Bigr],
\label{eq:grpo-loss}
\end{equation}
where $\rho_t = \exp(\ell_t^{\mathrm{stu}} - \ell_t^{\mathrm{old}})$, $\ell_t^{\mathrm{old}} = \log \pi_{\theta_{\mathrm{old}}}(a_t \mid h_t)$, $\varepsilon > 0$ is the clipping threshold, and $m_{\mathrm{act}}$ masks valid action tokens.
For failed rollouts ($\mathbb{I}[R(\tau) < R_{\mathrm{succ}}]$), confidence-gated OPD~\citep{agarwal2024gkd,lu2026sdar} (More details in App.~\ref{app:opd}) provides dense token-level targets:
\begin{equation}
\mathcal{L}_{\mathrm{OPD}}(\theta) = \mathbb{E}\Bigl[ \mathbb{I}[R(\tau)<R_{\mathrm{succ}}] \, m_{\mathrm{act}} \, g_t \, (\operatorname{sg}[\ell_t^{\mathrm{tea}}] - \ell_t^{\mathrm{stu}}) \Bigr],
\label{eq:opd-main}
\end{equation}
where $g_t = \sigma(\beta_{\mathrm{opd}} \delta_t)$ is the confidence gating weight, $\delta_t = \ell_t^{\mathrm{tea}} - \ell_t^{\mathrm{stu}}$ measures the privileged log-probability gap, $\sigma(\cdot)$ is the sigmoid function, and $\beta_{\mathrm{opd}} > 0$ controls the gating temperature.

The overall joint training objective is formulated as:
\begin{equation}
\mathcal{L}(\theta) = \mathcal{L}_{\mathrm{GRPO}}(\theta) + \lambda_{\mathrm{opd}} \mathcal{L}_{\mathrm{OPD}}(\theta),
\label{eq:joint-objective}
\end{equation}
where $\lambda_{\mathrm{opd}} > 0$ balances RL exploration and token-level self-distillation.
When all rollouts within a group fail ($\sigma=0$), $\mathcal{L}_{\mathrm{GRPO}}$ vanishes while $\mathcal{L}_{\mathrm{OPD}}$ actively maintains informative policy gradients.

\paragraph{Inference.}
During deployment (Fig.~\ref{fig:overview}, top-right), Analyzer and Teacher branches are stripped away, executing $\pi_\theta$ strictly as an unprivileged Student $a_t \sim \pi_\theta(\cdot \mid h_t)$ with zero extra cost.

\section{Experiments}
\label{sec:exp}

\subsection{Experimental Setup}
\label{sec:setup}

\paragraph{Environments.}
To evaluate the learning dynamics and visual reasoning capabilities of VLM agents, we adopt the VAGEN~\citep{wang2025vagen} evaluation suite, comprising five agentic tasks across three paradigms: Cognitive Grid Puzzles, Embodied 3D Control, and Generative Reasoning. App.~\ref{app:envs} details the task families, environment specifications, horizon limits, and metrics.

\paragraph{Experimental Setup \& Reference Baselines.}
We evaluate \ourmodel{} across two backbone models: \texttt{Qwen2.5-VL-3B-Instruct}~\citep{bai2025qwen25vl} and \texttt{Qwen3-VL-2B-Instruct}~\citep{bai2025qwen3vl}.
We compare against three reference baselines:
(i) \textbf{\armGRPO{}}: Pure RL post-training initialized directly from raw Instruct checkpoints with Analyzer components disabled;
(ii) \textbf{\armSFTGRPO{}}: GRPO initialized from Stage~I Analyzer SFT weights $\theta_{\mathrm{sft}}$ (with OPD disabled during RL), isolating SFT initialization gains from token-level distillation; and
(iii) \textbf{SDAR}~\citep{lu2026sdar}: A reproduced state-of-the-art on-policy distillation baseline using a frozen \texttt{Qwen3-VL-7B-Instruct} teacher and Qwen2.5-VL-3B student (More details in App.~\ref{app:sdar}) with the same Gated OPD (Eq.~\ref{eq:opd-main}), isolating Gated OPD's contribution.

\paragraph{\ourmodel{} Variants \& Ablation Arms.}
All \ourmodel{} variants initialize $\pi_\theta$ from Stage~I weights $\theta_{\mathrm{sft}}$ and modulate the Teacher's privileged context $p$:
(i) \textbf{\armP{}}: Includes localized visual panel only, $p=(P_{\hat t})$;
(ii) \textbf{\armM{}}: Includes failure mode diagnosis only, $p=(\hat m)$;
(iii) \textbf{\armMP{}} (\emph{Default}): Combines localized visual panel, failure mode, and predicted pivot step, $p=(P_{\hat t},\hat m,\hat t)$; and
(iv) \textbf{\armMPR{}}: Extends \armMP{} by appending predicted guidance skill prompts, $p=(P_{\hat t},\hat m,\hat t,\hat r)$.
Additionally, Tab.~\ref{tab:pivot-effect} incorporates a negative control \textbf{\armMPRandR{}}, which replaces predicted pivot steps $\hat t$ with uniform random step indices to evaluate temporal localization sensitivity.

\paragraph{Metrics.}
Performance across puzzle and embodied control tasks is evaluated using the average Success Rate (SR) under sparse goal-completion rewards. For SVG Reconstruction, we report a composite visual similarity score averaging DINO and DreamSim embeddings.

\paragraph{Implementation Details.}
\textbf{\textit{SFT stage.}} For each environment we sample failed on-policy rollouts from the raw Instruct checkpoint and retain $\sim$960--1.7k accepted trajectories after loose factual filtering (Tab.~\ref{tab:sft-n}).
Each example pairs an trajectory collage with a compact action log; supervision is JSON $(t^\ast,m^\ast,r^\ast)$, with task solvers pinning the pivot step $t^\ast$ and failure mode $m^\ast$ and a frozen teacher optionally writing guidance skill $r^\ast$ (App.~\ref{app:hyper}).
We fine-tune the backbone for three epochs per environment to obtain a dedicated Analyzer $\theta_{\mathrm{sft}}$.
\textbf{\textit{RL stage.}} We set $\theta\leftarrow\theta_{\mathrm{sft}}$ for the shared Student ($\pi_\theta$), Analyzer, and Teacher. We run $250$ policy updates with a GRPO group size of $N{=}8$; the OPD loss ($\lambda_{\mathrm{opd}}{=}0.01$, gate $\beta_{\mathrm{opd}}{=}5$) is applied only on failed trajectories. Validation uses $128$ episodes per environment. Further SFT data, collage, optimization, reward and hyperparameter details, plus collages and JSON targets, are in Appendices~\ref{app:hyper} and~\ref{app:examples}.

\subsection{Main Results}
\label{sec:main}

\begin{table*}[t]
\centering
\caption{Main results across five VLM agentic benchmarks.
\armGRPO{}, \armSFTGRPO{}, and \armM{}/\armMP{}/\armMPR{} denote vanilla GRPO, GRPO after Analyzer SFT, and Teacher context with failure mode $\hat m$, pivot panel $P_{\hat t}$, or added skill text $\hat r$; SDAR~\citep{lu2026sdar} is our reproduced Gated OPD baseline.
Bold marks the best result; cyan percentages on \armMPR{} are relative gains over \armSFTGRPO{} in the same block; light-gray rows follow~\citep{wang2025vagen,li2026glance}.}
\label{tab:main}
\setlength{\tabcolsep}{2.4pt}
\arrayrulecolor{black}
\footnotesize
\newcommand{\dgrpo}[1]{\textcolor{cyan!50!black}{\tiny\,#1}}
\begin{adjustbox}{max width=\linewidth}
\begin{tabular}{@{}l cc ccc ccccc ccc c@{}}
\toprule
\multirow{3}{*}{Method} &
\multicolumn{2}{c}{Cognitive Grid Puzzles} &
\multicolumn{8}{c}{Embodied 3D Control} &
\multicolumn{3}{c}{Generative Reasoning} &
\multirow{3}{*}{All} \\
\cmidrule(lr){2-3}\cmidrule(lr){4-11}\cmidrule(lr){12-14}
& \multirow{2}{*}{Sokoban} &
\multirow{2}{*}{\shortstack[c]{Frozen\\Lake}} &
\multicolumn{3}{c}{Navigation} &
\multicolumn{5}{c}{PrimitiveSkill} &
\multicolumn{3}{c}{SVG} & \\
\cmidrule(lr){4-6}\cmidrule(lr){7-11}\cmidrule(lr){12-14}
& & & Base & Com. & Avg. & Place & Stack & Draw. & Align & Avg. & DINO & DS & Avg. & \\
\midrule
\multicolumn{15}{@{}l}{\textit{Open-source VLMs}} \\
\rowcolor{gray!8}
Qwen2.5-VL-72B & 0.20 & 0.44 & 0.70 & 0.77 & 0.74 & 1.00 & 0.50 & 0.00 & 1.00 & 0.63 & 0.84 & 0.62 & 0.73 & 0.55 \\
\rowcolor{gray!8}
Qwen2.5-VL-7B & 0.14 & 0.14 & 0.33 & 0.38 & 0.35 & 0.00 & 0.00 & 0.00 & 0.75 & 0.19 & 0.84 & 0.27 & 0.56 & 0.28 \\
\rowcolor{gray!8}
Qwen2.5-VL-3B~\citep{bai2025qwen25vl} & 0.13 & 0.14 & 0.20 & 0.26 & 0.23 & 0.00 & 0.00 & 0.00 & 0.00 & 0.00 & 0.79 & 0.30 & 0.54 & 0.21 \\
\rowcolor{gray!8}
VLM-R1-3B~\citep{shen2025vlm} & 0.16 & 0.15 & 0.33 & 0.34 & 0.34 & 0.00 & 0.00 & 0.00 & 0.00 & 0.00 & 0.79 & 0.27 & 0.54 & 0.24 \\
\midrule
\multicolumn{15}{@{}l}{\textit{Proprietary VLMs}} \\
\rowcolor{gray!8}
o4-mini & 0.44 & 0.82 & 0.75 & 0.75 & 0.75 & 1.00 & 0.50 & 0.00 & 0.75 & 0.56 & 0.90 & 0.66 & 0.78 & 0.67 \\
\rowcolor{gray!8}
GPT-4o & 0.43 & 0.54 & 0.75 & 0.69 & 0.72 & 0.50 & 0.63 & 0.00 & 0.88 & 0.50 & 0.91 & 0.69 & 0.80 & 0.60 \\
\rowcolor{gray!8}
Gemini~2.5~Pro & 0.58 & 0.78 & 0.63 & 0.63 & 0.63 & 0.63 & 0.63 & 0.00 & 0.75 & 0.50 & 0.93 & 0.78 & 0.86 & 0.67 \\
\rowcolor{gray!8}
Claude~4.5~Sonnet & 0.31 & 0.80 & 0.67 & 0.67 & 0.67 & 0.63 & 0.50 & 0.00 & 1.00 & 0.53 & 0.95 & 0.81 & 0.88 & 0.64 \\
\rowcolor{gray!8}
Claude~3.7~Sonnet & 0.25 & 0.69 & 0.48 & 0.47 & 0.47 & 0.63 & 0.13 & 0.00 & 1.00 & 0.44 & 0.94 & 0.77 & 0.85 & 0.54 \\
\midrule
\multicolumn{15}{@{}l}{\textit{Previous SOTA with World Model Reasoning for Visual State Reconstruction} (Qwen2.5-VL-3B-Instruct)} \\
\rowcolor{gray!8}
VAGEN-Full~\citep{wang2025vagen} & 0.79 & 0.72 & 0.80 & 0.81 & 0.81 & 1.00 & 0.88 & 1.00 & 1.00 & 0.97 & 0.90 & 0.66 & 0.78 & 0.81 \\
\rowcolor{gray!8}
\textsc{GLANCE}-Full~\citep{li2026glance} & 0.85 & 0.78 & 0.86 & 0.88 & 0.87 & 1.00 & 0.88 & 1.00 & 1.00 & 0.97 & 0.92 & 0.70 & 0.81 & 0.86 \\
\midrule
\multicolumn{15}{@{}l}{\textit{Reproduced OPD baseline} (Qwen2.5-VL-3B-Instruct)} \\
SDAR~\citep{lu2026sdar} & 0.52 & 0.77 & 0.88 & 0.84 & 0.86 & 1.00 & 1.00 & 1.00 & 1.00 & 1.00 & 0.91 & 0.68 & 0.80 & 0.79 \\
\midrule
\multicolumn{15}{@{}l}{\ourmodel{} (Qwen2.5-VL-3B-Instruct)} \\
\rowcolor{blue!4}
\armGRPO{} & 0.78 & 0.71 & 0.87 & 0.75 & 0.81 & 0.92 & 0.86 & 0.89 & 0.93 & 0.90 & 0.90 & 0.66 & 0.78 & 0.80 \\
\rowcolor{blue!4}
\armSFTGRPO{} & 0.82 & 0.75 & 0.93 & 0.77 & 0.85 & 0.96 & 0.90 & 0.93 & 0.97 & 0.94 & 0.92 & 0.68 & 0.80 & 0.83 \\
\rowcolor{blue!4}
\armM{} & 0.86 & 0.83 & 0.94 & 0.78 & 0.86 & \textbf{1.00} & \textbf{1.00} & \textbf{1.00} & \textbf{1.00} & \textbf{1.00} & \textbf{0.93} & 0.73 & 0.83 & 0.88 \\
\rowcolor{blue!4}
\armMP{} & 0.90 & 0.77 & 0.95 & \textbf{0.83} & 0.89 & \textbf{1.00} & \textbf{1.00} & \textbf{1.00} & \textbf{1.00} & \textbf{1.00} & \textbf{0.93} & \textbf{0.74} & \textbf{0.84} & 0.88 \\
\rowcolor{blue!8}
\armMPR{} & \textbf{0.95}\dgrpo{+16\%} & \textbf{0.84}\dgrpo{+12\%} & \textbf{0.97}\dgrpo{+4\%} & \textbf{0.83}\dgrpo{+8\%} & \textbf{0.90}\dgrpo{+6\%} & \textbf{1.00}\dgrpo{+4\%} & \textbf{1.00}\dgrpo{+11\%} & \textbf{1.00}\dgrpo{+8\%} & \textbf{1.00}\dgrpo{+3\%} & \textbf{1.00}\dgrpo{+6\%} & \textbf{0.93}\dgrpo{+1\%} & 0.71\dgrpo{+4\%} & 0.82\dgrpo{+3\%} & \textbf{0.90}\dgrpo{+8\%} \\
\midrule
\multicolumn{15}{@{}l}{\ourmodel{} (Qwen3-VL-2B-Instruct)} \\
\rowcolor{green!4}
\armGRPO{} & 0.77 & 0.89 & 0.97 & 0.73 & 0.85 & 0.77 & 0.77 & 0.77 & 0.77 & 0.77 & 0.90 & 0.68 & 0.79 & 0.81 \\
\rowcolor{green!4}
\armSFTGRPO{} & 0.80 & 0.87 & 0.92 & 0.74 & 0.83 & 0.83 & 0.80 & 0.81 & 0.84 & 0.82 & 0.91 & 0.67 & 0.79 & 0.82 \\
\rowcolor{green!4}
\armM{} & 0.82 & 0.81 & 0.93 & 0.75 & 0.84 & \textbf{1.00} & \textbf{1.00} & \textbf{1.00} & \textbf{1.00} & \textbf{1.00} & 0.92 & 0.70 & 0.81 & 0.86 \\
\rowcolor{green!4}
\armMP{} & \textbf{0.97} & 0.89 & 0.97 & 0.80 & 0.89 & \textbf{1.00} & \textbf{1.00} & \textbf{1.00} & \textbf{1.00} & \textbf{1.00} & \textbf{0.93} & 0.71 & 0.82 & 0.91 \\
\rowcolor{green!8}
\armMPR{} & 0.92\dgrpo{+15\%} & \textbf{0.90}\dgrpo{+3\%} & \textbf{0.98}\dgrpo{+7\%} & \textbf{0.91}\dgrpo{+23\%} & \textbf{0.95}\dgrpo{+14\%} & \textbf{1.00}\dgrpo{+20\%} & \textbf{1.00}\dgrpo{+25\%} & \textbf{1.00}\dgrpo{+23\%} & \textbf{1.00}\dgrpo{+19\%} & \textbf{1.00}\dgrpo{+22\%} & \textbf{0.93}\dgrpo{+2\%} & \textbf{0.76}\dgrpo{+13\%} & \textbf{0.85}\dgrpo{+8\%} & \textbf{0.92}\dgrpo{+12\%} \\
\bottomrule
\end{tabular}
\end{adjustbox}
\end{table*}

\paragraph{Main Results \& SOTA Comparison.}
As presented in Tab.~\ref{tab:main}, the full \ourmodel{} framework (\armMPR{}) consistently outperforms all reference baselines across both backbones.
On \texttt{Qwen2.5-VL-3B}, \armMPR{} achieves a \textbf{$0.90$ overall accuracy}, delivering a $+8\%$ gain over \armSFTGRPO{} ($0.83 \to 0.90$) and a $+13\%$ lift over pure \armGRPO{}.
Crucially, \ourmodel{} surpasses both the dense-reward SOTA (\textsc{GLANCE}-Full, $0.86$) and the reproduced on-policy distillation baseline SDAR ($0.79$; App.~\ref{app:sdar}), achieving these state-of-the-art results with zero inference-time overhead.
The gains are most pronounced in \emph{cognitive grid puzzles} and \emph{embodied control}—where early pivot blunders render trajectory suffixes unrecoverable—highlighted by Sokoban surging from $0.82$ to $0.95$ and PrimitiveSkill reaching $1.00$ (vs.\ $0.94$ under \armSFTGRPO{}).

\paragraph{Domain Insights \& Backbone Scaling.}
\ourmodel{} demonstrates strong generalizability and progressive scaling behavior across task domains and model families.
On \emph{generative reasoning} (SVG) with Qwen2.5-VL-3B, \armMP{} achieves a peak composite score of $0.84$; appending guidance skill prompts (\armMPR{}) keeps DINO at $0.93$ but lowers DreamSim ($0.74{\rightarrow}0.71$).
On \texttt{Qwen3-VL-2B} (\armSFTGRPO{} at $0.82$), enriching the Teacher's privileged context yields continuous performance lifts: \armM{} ($0.86$) $\to$ \armMP{} ($0.91$) $\to$ \armMPR{} (\textbf{$0.92$ overall}, $+12\%$ over \armSFTGRPO{}).
Under \armMPR{}, task-level scores reach $0.90$ on FrozenLake, $0.95$ on Navigation, and $0.85$ on SVG; on Sokoban, \armMP{} attains the peak ($0.97$).

\subsection{Ablation Study}
\label{sec:ablation}
\vspace{-8pt}

\noindent
\begin{minipage}[t]{0.64\textwidth}
\vspace{0pt}
\noindent\textbf{Impact of Pivot Step Localization.}
Tab.~\ref{tab:pivot-effect} dissects diagnostic contributions.
Compared to Vanilla-GRPO ($0.78$ Sokoban / $0.81$ Navigation), single channels $P$ and $M$ both raise accuracy to $\sim0.86$, their combination ($M+P$) reaches $0.90 / 0.89$, and the full stack ($M+P+R$) peaks at $0.95 / 0.90$.
Crucially, substituting the localized step $t^\ast$ with a random frame ($M+P\text{-Random}+R$) degrades performance below Vanilla-GRPO ($0.75 / 0.80$).
This negative control confirms that arbitrary visual context introduces harmful noise, proving precise pivot localization indispensable.
\end{minipage}\hfill
\sbox{\pivottabbox}{%
\small
\setlength{\tabcolsep}{4pt}%
\renewcommand{\arraystretch}{1.02}%
\begin{tabular}{@{}l cc@{}}
\toprule
Variant & Sok. & Nav. \\
\midrule
\armGRPO{} & 0.78 & 0.81 \\
\armP{} & 0.86 & 0.85 \\
\armM{} & 0.86 & 0.86 \\
\armMP{} & 0.90 & 0.89 \\
\rowcolor{blue!8}
\armMPR{} & \textbf{0.95} & \textbf{0.90} \\
\armMPRandR{} & 0.75 & 0.80 \\
\bottomrule
\end{tabular}}%
\begin{minipage}[t]{\wd\pivottabbox}
\vspace{0pt}
\centering
\usebox{\pivottabbox}
\vspace{-3pt}
\captionsetup{skip=5pt,aboveskip=5pt,belowskip=1pt,font=small,justification=centering,singlelinecheck=true}
\captionof{table}{Pivot-step ablation.}
\label{tab:pivot-effect}
\vspace{-2pt}
\end{minipage}

\noindent\textbf{Optimal Diagnostic Context.}
Tab.~\ref{tab:main} shows that progressively enriching privileged context from $M$ to $M+P+R$ delivers strong multimodal synergy, driving overall performance to $0.90$ on Qwen2.5-VL-3B ($+8\%$ over \armSFTGRPO{}) and $0.92$ on Qwen3-VL-2B ($+12\%$).
The full $M+P+R$ stack peaks on FrozenLake, Navigation, and PrimitiveSkill on both backbones, and on Sokoban for 3B ($0.95$); on Qwen3-VL-2B, $M+P$ attains the Sokoban peak ($0.97$).
SVG is mixed: on 3B the predicted guidance skill $\hat r$ slightly degrades the composite ($0.84{\rightarrow}0.82$), whereas on 2B $M+P+R$ is best ($0.85$).
Overall, pairing visual state restoration with failure modes and guidance skills forms the primary foundation for effective credit assignment.

\subsection{Further Analysis}
\label{sec:further-analysis}
\vspace{-8pt}

\noindent
\begin{minipage}[t]{0.61\textwidth}
\vspace{0pt}
\noindent\textbf{Self-Evolving Pivot Localization.}
Fig.~\ref{fig:frozenlake-pivot-acc} illustrates the co-evolution of the Analyzer's diagnostic accuracy ($\hat{t} = t^\ast$) alongside policy optimization on FrozenLake, as this environment shortest-path solver provides ground-truth pivot labels $t^\ast$ on failed rollouts.
Under \armMPR{}, 15-step moving accuracy starts near 0.56, rises after step 50 to $0.90$ by step 100, and later holds near $0.95$.
This trajectory confirms the Analyzer is genuinely self-evolving, forming a loop where sharper pivot localization yields finer credit.
\end{minipage}\hfill
\begin{minipage}[t]{0.39\textwidth}
\vspace{0pt}
\centering
\captionsetup{skip=5pt,aboveskip=5pt,belowskip=1pt,font=small,justification=centering,singlelinecheck=true}
\includegraphics[width=0.78\linewidth]{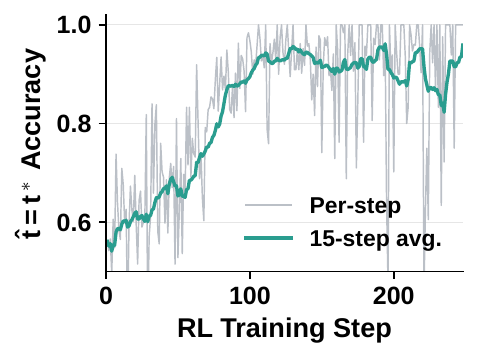}
\vspace{-8pt}
\captionof{figure}{FrozenLake $\hat t$ accuracy.}
\label{fig:frozenlake-pivot-acc}
\vspace{-4pt}
\end{minipage}

\noindent
\begin{minipage}[t]{0.61\textwidth}
  \noindent\textbf{Training Efficiency.}
  Fig.~\ref{fig:efficiency} compares the average per-step runtime during RL optimization on Sokoban. Despite incorporating a multi-role diagnostic branch, \ourmodel{} achieves a lower per-update latency ($95$\,s) than standard \armGRPO{} ($102$\,s). This advantage stems from \ourmodel{} rapidly suppressing unviable actions and converging toward shorter, successful trajectories, which significantly reduces rollout sampling time as it's the primary computational bottleneck in VLM RL. In contrast, delegating hindsight diagnosis to an external frozen \texttt{Qwen3-VL-8B} teacher introduces cross-model inference overhead, inflating the step time to $373$\,s ($3.9\times$). At test time, stripping all diagnostic branches leaves an unprivileged student with zero parameter or inference overhead.
\end{minipage}\hfill
\begin{minipage}[t]{0.39\textwidth}
\vspace{0pt}
\centering
\captionsetup{skip=5pt,aboveskip=5pt,belowskip=1pt,font=small,justification=centering,singlelinecheck=true}
\includegraphics[width=0.78\linewidth]{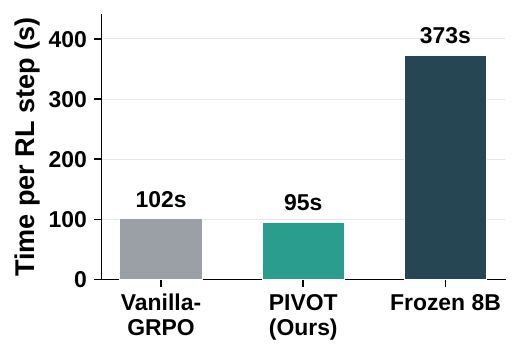}
\captionof{figure}{Time per RL step (s).}
\label{fig:efficiency}
\end{minipage}

\section{Related Work}
\label{sec:related}
\paragraph{RL and Hindsight Distillation for VLM Agents.}
VLMs serve as goal-driven agents across web navigation, device control, and embodied decision-making tasks~\citep{koh2024visualwebarena,shridhar2021alfworld}.
Beyond test-time prompting with frozen backbones~\citep{yao2023react,shinn2023reflexion}, task adaptability is enhanced via fine-tuning and online RL~\citep{bai2025qwen25vl,shen2025vlm,ouyang2022instructgpt}.
However, standard outcome RL and process supervision~\citep{schulman2017proximal,shao2024deepseekmath,guo2025deepseekr1,li2026glance,lightman2024prm,wang2024mathshepherd} rely on trial-and-error or dense shaping, providing coarse feedback that fails to isolate unrecoverable decision points.
Retrospective feedback from hindsight, reflection, and memory systems~\citep{andrychowicz2017hindsight,madaan2023selfrefine,zhao2024expel,wang2023voyager} yields post-hoc signals but incurs severe prompt-length and latency overhead at inference.
While on-policy distillation internalizes feedback into parameters via self-generated rollouts~\citep{ross2011dagger,agarwal2024gkd,lu2026sdar,wang2026skillsd}, agentic self-teachers remain structurally decoupled or prompt-dependent~\citep{zhong2026seed}.
In contrast, \ourmodel{} localizes pivot turns from visual histories and distills token-level hindsight directly into policy parameters without dense shaping, extra prompts, or environment rollbacks.

\paragraph{Policy Entropy Collapse and Multi-turn Credit Assignment.}
Outcome-based RL frequently suffers from policy entropy collapse, prematurely shrinking pass@$k$ exploration diversity and trapping policies in local optima~\citep{yue2025limitrlvr,cui2025entropy,yu2025dapo}.
This degeneration is acute in multi-turn agent interaction~\citep{wang2025vagen,li2026glance}, where standard policy gradients over-reinforce frequent success paths while suppressing alternative exploration branches~\citep{schulman2017proximal,cui2025entropy}.
In sequential settings, an uncorrected early blunder distorts subsequent state observations, compounding downstream execution errors and pruning viable subtrees~\citep{andrychowicz2017hindsight,sinha2026ips}.
Although recent diversity-preserving techniques attempt to mitigate policy sharpening, they remain bound to trajectory-level returns or group-level advantages~\citep{sinha2026ips,shao2024deepseekmath,yu2025dapo,kazemnejad2025vineppo}.
Consequently, they sustain only surface-level diversity without identifying the exact pivot turn where task failure originated~\citep{uesato2022process,lightman2024prm}.
\ourmodel{} bridges this credit-assignment gap by isolating critical pivot steps via visual panels and distilling confidence-gated token-level gradients onto failed rollouts, maintaining informative optimization even within zero-variance failure groups where conventional RL advantages vanish.

\section{Discussion and Limitations}
\label{sec:limit}
While \ourmodel{} demonstrates strong performance and efficiency across multi-turn VLM agent tasks, several scope boundaries and promising directions for future work remain:

\begin{itemize}
    \item \textbf{Pivot Supervision Source \& Discovery:} Stage I SFT currently relies on offline target labels generated via feasibility heuristics. Extending \ourmodel{} to complex environments lacking explicit solvers presents an exciting opportunity for unsupervised or self-diagnostic pivot discovery. Inspired by recent insights on token-level uncertainty and self-adaptation~\citep{ding2026does}, the pivot step $t^\ast$ could be determined intrinsically using token-level entropy metrics (e.g., locating critical state transitions via minimum-entropy positions), or empirically via rollout sampling by identifying the step where the suffix return lift growth is minimized.
    \item \textbf{Task Domain Applicability:} The current definition of the pivot step $t^\ast$ relies on remaining-budget goal reachability, which naturally aligns with structured visual environments. Adapting it to open-ended dialogue or continuous control may use dynamic, probabilistic, or entropy-guided reachability.
    \item \textbf{Model Scale and Physical Realization:} Our evaluation focuses on compact VLM backbones (Qwen2.5-VL-3B and Qwen3-VL-2B) within simulated benchmarks. Scaling this internalized distillation recipe to larger parameters and non-rewindable physical robotic systems represents an important direction for future exploration.
\end{itemize}

\section{Conclusion}
\label{sec:concl}

In this work, we address the credit assignment bottleneck of outcome-based RL in multi-turn VLM agents. 
Our analysis shows that post-failure recovery is driven by physical state restoration at pivot steps ($t^\ast$), rather than textual skill guidance. 
To avoid prohibitive simulator rollbacks, we present \ourmodel{}, which internalizes visual state restoration into a unified Student-Analyzer-Teacher policy. 
By localizing $t^\ast$ and distilling gated hindsight onto failed rollouts, \ourmodel{} removes physical resets at training and prompt overhead at deployment, outperforming SOTAs and prior distillation methods.


\newpage
\bibliography{iclr2027_conference}
\bibliographystyle{iclr2027_conference}

\newpage
\appendix
\setcounter{secnumdepth}{1}
\newcommand{\appblock}[1]{\par\noindent\textbf{#1} }
\newcommand{\appenvheading}[1]{\par\noindent\textbf{\large #1}\par}
\begin{center}
{\LARGE\sc PIVOT: Pivot-Aware On-Policy Self-distillation for Multi-Turn VLM Agents\par}
{\LARGE \emph{--Supplementary Material (Text)--}}
\end{center}

\section*{Appendix Contents}
\label{app:contents}
\newcommand{\apptocline}[1]{%
  \noindent
  \hyperref[#1]{\ref{#1}\hspace{0.75em}\nameref{#1}}%
  \nobreak\dotfill\nobreak
  \hyperref[#1]{\pageref{#1}}\par
}
{\setlength{\parskip}{0.28em}\setlength{\parindent}{0pt}%
\apptocline{app:notation}%
\apptocline{app:certs}%
\apptocline{app:sec2}%
\apptocline{app:cf-recovery}%
\apptocline{app:envs}%
\apptocline{app:grpo}%
\apptocline{app:hyper}%
\apptocline{app:examples}%
\apptocline{app:sdar}%
\apptocline{app:curves}%
}

\section{Notation}
\label{app:notation}

Tab.~\ref{tab:notation} lists symbols used in Sec.~\ref{sec:pivot-analysis}--\ref{sec:loss} and their meanings.

\begin{table}[t]
\centering
\small
\setlength{\tabcolsep}{5pt}
\caption{Notation (Sec.~\ref{sec:pivot-analysis}--\ref{sec:loss}).}
\label{tab:notation}
\begin{tabularx}{\linewidth}{lX}
\toprule
\textbf{Symbol} & \textbf{Meaning} \\
\midrule
$\mathcal{M}$ & Partially observable Markov decision process. \\
$\mathcal{S},\mathcal{O},\mathcal{A}$ & Latent-state, observation, and action spaces. \\
$\mathcal{P},\Omega$ & State-transition and observation kernels. \\
$\mathcal{R},\gamma$ & Per-step reward function and discount factor. \\
$q,\mathcal{Q}$ & Task prompt and task-prompt distribution. \\
$s_t,o_t,a_t,r_t$ & Latent state, observation, action, and reward at turn $t$. \\
$h_t$ & Observable interaction history $(o_0,a_0,\ldots,o_t)$ (Sec.~\ref{sec:grpo-blindspot}). \\
$\tilde h_t$ & Privileged Teacher history with diagnostic context prepended, $\tilde h_t=(p,h_t)$ (Eq.~\eqref{eq:teacher}). \\
$\pi_\theta$ & Shared autoregressive VLM: Student $a_t\sim\pi_\theta(\cdot\mid h_t)$; Analyzer outputs $(\hat t,\hat m,\hat r)$ from $x_\tau$ (Eq.~\eqref{eq:analyzer}); Teacher supplies $\ell^{\mathrm{tea}}$ under $\tilde h_t$. \\
$\theta$ & Trainable weights shared by all roles of $\pi_\theta$. \\
$J(\theta)$ & Outcome-based objective $\mathbb{E}_{q\sim\mathcal{Q},\,\tau\sim\pi_\theta}[R(\tau)]$. \\
$\tau,T,T_{\max}$ & Trajectory, realized trajectory length, and maximum horizon. \\
$R(\tau)$ & Discounted trajectory return $\sum_{t=0}^{T-1}\gamma^t r_t$. \\
$R_{\mathrm{succ}}$ & Success-return threshold; rollouts with $R(\tau)<R_{\mathrm{succ}}$ are failures for OPD (Eq.~\eqref{eq:opd-main}). \\
$N^{-}$ & Failed-rollout set $\{\tau:R(\tau)<R_{\mathrm{succ}}\}$; $|N^{-}|$ is its cardinality. \\
$\mathcal{G}_q$ & On-policy GRPO group $\{\tau_q^{(1)},\ldots,\tau_q^{(N)}\}$ for prompt $q$. \\
$N$ & Number of on-policy trajectories sampled per prompt. \\
$\mu,\sigma,\epsilon$ & Group return mean, standard deviation, and stabilizer in Eq.~\eqref{eq:grpo-adv} (distinct from clip $\varepsilon$). \\
$A_{n}^{\mathrm{rl}}$ & Group-relative advantage of rollout $n$; per-prompt form $A_{q,n}^{\mathrm{rl}}$. \\
$k_t$ & Remaining action budget after step $t$ in Eq.~\eqref{eq:pivot}: $T-t-1$. \\
$\mathsf{Feas}(s,k)$ & Indicator that state $s$ is goal-reachable within $k$ actions. \\
$t^\ast$ & Pivot step (Eq.~\eqref{eq:pivot}): first infeasible action, or $T{-}1$ if none. \\
$\tau_{t^\ast:}$ & Original trajectory suffix from pivot step $t^\ast$. \\
$\tilde\tau_{t^\ast:}$ & Resampled suffix after state restoration at $t^\ast$. \\
$\Delta R$ & Suffix-return lift rate in Eq.~\eqref{eq:lift-rate} (distinct from the OPD log-prob gap $\delta_t$). \\
$\mathcal{M}_{\mathrm{fail}}$ & Closed vocabulary of discrete failure modes. \\
$m^\ast,r^\ast$ & Ground-truth failure mode and optional guidance skill in SFT labels. \\
$\hat t,\hat m,\hat r$ & Analyzer-predicted pivot, failure mode, and guidance skill. \\
$z_\tau$ & SFT diagnosis triplet $(t^\ast,m^\ast,r^\ast)$; $r^\ast$ may be empty. \\
$C(\tau)$ & Full-trajectory visual collage $C(\tau)=\mathrm{Grid}(o_0,\ldots,o_{T-1})$ with step labels (Sec.~\ref{sec:sft}). \\
$\acttxt_{0:T-1}$ & Compact textual action transcript fed to the Analyzer (serializes the action sequence $a_{0:T-1}$ in Eqs.~\eqref{eq:analyzer} and~\eqref{eq:sft-loss}). \\
$x_\tau$ & Serialized Analyzer input $(q,C(\tau),\acttxt_{0:T-1})$. \\
$P_{\hat t}$ & Local visual panel $[o_{\hat t-1},o_{\hat t},o_{\hat t+1}]$ (boundary frames zero-padded; Sec.~\ref{sec:loss}). \\
$p$ & Teacher privileged context; ablation arms use $(P_{\hat t})$, $(P_{\hat t},\hat m)$, $(P_{\hat t},\hat m,\hat t)$, or $(P_{\hat t},\hat m,\hat t,\hat r)$ (Sec.~\ref{sec:setup}). \\
$\pi_{\mathrm{base}}$ & Fixed untrained instruct checkpoint for offline rollouts and the Sec.~\ref{sec:cf-recovery} probe. \\
$\pi_{\theta_{\mathrm{old}}}$ & Frozen behavior policy for on-policy collection; synced from $\theta$ after each RL update. \\
$\nu_\tau$ & Loose factual-filter accept flag (App.~\ref{app:cf}). \\
$\mathcal{D}_{\mathrm{sft}}$ & Accepted SFT set after loose filtering. \\
$m_{\mathrm{act}}$ & Valid-action token mask (distinct from failure mode $m^\ast$). \\
$\ell^{\mathrm{stu}},\ell^{\mathrm{old}},\ell^{\mathrm{tea}}$ & Student, frozen-behavior, and Teacher token log-probabilities. \\
$\rho$ & Token-level importance ratio $\exp(\ell^{\mathrm{stu}}-\ell^{\mathrm{old}})$. \\
$\varepsilon$ & GRPO PPO-style clip radius in Eq.~\eqref{eq:grpo-loss}. \\
$\delta_t,g_t$ & Privileged log-prob gap $\delta_t=\ell^{\mathrm{tea}}_t-\ell^{\mathrm{stu}}_t$ and gate $g_t=\sigma(\beta_{\mathrm{opd}}\delta_t)$ (Eq.~\eqref{eq:opd-main}; token-level form in Eq.~\eqref{eq:opd-gate}). \\
$\beta_{\mathrm{opd}},\lambda_{\mathrm{opd}}$ & OPD gate sharpness and joint-loss weight. \\
\bottomrule
\end{tabularx}
\end{table}

\section{Pivot Certificates}
\label{app:certs}

Offline Analyzer SFT and counterfactual probes pin a diagnostic pair $(t^\ast,m^\ast)$ on each failed rollout.
Tab.~\ref{tab:modes} lists the certificate used per VAGEN task; all modes lie in the closed vocabulary $\mathcal{M}_{\mathrm{fail}}$ referenced in Sec.~\ref{sec:sft}.

\appblock{Pivot step $t^\ast$.}
When a feasibility test $\mathsf{Feas}(s,k)\in\{0,1\}$ is defined (Sokoban, FrozenLake, Navigation proxy, PrimitiveSkill stage budget), we use the main-text pivot step Eq.~\eqref{eq:pivot}: the first turn after which the goal is unreachable within the remaining $k_t=T-t-1$ actions, defaulting to $T{-}1$ if the trajectory never becomes infeasible.
Navigation may pin an \emph{earlier} $t^\ast$ when a terminal event fires first (invalid-action suffix, blocked move, or stagnation window); only if none fire do we fall back to the travel-time proxy in Eq.~\eqref{eq:pivot}.
SVG does not use spatial $\mathsf{Feas}$: each step submits a full canvas, so we set $t^\ast$ to the index of the last scored submission (typically $T{-}1$ with $T_{\max}{=}2$).

\appblock{Failure mode $m^\ast$ on grid puzzles.}
For Sokoban and FrozenLake, after $t^\ast$ from Eq.~\eqref{eq:pivot} we label timeout versus irreversible deadlock with two reachability checks at the post-action state $s_{t^\ast+1}$:
\begin{equation}
m^\ast=
\begin{cases}
\texttt{timeout}, & \mathsf{Feas}(s_{t^\ast+1},k_{t^\ast})=0
\ \wedge\ \mathsf{Feas}(s_{t^\ast+1},\infty)=1,\\[0.35em]
\texttt{deadlock}, & \mathsf{Feas}(s_{t^\ast+1},\infty)=0,
\end{cases}
\label{eq:failure-mode}
\end{equation}
where $k_{t^\ast}=T-t^\ast-1$ and the $\mathsf{Feas}(\cdot,\infty)$ probe uses unbounded search on grid worlds.

\appblock{Failure modes on other benchmarks.}
Eq.~\eqref{eq:failure-mode} does \emph{not} apply outside the two grid puzzles.
Navigation assigns the earliest matching event among \texttt{format}, \texttt{blocked}, and \texttt{stagnation}; if none occur, \texttt{budget} marks insufficient remaining moves under a Euclidean travel-time lower bound.
PrimitiveSkill compares unfinished manipulation stages with the remaining turn budget (\texttt{timeout}) and may further tag format, target, or ordering mistakes at $t^\ast$.
SVG reads parsing validity and DINO/DreamSim progression on the final canvas (\texttt{format}, \texttt{regression}, \texttt{stagnation}, \texttt{mismatch}).
Tab.~\ref{tab:modes} summarizes each environment in one place.

\begin{table}[t]
\centering
\footnotesize
\setlength{\tabcolsep}{4pt}
\renewcommand{\arraystretch}{1.12}
\caption{Offline pivot certificate $(t^\ast,m^\ast)$ by environment (Eq.~\eqref{eq:pivot} when applicable; Eq.~\eqref{eq:failure-mode} for Sokoban/FrozenLake only).}
\label{tab:modes}
\begin{tabularx}{\linewidth}{@{}>{\raggedright\arraybackslash}p{0.14\linewidth} >{\raggedright\arraybackslash}X >{\raggedright\arraybackslash}p{0.30\linewidth}@{}}
\toprule
\textbf{Environment} & \textbf{Pivot} $t^\ast$ & \textbf{Failure modes} $m^\ast$ \\
\midrule
Sokoban & Eq.~\eqref{eq:pivot}; $\mathsf{Feas}$ = box-pushing plan $\le k$ & \texttt{timeout}, \texttt{deadlock} \\
FrozenLake & Eq.~\eqref{eq:pivot}; $\mathsf{Feas}$ = hole-free path $\le k$ & \texttt{timeout}, \texttt{deadlock} \\
Navigation & earliest terminal event; else Eq.~\eqref{eq:pivot} (travel-time proxy) & \texttt{format}, \texttt{blocked}, \texttt{stagnation}, \texttt{budget} \\
PrimitiveSkill & Eq.~\eqref{eq:pivot}; unfinished stages vs.\ remaining turns & \texttt{timeout}, \texttt{format}, \newline \texttt{wrong\_target}, \texttt{order\_error} \\
SVG & index of last scored canvas & \texttt{format}, \texttt{regression}, \newline \texttt{stagnation}, \texttt{mismatch} \\
\bottomrule
\end{tabularx}
\end{table}

\section{GRPO Grouping and Zero-Gradient Groups}
\label{app:sec2}

We form GRPO groups as contiguous blocks of $N{=}8$ rollouts that share the same prompt key $(\texttt{task\_id},\texttt{goal\_idx},\texttt{eval\_set})$.
Blocks with fewer than eight rollouts are discarded, leaving $720$ of $1680$ Navigation rollouts and seven PrimitiveSkill rollouts unused.
Tab.~\ref{tab:grpo-waste} reports two group-level rates: \textbf{All-fail} is the fraction of groups with no successful rollout; \textbf{$\sigma{=}0$} is the fraction whose eight returns match at $10^{-8}$ rounding.
These two rates need not agree when failures receive different partial credit, as on FrozenLake and SVG.

\begin{table}[t]
\centering
\small
\setlength{\tabcolsep}{3pt}
\begin{minipage}[t]{0.58\linewidth}
\centering
\caption{Untrained 3B GRPO groups ($N{=}8$).}
\label{tab:grpo-waste}
\begin{tabular}{lrrrrr}
\toprule
Environment & Traj. & Grps. & SR (\%) & All-fail (\%) & $\sigma{=}0$ (\%) \\
\midrule
Sokoban & 1440 & 180 & 17.8 & 48.3 & 48.3 \\
FrozenLake & 1728 & 216 & 20.3 & 39.4 & 0.0 \\
Navigation & 960 & 120 & 43.0 & 5.8 & 0.0 \\
PrimitiveSkill & 2912 & 364 & 8.9 & 46.2 & 16.5 \\
SVG & 1728 & 216 & 0.1 & 99.5 & 0.9 \\
\midrule
Pooled & 8768 & 1096 & --- & 51.3 & 13.6 \\
\bottomrule
\end{tabular}
\end{minipage}
\hfill
\begin{minipage}[t]{0.38\linewidth}
\centering
\caption{Counterfactual rollback arms.}
\label{tab:cf-arms}
\begin{tabular}{@{}l>{\raggedright\arraybackslash}p{0.62\linewidth}@{}}
\toprule
Arm & Injected at restore \\
\midrule
no-hint & none \\
3B-self & $m^\ast$ pinned; $\hat r$ from untrained 3B \\
8B-teach. & $m^\ast$ pinned; $\hat r$ from \texttt{Qwen3-VL-8B} \\
\bottomrule
\end{tabular}
\end{minipage}
\end{table}

\begin{table}[t]
\centering
\small
\setlength{\tabcolsep}{3pt}
\begin{minipage}[t]{0.63\linewidth}
\centering
\caption{Suffix lift $\Delta R$ at $t^\ast$ ($n{=}1$; \% with $R(\tilde\tau_{t^\ast:})>R(\tau_{t^\ast:})$).}
\label{tab:cf-recovery}
\begin{tabular}{lrrrr}
\toprule
Environment & Fail & no-hint & 3B-self & 8B-teach. \\
\midrule
Sokoban & 1165 & 3.3 & 4.1 & 4.4 \\
FrozenLake & 1378 & 18.2 & 16.8 & 19.5 \\
Navigation & 969 & 56.6 & 50.8 & 52.7 \\
PrimitiveSkill & 2898 & 17.7 & 20.3 & 19.6 \\
SVG & 1727 & 22.4 & 22.8 & 24.3 \\
\bottomrule
\end{tabular}
\end{minipage}
\hfill
\begin{minipage}[t]{0.34\linewidth}
\centering
\caption{$n{=}1$ no-hint lift at $t^\ast$ vs.\ $t^\ast{+}1$ (\%).}
\label{tab:cf-tstarplus1}
\begin{tabular}{lrr}
\toprule
Environment & $t^\ast$ & $t^\ast{+}1$ \\
\midrule
Sokoban & 3.3 & 0.9 \\
FrozenLake & 18.2 & 8.1 \\
Navigation & 56.6 & 53.6 \\
PrimitiveSkill & 17.7 & 8.3 \\
SVG & 22.4 & 16.2 \\
\bottomrule
\end{tabular}
\end{minipage}
\end{table}

\section{Counterfactual Recovery}
\label{app:cf-recovery}

\appblock{Counterfactual recovery protocol.}
Tab.~\ref{tab:cf-arms} summarizes the rollback arms used in Sec.~\ref{sec:cf-recovery}.
Every arm starts from the same failed untrained \texttt{Qwen2.5-VL-3B-Instruct} rollouts, restores the environment at the solver-pinned pivot step $t^\ast$ (Eq.~\eqref{eq:pivot}), and resamples the suffix with $\pi_{\mathrm{base}}$.
A retry counts as a lift when the new suffix return strictly exceeds the original suffix return from $t^\ast$.
By default we restore at $t_{\mathrm{restore}}{=}t^\ast$ and roll forward to the horizon; the $t^\ast{+}1$ control instead restores one step later as a negative control, and lift collapses on all five environments.
Hinted arms inject solver-pinned $m^\ast$ plus an arm-specific guidance string into the actor prompt at restore; the no-hint arm skips injection entirely.
Tab.~\ref{tab:cf-recovery} gives the main-text $n{=}1$ results at $t^\ast$, and Tab.~\ref{tab:cf-tstarplus1} reports the matching no-hint run at $t^\ast{+}1$.

\section{Environments and Rewards}
\label{app:envs}

\begin{table}[t]
\centering
\small
\setlength{\tabcolsep}{3pt}
\newsavebox{\apptabenvsbox}
\newsavebox{\apptabrewardsbox}
\newsavebox{\apptabsftbox}
\newsavebox{\apptabhyperbox}
\savebox{\apptabenvsbox}{%
\begin{minipage}[t]{0.47\linewidth}
\centering
\caption{VAGEN tasks: $T_{\max}$ and validation metrics.}
\label{tab:envs}
\footnotesize
\begin{tabularx}{\linewidth}{@{}>{\raggedright\arraybackslash}X c >{\raggedright\arraybackslash}X@{}}
\toprule
\textbf{Task} & $T_{\max}$ & \textbf{Metric} \\
\midrule
\multicolumn{3}{@{}l}{\textit{Cognitive Grid}} \\
Sokoban $[6,6]$, 1 box & 9 & success rate \\
FrozenLake $4{\times}4$ & 9 & success rate \\
\midrule
\multicolumn{3}{@{}l}{\textit{Embodied 3D}} \\
Navigation (Base / Com.) & 25 & avg.\ SR \\
PrimitiveSkill (4 skills) & 6 & mean SR \\
\midrule
\multicolumn{3}{@{}l}{\textit{Generative}} \\
SVG reconstruction & 2 & mean DINO \& DreamSim \\
\bottomrule
\end{tabularx}
\end{minipage}}%
\savebox{\apptabrewardsbox}{%
\begin{minipage}[t]{0.47\linewidth}
\centering
\caption{Episode return $R(\tau)$ (GRPO).}
\label{tab:rewards}
\footnotesize
\begin{tabularx}{\linewidth}{@{}l >{\raggedright\arraybackslash}X@{}}
\toprule
\textbf{Env} & \textbf{Reward structure} \\
\midrule
Sokoban & $-0.1$/step; $+11$ goal; fail $R=-0.9$ \\
FrozenLake & $-0.1$/step; sparse goal \\
Navigation & $-0.1$/step; $+10$ on target \\
PrimitiveSkill & $-0.1$/step; $+2$/stage; $+10$ success \\
SVG & last canvas: $\tfrac{1}{2}(\mathrm{DINO}+\mathrm{DS})$ \\
\bottomrule
\end{tabularx}
\end{minipage}}%
\savebox{\apptabsftbox}{%
\begin{minipage}[t]{0.47\linewidth}
\centering
\caption{Analyzer SFT splits ($90/10$; 3B packs, same recipe on 2B).}
\label{tab:sft-n}
\begin{tabular}{lrrr}
\toprule
Environment & Acc. & Train & Val \\
\midrule
Sokoban & 1170 & 1053 & 117 \\
FrozenLake & 1375 & 1237 & 138 \\
Navigation & 959 & 863 & 96 \\
PrimitiveSkill & 1677 & 1509 & 168 \\
SVG & 1719 & 1547 & 172 \\
\bottomrule
\end{tabular}
\end{minipage}}%
\savebox{\apptabhyperbox}{%
\begin{minipage}[t]{0.47\linewidth}
\centering
\caption{Stage~II RL hyperparameters (3B \& 2B).}
\label{tab:hyper}
\footnotesize
\begin{tabular}{@{}>{\raggedright\arraybackslash}p{0.42\linewidth}@{}>{\raggedright\arraybackslash}p{0.52\linewidth}@{}}
\toprule
Item & Value \\
\midrule
Shared $\pi_\theta$ & \texttt{Qwen2.5-VL-3B} / \texttt{Qwen3-VL-2B} \\
Analyzer / Teacher & \texttt{policy\_vllm}; panel $P_{\hat t}$ \\
GRPO $N$, clip $\varepsilon$ & $8$; $0.2$ \\
$\gamma$; actor LR & $0.95$; $1{\times}10^{-6}$ \\
KL coeff. & \texttt{low\_var\_kl}, $0.01$ \\
$\lambda_{\mathrm{opd}}$ / $\beta_{\mathrm{opd}}$ & $0.01$ / $5$ \\
OPD mask & failed traj., all valid tokens \\
Invalid-action pen. & $0.1$ \\
Temps (train / val) & $1.0$ / $0.4$ \\
History; max response & $2$; $512$ \\
Batch / val $n$ & $16$ / $128$ (Nav.\ $8$) \\
Updates; GPUs & $250$ (Sok.\ $350$); $8$ \\
\bottomrule
\end{tabular}
\end{minipage}}%
\noindent\usebox{\apptabenvsbox}\hfill\usebox{\apptabhyperbox}

\medskip
\noindent\usebox{\apptabrewardsbox}\hfill\usebox{\apptabsftbox}
\end{table}

\appblock{Benchmarks.}
We evaluate five tasks from VAGEN~\citep{wang2025vagen} (Tab.~\ref{tab:envs}), grouped into the three families of Tab.~\ref{tab:main}.

\paragraph{Sokoban~\citep{schrader2018sokoban}.}
In this classic puzzle, the agent must push all boxes onto target locations.
The visual state is a $6{\times}6$ grid ($T_{\max}{=}9$), and the action space is discrete (up, down, left, right).
Validation reports success rate.

\paragraph{FrozenLake~\citep{towers2026gymnasium}.}
The agent navigates a $4{\times}4$ grid to reach a goal while avoiding holes ($T_{\max}{=}9$).
The visual state and discrete action space match Sokoban; we disable the slippery setting for determinism.
Validation reports success rate.

\paragraph{Navigation~\citep{kolve2017ai2,yang2025embodiedbench}.}
A 3D embodied task ($T_{\max}{=}25$): the agent follows instructions to find an object from a first-person view with discrete actions (e.g., \texttt{moveahead}).
We average success rate on the Base and Common-sense splits.

\paragraph{PrimitiveSkill~\citep{tao2025maniskill3,hiranaka2023primitive}.}
The agent controls a Panda arm to perform tabletop manipulation ($T_{\max}{=}6$).
Actions use a hybrid space (e.g., \texttt{pick($x,y,z$)}): the policy must ground objects in a third-person 3D scene to coordinates.
Validation reports mean success rate over four skills.

\paragraph{SVG reconstruction~\citep{rodriguez2025starvector}.}
The agent generates SVG code that replicates a target image ($T_{\max}{=}2$).
The action space is open-ended text: each step submits a full canvas.
Validation averages DINO and DreamSim on the final submission.

\appblock{Rewards and task specs.}
Tables~\ref{tab:envs}--\ref{tab:hyper} collect per-task horizons and validation metrics, the episode-return structure used in GRPO (logged as \texttt{final\_reward} with discount $\gamma{=}0.95$), Analyzer SFT split sizes, and Stage~II training settings.
Invalid actions receive an additional $0.1$ penalty on top of the domain reward.
For OPD masking we mark a rollout as failed when its logged return falls below the success threshold $R_{\mathrm{succ}}{=}1.0$ whenever the environment does not expose a separate success bit.

\section{Training Objectives}
\label{app:grpo}
\label{app:sft-loss}
\label{app:opd}

The main text defines group advantages (Eq.~\eqref{eq:grpo-adv}), the GRPO surrogate (Eq.~\eqref{eq:grpo-loss}), gated OPD (Eq.~\eqref{eq:opd-main}), and their sum (Eq.~\eqref{eq:joint-objective}).
Below we give the Stage~I Analyzer loss explicitly and expand Stage~II objectives to token indices, including the KL regularizer omitted from the main display.

\appblock{Analyzer SFT.}
Stage~I (Sec.~\ref{sec:sft}) trains the shared checkpoint to read a failed rollout without environment access: input $x_\tau=(q,C(\tau),\acttxt_{0:T-1})$ matches Eq.~\eqref{eq:analyzer}, and the label is the solver-supervised triplet $z_\tau=(t^\ast,m^\ast,r^\ast)$.
Pivot step $t^\ast$ and mode $m^\ast$ are pinned offline as in App.~\ref{app:certs}; optional guidance text $r^\ast$ is produced by a frozen teacher on records that pass the loose factual filter (App.~\ref{app:hyper}).
Let $z_{\tau,1:L_\tau}$ denote the tokenized JSON string for that object.
We minimize length-normalized autoregressive negative log-likelihood,
\begin{equation}
\mathcal{L}_{\mathrm{sft}}(\theta)
=
-\mathbb{E}_{(x_\tau,z_\tau)\sim\mathcal{D}_{\mathrm{sft}}}
\left[
\frac{1}{L_\tau}
\sum_{\ell=1}^{L_\tau}
\log\pi_\theta
\bigl(z_{\tau,\ell}\mid x_\tau,z_{\tau,<\ell}\bigr)
\right].
\label{eq:sft-loss}
\end{equation}
The resulting weights $\theta_{\mathrm{sft}}$ initialize Stage~II and supply the Analyzer used at training time; at inference only the unprivileged Student is deployed (Sec.~\ref{sec:sft}).


\appblock{Confidence-gated OPD.}
Sec.~\ref{sec:loss} uses the same $\pi_\theta$ as a detached Teacher: privileged log-probs $\ell^{\mathrm{tea}}_{q,n,t,\ell}$ condition on $\tilde h_t=(p,h_t)$ from Eq.~\eqref{eq:teacher}, while $\ell^{\mathrm{stu}}_{q,n,t,\ell}$ uses the unprivileged history $h_t$ only.
Rollout actions are not resampled; both branches re-score the tokens already taken on the failed trajectory.
For each failed rollout we define the hindsight gap and gate
\begin{equation}
\Delta_{q,n,t,\ell}
=
\operatorname{sg}\!\left[
\ell^{\mathrm{tea}}_{q,n,t,\ell}
-
\ell^{\mathrm{stu}}_{q,n,t,\ell}
\right],
\qquad
g_{q,n,t,\ell}
=
\sigma\!\left(\beta_{\mathrm{opd}}\Delta_{q,n,t,\ell}\right),
\label{eq:opd-gate}
\end{equation}
matching the main-text $\delta_t$ and $g_t$ in Eq.~\eqref{eq:opd-main} ($\lambda_{\mathrm{opd}}$ and $\beta_{\mathrm{opd}}$ in Tab.~\ref{tab:hyper}).
The failed-trajectory loss aggregates over all valid action tokens on $R(\tau_q^{(n)})<R_{\mathrm{succ}}$:
\begin{equation}
\mathcal{L}_{\mathrm{opd}}(\theta)
=
\mathbb{E}_{q,n,t,\ell}
\left[
\mathbb{I}\!\left[R(\tau_q^{(n)})<R_{\mathrm{succ}}\right]
m_{\mathrm{act},q,n,t,\ell}\,
g_{q,n,t,\ell}
\left(
\operatorname{sg}
\bigl[\ell^{\mathrm{tea}}_{q,n,t,\ell}\bigr]
-
\ell^{\mathrm{stu}}_{q,n,t,\ell}
\right)
\right].
\label{eq:opd-loss}
\end{equation}
Stop-gradient on Teacher logits and on $\Delta$ leaves the update on Student parameters:
\begin{equation}
\nabla_\theta\mathcal{L}_{\mathrm{opd}}
=
-\mathbb{E}_{q,n,t,\ell}
\left[
\mathbb{I}\!\left[R(\tau_q^{(n)})<R_{\mathrm{succ}}\right]
m_{\mathrm{act},q,n,t,\ell}\,
g_{q,n,t,\ell}\,
\nabla_\theta\ell^{\mathrm{stu}}_{q,n,t,\ell}
\right].
\label{eq:opd-gradient}
\end{equation}
When a GRPO group has $\sigma_q{=}0$, $A^{\mathrm{rl}}_{q,n}$ vanishes for every token but this term can still train the Student on failed rollouts (Sec.~\ref{sec:loss}).

\appblock{SFT and RL optimization.}
Analyzer SFT optimizes Eq.~\eqref{eq:sft-loss} with AdamW at learning rate $2{\times}10^{-6}$ (batch size $8$, maximum sequence length $8192$, three epochs per environment).
Each environment yields its own $\theta_{\mathrm{sft}}$ checkpoint, which initializes Stage~II for that task only.
All Stage~II knobs not listed here (GRPO group size, OPD weights, KL coefficient, batching, and validation counts) match Tab.~\ref{tab:hyper}.

\section{SFT Data and Optimization}
\label{app:hyper}

\appblock{Loose factual filter for Analyzer SFT.}
\label{app:cf}
Sec.~\ref{sec:sft} trains on accepted failures only; each example must pass a loose factual filter $\nu_\tau$ after the solver pins $(t^\ast,m^\ast)$ (App.~\ref{app:certs}).
We require parseable Analyzer JSON, $m^\ast\in\mathcal{M}_{\mathrm{fail}}$, and consistency between the solver replay and the logged trajectory.
When a guidance field $r^\ast$ is present, it must be non-empty and refer to the action at $t^\ast$.
Task-specific checks further drop bad labels: Sokoban \texttt{deadlock} unless the replay matches the pinned state; FrozenLake skills that contradict hole-fall or budget markers; Navigation skills that mention an unreached goal.
Any failed check removes the record; we do not apply lift-based replay filtering when building $\mathcal{D}_{\mathrm{sft}}$.

\appblock{Analyzer SFT data.}
Data come from failed rollouts of the raw instruct checkpoint at sampling temperature $1.0$ (one dataset per environment).
The environment solver supplies $(t^\ast,m^\ast)$; a frozen \texttt{Qwen3-VL-8B} model writes optional $r^\ast$ on records that survive $\nu_\tau$.
Tab.~\ref{tab:sft-n} lists the resulting $90/10$ train/validation counts for \texttt{Qwen2.5-VL-3B} packs (\texttt{Qwen3-VL-2B} follows the same pipeline).
Example collages, role prompts, and JSON targets appear in App.~\ref{app:examples}; AdamW and Stage~II settings are in Tab.~\ref{tab:hyper}.

\section{Analyzer Collages, Prompts, and SFT Targets}
\label{app:examples}

This section documents training-time prompts and the remaining Analyzer supervision visuals. Representative FrozenLake, Navigation, and PrimitiveSkill pairs are shown in Fig.~\ref{fig:sft-examples}.
Prompt templates appear first, followed by multi-page collage figures.

\appblock{Role prompts by environment.}
For each VAGEN task we include three prompt families: the unprivileged Student (task instruction plus short RGB history), the Analyzer (full collage $C(\tau)$ and action transcript $\acttxt_{0:T-1}$, JSON output), and the privileged Teacher under \armMP{} or \armMPR{} (panel $P_{\hat t}$ plus mode and optional skill text; see Fig.~\ref{fig:panels}).
Vision placeholders use \texttt{\textless{}image\textgreater{}}; closed failure-mode lists follow App.~\ref{app:certs}.
Ground-truth SFT targets $(t^\ast,m^\ast,r^\ast)$ are built as in App.~\ref{app:hyper}.

\appenvheading{Sokoban}
\begin{promptbox}{Sokoban --- Student ($\pi_\theta$, RL rollout)}
You are an expert agent operating in the Sokoban environment. \allowbreak Your goal is to push all the boxes onto the target spots. \allowbreak Once all boxes are on the targets, \allowbreak you win!

\# Rules
You can only push boxes. \allowbreak You can't pull them, \allowbreak so plan ahead to avoid getting stuck.
You can't walk through or push boxes into walls.
To avoid traps, \allowbreak do not push boxes into corners or against walls where they can't be moved again.

\# Visual Elements in the Image:
Character: A small, \allowbreak green alien-like figure with two antennae and black eyes. \allowbreak It represents you.
Box: A yellow crate marked with an orange "X" across its front. \allowbreak It is the box you need to push.
Target: A black tile outlined in red, \allowbreak with a small red diamond shape in the center. \allowbreak It marks the destination where a box should be pushed.

\# Current Step
Your current observation is shown in the image: \textless{}image\textgreater{}
Your admissible actions are ["up", \allowbreak "down", \allowbreak "left", \allowbreak "right"].

Now it's your turn to make a move (choose ONE action only for the current step).
You should first reason step-by-step about the current situation — observe the positions of boxes and targets, \allowbreak plan a path to push a box toward a target, \allowbreak and avoid traps like corners or walls. \allowbreak This reasoning process MUST be enclosed within \textless{}redacted\_thinking\textgreater{} \textless{}/redacted\_thinking\textgreater{} tags. \allowbreak 
Once you've finished your reasoning, \allowbreak you should choose an admissible action for current step and present it within \textless{}action\textgreater{} \textless{}/action\textgreater{} tags.
\end{promptbox}
\begin{promptbox}{Sokoban --- Analyzer (failed rollout, SFT/online)}
Screenshot collage (\textless{}image\textgreater{}): 3x3 grid in row-major order, \allowbreak each panel labeled by original 0-based step index (0..8). \allowbreak Match the panel label with that step in the trajectory text below.

Analyze this failed agent episode from the screenshot collage and the compact action log. \allowbreak Return ONLY valid JSON.

You need to complete three fields:
1. \allowbreak pivot\_step: the 0-based index of the first step after which the remaining step budget cannot solve the puzzle.
2. \allowbreak failure\_mode: "timeout" if the puzzle is still solvable with more steps, \allowbreak or "deadlock" if the box is irreversibly stuck.
3. \allowbreak failure\_reason: extract the failed trajectory into avoidance rules (the core mistake and warning signs), \allowbreak then one short imperative sentence the policy can act on at pivot\_step, \allowbreak not a retrospective explanation. \allowbreak Put both parts in this one string.

Important constraints:
- Use the screenshot collage together with the compact action log;\allowbreak  do not ignore the images.
- Step indices are 0-based integers.
- pivot\_step MUST be an integer in [0, \allowbreak 8] matching a labeled collage panel.
- Do not copy an example index;\allowbreak  pick the true fatal step for THIS trajectory.
- Prefer a valid box-push that made the remaining budget insufficient. \allowbreak Invalid / no-op actions (action=invalid or moved=false) are usually not the pivot.
- failure\_mode must be exactly timeout or deadlock.
- Return only these top-level fields: pivot\_step, \allowbreak failure\_mode, \allowbreak failure\_reason.

Return format:
\{
  "pivot\_step": \textless{}integer\textgreater{},
  "failure\_mode": "timeout",
  "failure\_reason": "Avoid repeating the core mistake. \allowbreak At the pivot, \allowbreak take the productive action instead of that compact-log action."
\}

Episode context:
- Task description: Push the box onto the target without trapping it against a wall or corner.
- episode\_success: failure
- Number of steps: 9
- Valid pivot\_step range: [0, \allowbreak 8]
- Compact action log:
step 0: action=left moved=true
step 1: action=right moved=true
step 2: action=right moved=false
step 3: action=left moved=true
step 4: action=left moved=true
step 5: action=up moved=false
step 6: action=right moved=true
step 7: action=left moved=true
step 8: action=down moved=true

\end{promptbox}
\begin{promptbox}{Sokoban --- Teacher (\armMPR\{\}, privileged scoring)}
You are an expert agent operating in the Sokoban environment. \allowbreak Your goal is to push all the boxes onto the target spots. \allowbreak Once all boxes are on the targets, \allowbreak you win!

\# Rules
You can only push boxes. \allowbreak You can't pull them, \allowbreak so plan ahead to avoid getting stuck.
You can't walk through or push boxes into walls.
To avoid traps, \allowbreak do not push boxes into corners or against walls where they can't be moved again.

\# Visual Elements in the Image:
Character: A small, \allowbreak green alien-like figure with two antennae and black eyes. \allowbreak It represents you.
Box: A yellow crate marked with an orange "X" across its front. \allowbreak It is the box you need to push.
Target: A black tile outlined in red, \allowbreak with a small red diamond shape in the center. \allowbreak It marks the destination where a box should be pushed.

\# Current Step
Your current observation is shown in the image: \textless{}image\textgreater{}

Hindsight 3-panel around the predicted failure pivot (not available when selecting the current action):
\textless{}image\textgreater{}

Your admissible actions are ["up", \allowbreak "down", \allowbreak "left", \allowbreak "right"].

**Episode-Level Skill**
Refer to this episode-level skill when deciding what action to take in the current episode:
[Failure mode: deadlock
Refer to this episode-level skill when deciding what action to take: Avoid pushing the box into a corner where it becomes trapped. \allowbreak At step 4, \allowbreak do not push up again;\allowbreak  instead, \allowbreak move right to create space.]

Now it's your turn to make a move (choose ONE action only for the current step).
You should first reason step-by-step about the current situation — observe the positions of boxes and targets, \allowbreak plan a path to push a box toward a target, \allowbreak and avoid traps like corners or walls. \allowbreak This reasoning process MUST be enclosed within \textless{}redacted\_thinking\textgreater{} \textless{}/redacted\_thinking\textgreater{} tags. \allowbreak 
Once you've finished your reasoning, \allowbreak you should choose an admissible action for current step and present it within \textless{}action\textgreater{} \textless{}/action\textgreater{} tags.
\end{promptbox}
\vspace{0.35em}
\appenvheading{FrozenLake}
\begin{promptbox}{FrozenLake --- Student ($\pi_\theta$, RL rollout)}
You are an expert agent operating in the FrozenLake environment. \allowbreak Reach the goal (G) without falling into holes (O).

\# Rules
Move carefully on the ice. \allowbreak Holes end the episode. \allowbreak The goal tile wins.

\# Current Step
Your current observation is shown in the image: \textless{}image\textgreater{}
Your admissible actions are ["up", \allowbreak "down", \allowbreak "left", \allowbreak "right"].

Now it's your turn to make a move (choose ONE action only for the current step).
You should first reason step-by-step about the current situation. \allowbreak This reasoning process MUST be enclosed within \textless{}redacted\_thinking\textgreater{} \textless{}/redacted\_thinking\textgreater{} tags.
Once you've finished your reasoning, \allowbreak you should choose an admissible action for current step and present it within \textless{}action\textgreater{} \textless{}/action\textgreater{} tags.
\end{promptbox}
\begin{promptbox}{FrozenLake --- Analyzer (failed rollout, SFT/online)}
Screenshot collage (\textless{}image\textgreater{}): 2x2 grid in row-major order, \allowbreak each panel labeled by original 0-based step index (0..2). \allowbreak Match the panel label with that step in the trajectory text below.

Analyze this failed agent episode from the screenshot collage and the compact action log. \allowbreak Return ONLY valid JSON.

You need to complete three fields:
1. \allowbreak pivot\_step: the 0-based index of the first step after which the remaining step budget cannot reach the goal.
2. \allowbreak failure\_mode: "timeout" if the agent is still on safe ice and more steps could reach the goal, \allowbreak or "deadlock" if the agent fell into a hole (irreversible).
3. \allowbreak failure\_reason: extract the failed trajectory into avoidance rules (the core mistake and warning signs), \allowbreak then one short imperative sentence the policy can act on at pivot\_step, \allowbreak not a retrospective explanation. \allowbreak Put both parts in this one string.

Important constraints:
- Use the screenshot collage together with the compact action log;\allowbreak  do not ignore the images.
- Step indices are 0-based integers.
- pivot\_step MUST be an integer in [0, \allowbreak 2] matching a labeled collage panel.
- Do not copy an example index;\allowbreak  pick the true fatal step for THIS trajectory.
- Prefer the step that walked into a hole, \allowbreak or a valid move that made the remaining budget insufficient. \allowbreak Invalid / empty / no-op actions (action=invalid or moved=false) are usually not the pivot, \allowbreak unless the episode ends on that empty action.
- failure\_mode must be exactly timeout or deadlock.
- Return only these top-level fields: pivot\_step, \allowbreak failure\_mode, \allowbreak failure\_reason.

Return format:
\{
  "pivot\_step": \textless{}integer\textgreater{},
  "failure\_mode": "timeout",
  "failure\_reason": "Avoid repeating the core mistake. \allowbreak At the pivot, \allowbreak take the productive action instead of that compact-log action."
\}

Episode context:
- Task description: Navigate from the start to the goal on the frozen lake without falling into holes.
- episode\_success: failure
- Number of steps: 3
- Valid pivot\_step range: [0, \allowbreak 2]
- Compact action log:
step 0: action=left moved=true remaining\_turns=8
step 1: action=right moved=true remaining\_turns=7
step 2: action=right moved=true remaining\_turns=6

\end{promptbox}
\begin{promptbox}{FrozenLake --- Teacher (\armMPR\{\}, privileged scoring)}
You are an expert agent operating in the FrozenLake environment. \allowbreak Reach the goal (G) without falling into holes (O).

\# Rules
Move carefully on the ice. \allowbreak Holes end the episode. \allowbreak The goal tile wins.

\# Current Step
Your current observation is shown in the image: \textless{}image\textgreater{}

Hindsight 3-panel around the predicted failure pivot (not available when selecting the current action):
\textless{}image\textgreater{}

Your admissible actions are ["up", \allowbreak "down", \allowbreak "left", \allowbreak "right"].

**Episode-Level Skill**
Refer to this episode-level skill when deciding what action to take in the current episode:
[Failure mode: deadlock
Refer to this episode-level skill when deciding what action to take: Avoid stepping onto hole tiles when a safe path exists. \allowbreak At step 1, \allowbreak move right along safe ice instead of up into the hole.]

Now it's your turn to make a move (choose ONE action only for the current step).
You should first reason step-by-step about the current situation. \allowbreak This reasoning process MUST be enclosed within \textless{}redacted\_thinking\textgreater{} \textless{}/redacted\_thinking\textgreater{} tags.
Once you've finished your reasoning, \allowbreak you should choose an admissible action for current step and present it within \textless{}action\textgreater{} \textless{}/action\textgreater{} tags.
\end{promptbox}
\vspace{0.35em}
\appenvheading{Navigation}
\begin{promptbox}{Navigation --- Student ($\pi_\theta$, RL rollout)}
You are an embodied agent navigating an AI2-THOR indoor environment from egocentric RGB.

\# Instruction
navigate to the Bread in the room and be as close as possible to it

\# Actions (choose ONE)
- moveahead / moveback / moveright / moveleft: translate 0.5m
- rotateright / rotateleft: yaw 90 degrees
- lookup / lookdown: pitch 30 degrees

Prior steps: 3. \allowbreak Recent actions (2): step 2: moveahead;\allowbreak  step 3: rotateright
Now at step 4. \allowbreak Current egocentric view: \textless{}image\textgreater{}

Reason in \textless{}redacted\_thinking\textgreater{} \textless{}/redacted\_thinking\textgreater{}, \allowbreak then put ONE action in \textless{}action\textgreater{} \textless{}/action\textgreater{}.
\end{promptbox}
\begin{promptbox}{Navigation --- Analyzer (failed rollout, SFT/online)}
Screenshot collage (\textless{}image\textgreater{}): 5x5 grid in row-major order, \allowbreak each panel labeled by original 0-based step index (0..24). \allowbreak Match the panel label with that step in the trajectory text below.

Analyze this failed agent episode from the screenshot collage and the compact action log. \allowbreak Return ONLY valid JSON.

You need to complete three fields:
1. \allowbreak pivot\_step: the 0-based index of the earliest structural mistake — a terminal blocked translation, \allowbreak the start of a no-progress loop, \allowbreak a trailing invalid suffix, \allowbreak or else the first step whose remaining translations cannot cover the Euclidean distance.
2. \allowbreak failure\_mode: exactly one of "budget", \allowbreak "format", \allowbreak "blocked", \allowbreak "stagnation". \allowbreak budget = remaining translations cannot cover remaining distance. \allowbreak format = empty/invalid action exhausted slack. \allowbreak blocked = a move was rejected by a wall/door. \allowbreak stagnation = recent moves did not reduce distance.
3. \allowbreak failure\_reason: extract the failed trajectory into avoidance rules (the core mistake and warning signs), \allowbreak then one short imperative sentence the policy can act on at pivot\_step, \allowbreak not a retrospective explanation. \allowbreak Put both parts in this one string.

Important constraints:
- Use the screenshot collage together with the compact action log;\allowbreak  do not ignore the images.
- Step indices are 0-based integers.
- pivot\_step MUST be an integer in [0, \allowbreak 24] matching a labeled collage panel.
- Do not copy an example index;\allowbreak  pick the true fatal step for THIS trajectory.
- Prefer the earliest real navigation action that matches that mode. \allowbreak Rotate/look are valid and may be a stagnation pivot. \allowbreak Empty/invalid is the format pivot when it is a trailing suffix.
- failure\_mode must be exactly one of: budget, \allowbreak format, \allowbreak blocked, \allowbreak stagnation.
- Return only these top-level fields: pivot\_step, \allowbreak failure\_mode, \allowbreak failure\_reason.

Return format:
\{
  "pivot\_step": \textless{}integer\textgreater{},
  "failure\_mode": "budget",
  "failure\_reason": "Avoid repeating the core mistake. \allowbreak At the pivot, \allowbreak take the productive action instead of that compact-log action."
\}

Episode context:
- Task description: navigate to the Bread in the room and be as close as possible to it
- episode\_success: failure
- Number of steps: 25
- Valid pivot\_step range: [0, \allowbreak 24]
- Compact action log:
step 0: action=moveright moved=false last\_action\_success=false action\_is\_valid=true dist=2.87 remaining\_turns=24
step 1: action=moveright moved=false last\_action\_success=false action\_is\_valid=false dist=2.87 remaining\_turns=23
step 2: action=moveahead moved=true last\_action\_success=true action\_is\_valid=true dist=2.39 remaining\_turns=22
step 3: action=moveleft moved=true last\_action\_success=true action\_is\_valid=true dist=2.56 remaining\_turns=21
step 4: action=rotateright moved=false last\_action\_success=false action\_is\_valid=false dist=2.56 remaining\_turns=20
step 5: action=moveahead moved=false last\_action\_success=false action\_is\_valid=false dist=2.39 remaining\_turns=19
step 6: action=moveright moved=true last\_action\_success=true action\_is\_valid=true dist=2.87 remaining\_turns=18
step 7: action=moveright moved=true last\_action\_success=true action\_is\_valid=true dist=3.36 remaining\_turns=17
step 8: action=moveleft moved=true last\_action\_success=true action\_is\_valid=true dist=2.87 remaining\_turns=16
step 9: action=rotateleft moved=false last\_action\_success=true action\_is\_valid=true dist=2.87 remaining\_turns=15
step 10: action=moveleft moved=false last\_action\_success=false action\_is\_valid=false dist=3.02 remaining\_turns=14
step 11: action=moveahead moved=true last\_action\_success=true action\_is\_valid=true dist=2.56 remaining\_turns=13
step 12: action=invalid moved=false last\_action\_success=false action\_is\_valid=true remaining\_turns=12
step 13: action=invalid moved=false last\_action\_success=false action\_is\_valid=true remaining\_turns=11
step 14: action=invalid moved=false last\_action\_success=false action\_is\_valid=true remaining\_turns=10
step 15: action=moveleft moved=true last\_action\_success=true action\_is\_valid=true dist=2.82 remaining\_turns=9
step 16: action=invalid moved=false last\_action\_success=false action\_is\_valid=false remaining\_turns=8
step 17: action=invalid moved=false last\_action\_success=false action\_is\_valid=false remaining\_turns=7
step 18: action=moveleft moved=false last\_action\_success=false action\_is\_valid=true dist=2.82 remaining\_turns=6
step 19: action=moveleft moved=false last\_action\_success=false action\_is\_valid=true dist=2.82 remaining\_turns=5
step 20: action=invalid moved=false last\_action\_success=false action\_is\_valid=false remaining\_turns=4
step 21: action=moveahead moved=false last\_action\_success=false action\_is\_valid=true dist=2.82 remaining\_turns=3
step 22: action=rotateright moved=false last\_action\_success=true action\_is\_valid=true dist=2.82 remaining\_turns=2
step 23: action=rotateright moved=false last\_action\_success=true action\_is\_valid=true dist=2.82 remaining\_turns=1
step 24: action=moveahead moved=false last\_action\_success=false action\_is\_valid=true dist=2.82 remaining\_turns=0

\end{promptbox}
\begin{promptbox}{Navigation --- Teacher (\armMPR\{\}, privileged scoring)}
You are an embodied agent navigating an AI2-THOR indoor environment from egocentric RGB.

\# Instruction
navigate to the Bread in the room and be as close as possible to it

\# Actions (choose ONE)
- moveahead / moveback / moveright / moveleft: translate 0.5m
- rotateright / rotateleft: yaw 90 degrees
- lookup / lookdown: pitch 30 degrees

Prior steps: 3. \allowbreak Recent actions (2): step 2: moveahead;\allowbreak  step 3: rotateright
Now at step 4. \allowbreak Current egocentric view: \textless{}image\textgreater{}

Reason in \textless{}redacted\_thinking\textgreater{} \textless{}/redacted\_thinking\textgreater{}, \allowbreak then put ONE action in \textless{}action\textgreater{} \textless{}/action\textgreater{}.

**Episode-Level Skill**
Refer to this episode-level skill when deciding what action to take in the current episode:
[Failure mode: blocked
Refer to this episode-level skill when deciding what action to take: Avoid repeated moveahead into a closed door. \allowbreak At step 22, \allowbreak rotate or sidestep before advancing again.]

Hindsight 3-panel around the predicted failure pivot (not available when selecting the current action):
\textless{}image\textgreater{}
\end{promptbox}
\vspace{0.35em}
\appenvheading{PrimitiveSkill}
\begin{promptbox}{PrimitiveSkill --- Student ($\pi_\theta$, RL rollout)}
You are an AI assistant controlling a Franka Emika robot arm. \allowbreak Your goal is to understand human instructions and translate them into a sequence of executable actions for the robot, \allowbreak based on visual input and the instruction.

Action Space Guide
You can command the robot using the following actions:

1. \allowbreak pick(x, \allowbreak y, \allowbreak z) \# To grasp an object located at position(x,y,z) in the robot's workspace.
2. \allowbreak place(x, \allowbreak y, \allowbreak z) \# To place the object currently held by the robot's gripper at the target position (x,y,z).
3. \allowbreak push(x1, \allowbreak y1, \allowbreak z1, \allowbreak x2, \allowbreak y2, \allowbreak z2) \# To push an object from position (x1,y1,z1) to (x2,y2,z2).

Hints: 
1. \allowbreak The coordinates (x, \allowbreak y, \allowbreak z) are in millimeters and are all integers.
2. \allowbreak Please ensure that the coordinates are within the workspace limits.
3. \allowbreak The position is the center of the object, \allowbreak when you place, \allowbreak please consider the volume of the object. \allowbreak It's always fine to set z much higher when placing an item.
4. \allowbreak We will provide the object positions to you, \allowbreak but you need to match them to the object in the image by yourself. \allowbreak You're facing toward the negative x-axis, \allowbreak and the negative y-axis is to your left, \allowbreak the positive y-axis is to your right, \allowbreak and the positive z-axis is up. \allowbreak 

Examples:
round1:
image1
Human Instruction: Put red cube on green cube and yellow cube on left target
Object positions:
[(62,-55,20),(75,33,20),(-44,100,20),(100,-43,0),(100,43,0)]
Reasoning: I can see from the picture that the red cube is on my left and green cube is on my right and near me. \allowbreak 
Since I'm looking toward the negative x axis, \allowbreak and negative y-axis is to my left, \allowbreak (62,-55,20) would be the position of the red cube, \allowbreak (75,33,20) would be the position of the green cube and (-44,100,20) is the position of the yellow cube. \allowbreak 
Also the (100,-43,0) would be the position of the left target, \allowbreak and (100,43,0) would be the porition of the right target.
I need to pick up red cube first and place it on the green cube, \allowbreak when placing, \allowbreak I should set z much higher.
Anwer: pick(62,-55,20)|\allowbreak place(75,33,50)
round2:
image2
Human Instruction: Put red cube on green cube and yellow cube on left target
Object positions:
[(75,33,50),(75,33,20),(-44,100,20),(100,-43,0),(100,43,0)]
Reasoning: Now the red cube is on the green cube, \allowbreak so I need to pick up the yellow cube and place it on the left target.
Anwer: pick(-44,100,20)|\allowbreak place(100,-43,50)

You can take up to 2 action(s) at a time, \allowbreak separated by |\allowbreak .
You should first give your thought process, \allowbreak and then your answer.
Your response should be in the format of:
\textless{}redacted\_thinking\textgreater{}...\textless{}/redacted\_thinking\textgreater{}\textless{}answer\textgreater{}...\textless{}/answer\textgreater{}
e.g. \allowbreak \textless{}redacted\_thinking\textgreater{}I need to pick the red\_cube\_pos at (10,20,30) and place it on the green\_block\_pos at (50,60,40).\textless{}/redacted\_thinking\textgreater{}\textless{}answer\textgreater{}pick(10,20,30)|\allowbreak place(50,60,70)\textless{}/answer\textgreater{}

Prior steps: 3. \allowbreak Recent actions (2): step 2: pick(62,-55,20)|\allowbreak place(75,33,50);\allowbreak  step 3: pick(-44,100,20)
Now at step 4.

[Current Observation]:
\textless{}image\textgreater{}
Human Instruction: Please align the cubes in the y-axis, \allowbreak which means the x-coordinates of both cubes should be 0 (+-10mm)
x\_workspace\_limit: [-100, \allowbreak 100]
y\_workspace\_limit: [-100, \allowbreak 100]
z\_workspace\_limit: [0, \allowbreak 120]
Object positions: 
[(62,-55,20),(75,33,20),(-44,100,20),(100,-43,0),(100,43,0)]
Other information:
(none)
Decide your next action(s).
\end{promptbox}
\begin{promptbox}{PrimitiveSkill --- Analyzer (failed rollout, SFT/online)}
Screenshot collage (\textless{}image\textgreater{}): 3x2 grid in row-major order, \allowbreak each panel labeled by original 0-based step index (0..5). \allowbreak Match the panel label with that step in the trajectory text below.

Analyze this failed agent episode from the screenshot collage and the compact action log. \allowbreak Return ONLY valid JSON.

You need to complete three fields:
1. \allowbreak pivot\_step: the 0-based index of the first step after which the remaining turns cannot finish the remaining task stages.
2. \allowbreak failure\_mode: exactly one of "timeout", \allowbreak "format", \allowbreak "wrong\_target", \allowbreak "order\_error". \allowbreak timeout = remaining stages exceed remaining turns. \allowbreak format = no legal pick/place/push. \allowbreak wrong\_target = a legal action missed the object or target. \allowbreak order\_error = place/push before a required pick.
3. \allowbreak failure\_reason: extract the failed trajectory into avoidance rules (the core mistake and warning signs), \allowbreak then one short imperative sentence the policy can act on at pivot\_step, \allowbreak not a retrospective explanation. \allowbreak Put both parts in this one string.

Important constraints:
- Use the screenshot collage together with the compact action log;\allowbreak  do not ignore the images.
- Step indices are 0-based integers.
- pivot\_step MUST be an integer in [0, \allowbreak 5] matching a labeled collage panel.
- Do not copy an example index;\allowbreak  pick the true fatal step for THIS trajectory.
- Prefer a real pick/place/push that made the remaining stage budget insufficient. \allowbreak Empty/invalid is usually not the pivot unless that empty action exhausted the last slack.
- failure\_mode must be exactly one of: timeout, \allowbreak format, \allowbreak wrong\_target, \allowbreak order\_error.
- Return only these top-level fields: pivot\_step, \allowbreak failure\_mode, \allowbreak failure\_reason.

Return format:
\{
  "pivot\_step": \textless{}integer\textgreater{},
  "failure\_mode": "timeout",
  "failure\_reason": "Avoid repeating the core mistake. \allowbreak At the pivot, \allowbreak take the productive action instead of that compact-log action."
\}

Episode context:
- Task description: Please align the cubes in the y-axis, \allowbreak which means the x-coordinates of both cubes should be 0 (+-10mm)
- episode\_success: failure
- Number of steps: 6
- Valid pivot\_step range: [0, \allowbreak 5]
- Compact action log:
step 0: action=empty legal=false remaining\_stages=2 remaining\_turns=5
step 1: action=empty legal=false remaining\_stages=2 remaining\_turns=4
step 2: action=empty legal=false remaining\_stages=2 remaining\_turns=3
step 3: action=empty legal=false remaining\_stages=2 remaining\_turns=2
step 4: action=empty legal=false remaining\_stages=2 remaining\_turns=1
step 5: action=pick(107,64,20)|\allowbreak place(0,-185,120) legal=true remaining\_stages=1 remaining\_turns=0

\end{promptbox}
\begin{promptbox}{PrimitiveSkill --- Teacher (\armMPR\{\}, privileged scoring)}
You are an AI assistant controlling a Franka Emika robot arm. \allowbreak Your goal is to understand human instructions and translate them into a sequence of executable actions for the robot, \allowbreak based on visual input and the instruction.

Action Space Guide
You can command the robot using the following actions:

1. \allowbreak pick(x, \allowbreak y, \allowbreak z) \# To grasp an object located at position(x,y,z) in the robot's workspace.
2. \allowbreak place(x, \allowbreak y, \allowbreak z) \# To place the object currently held by the robot's gripper at the target position (x,y,z).
3. \allowbreak push(x1, \allowbreak y1, \allowbreak z1, \allowbreak x2, \allowbreak y2, \allowbreak z2) \# To push an object from position (x1,y1,z1) to (x2,y2,z2).

Hints: 
1. \allowbreak The coordinates (x, \allowbreak y, \allowbreak z) are in millimeters and are all integers.
2. \allowbreak Please ensure that the coordinates are within the workspace limits.
3. \allowbreak The position is the center of the object, \allowbreak when you place, \allowbreak please consider the volume of the object. \allowbreak It's always fine to set z much higher when placing an item.
4. \allowbreak We will provide the object positions to you, \allowbreak but you need to match them to the object in the image by yourself. \allowbreak You're facing toward the negative x-axis, \allowbreak and the negative y-axis is to your left, \allowbreak the positive y-axis is to your right, \allowbreak and the positive z-axis is up. \allowbreak 

Examples:
round1:
image1
Human Instruction: Put red cube on green cube and yellow cube on left target
Object positions:
[(62,-55,20),(75,33,20),(-44,100,20),(100,-43,0),(100,43,0)]
Reasoning: I can see from the picture that the red cube is on my left and green cube is on my right and near me. \allowbreak 
Since I'm looking toward the negative x axis, \allowbreak and negative y-axis is to my left, \allowbreak (62,-55,20) would be the position of the red cube, \allowbreak (75,33,20) would be the position of the green cube and (-44,100,20) is the position of the yellow cube. \allowbreak 
Also the (100,-43,0) would be the position of the left target, \allowbreak and (100,43,0) would be the porition of the right target.
I need to pick up red cube first and place it on the green cube, \allowbreak when placing, \allowbreak I should set z much higher.
Anwer: pick(62,-55,20)|\allowbreak place(75,33,50)
round2:
image2
Human Instruction: Put red cube on green cube and yellow cube on left target
Object positions:
[(75,33,50),(75,33,20),(-44,100,20),(100,-43,0),(100,43,0)]
Reasoning: Now the red cube is on the green cube, \allowbreak so I need to pick up the yellow cube and place it on the left target.
Anwer: pick(-44,100,20)|\allowbreak place(100,-43,50)

You can take up to 2 action(s) at a time, \allowbreak separated by |\allowbreak .
You should first give your thought process, \allowbreak and then your answer.
Your response should be in the format of:
\textless{}redacted\_thinking\textgreater{}...\textless{}/redacted\_thinking\textgreater{}\textless{}answer\textgreater{}...\textless{}/answer\textgreater{}
e.g. \allowbreak \textless{}redacted\_thinking\textgreater{}I need to pick the red\_cube\_pos at (10,20,30) and place it on the green\_block\_pos at (50,60,40).\textless{}/redacted\_thinking\textgreater{}\textless{}answer\textgreater{}pick(10,20,30)|\allowbreak place(50,60,70)\textless{}/answer\textgreater{}

Prior steps: 3. \allowbreak Recent actions (2): step 2: pick(62,-55,20)|\allowbreak place(75,33,50);\allowbreak  step 3: pick(-44,100,20)
Now at step 4.

[Current Observation]:
\textless{}image\textgreater{}
Human Instruction: Please align the cubes in the y-axis, \allowbreak which means the x-coordinates of both cubes should be 0 (+-10mm)
x\_workspace\_limit: [-100, \allowbreak 100]
y\_workspace\_limit: [-100, \allowbreak 100]
z\_workspace\_limit: [0, \allowbreak 120]
Object positions: 
[(62,-55,20),(75,33,20),(-44,100,20),(100,-43,0),(100,43,0)]
Other information:
(none)

**Episode-Level Skill**
Refer to this episode-level skill when deciding what action to take in the current episode:
[Failure mode: wrong\_target
Refer to this episode-level skill when deciding what action to take: Avoid placing on the wrong cube. \allowbreak At step 4, \allowbreak pick the yellow cube and place on the left target.]

Hindsight 3-panel around the predicted failure pivot (not available when selecting the current action):
\textless{}image\textgreater{}

Decide your next action(s).
\end{promptbox}
\vspace{0.35em}
\appenvheading{SVG Reconstruction}
\begin{promptbox}{SVG Reconstruction --- Student ($\pi_\theta$, RL rollout)}
You are a precise SVG code generator.

SVG Quick Guide
Goal: Transform the provided image into precise SVG code that replicates the image.

Process:
1. \allowbreak First analyze the image carefully, \allowbreak identifying distinct visual elements
2. \allowbreak Identify colors, \allowbreak dimensions, \allowbreak positions, \allowbreak and relationships between elements
3. \allowbreak Generate accurate SVG code that reproduces the image;\allowbreak  you can use \textless{}path\textgreater{} for better shape

Rewards:
- Overall visual similarity: +5.0
- Structural accuracy: +10.0

Example:
\textless{}redacted\_thinking\textgreater{}I can see the image contains a red circle and a blue rectangle. \allowbreak The circle is positioned at the top-left, \allowbreak while the rectangle is at the bottom-right.\textless{}/redacted\_thinking\textgreater{}
\textless{}answer\textgreater{}\textless{}svg viewBox="0 0 100 100" xmlns="http://www.w3.org/2000/svg"\textgreater{}
  \textless{}circle cx="25" cy="25" r="15" fill="red" /\textgreater{}
  \textless{}rect x="60" y="60" width="30" height="20" fill="blue" /\textgreater{}
\textless{}/svg\textgreater{}\textless{}/answer\textgreater{}

You can take up to 1 action(s) at a time, \allowbreak separated by ,.
You should first give your thought process, \allowbreak and then your answer.
Your response should be in the format of:
\textless{}redacted\_thinking\textgreater{}...\textless{}/redacted\_thinking\textgreater{}\textless{}answer\textgreater{}...\textless{}/answer\textgreater{}
e.g. \allowbreak \textless{}redacted\_thinking\textgreater{}I can see the image contains a red circle and a blue rectangle. \allowbreak The circle is positioned at the top-left, \allowbreak while the rectangle is at the bottom-right.\textless{}/redacted\_thinking\textgreater{}
\textless{}answer\textgreater{}\textless{}svg viewBox="0 0 100 100" xmlns="http://www.w3.org/2000/svg"\textgreater{}
  \textless{}circle cx="25" cy="25" r="15" fill="red" /\textgreater{}
  \textless{}rect x="60" y="60" width="30" height="20" fill="blue" /\textgreater{}
\textless{}/svg\textgreater{}\textless{}/answer\textgreater{}

[Initial Observation]:
Your target image is shown here: \textless{}image\textgreater{}
Please carefully observe the image, \allowbreak and generate SVG code that reproduces it as accurately as possible.
Decide on your SVG code.
\end{promptbox}
\begin{promptbox}{SVG Reconstruction --- Analyzer (failed rollout, SFT/online)}
Screenshot collage (\textless{}image\textgreater{}): 2x2 grid in row-major order. \allowbreak The panel labeled target is the ground-truth icon;\allowbreak  panels labeled step k are the rasterized SVG submitted at that 0-based step. \allowbreak Match step k with the compact action log.

Analyze this failed agent episode from the screenshot collage and the compact action log. \allowbreak Return ONLY valid JSON.

You need to complete three fields:
1. \allowbreak pivot\_step: the 0-based index of the terminating submission (the drawing that was scored, \allowbreak or the empty action that ended the episode).
2. \allowbreak failure\_mode: "timeout" if the task is still solvable but the remaining budget is not enough, \allowbreak or "deadlock" if the failure is irreversible. \allowbreak Prefer timeout;\allowbreak  pin the last submission (empty SVG stays action=empty in the log).
3. \allowbreak failure\_reason: extract the failed trajectory into avoidance rules (the core mistake and warning signs), \allowbreak then one short imperative sentence the policy can act on at pivot\_step, \allowbreak not a retrospective explanation. \allowbreak Put both parts in this one string.

Important constraints:
- Use the screenshot collage together with the compact action log;\allowbreak  do not ignore the images.
- Step indices are 0-based integers.
- pivot\_step MUST be an integer in [0, \allowbreak 1] matching a labeled collage panel.
- Do not copy an example index;\allowbreak  pick the true fatal step for THIS trajectory.
- Empty submissions are still a timeout pivot;\allowbreak  do not skip them. \allowbreak For a valid SVG, \allowbreak prefer the last submission.
- failure\_mode must be exactly timeout or deadlock.
- Return only these top-level fields: pivot\_step, \allowbreak failure\_mode, \allowbreak failure\_reason.

Return format:
\{
  "pivot\_step": \textless{}integer\textgreater{},
  "failure\_mode": "timeout",
  "failure\_reason": "Avoid repeating the core mistake. \allowbreak At the pivot, \allowbreak take the productive action instead of that compact-log action."
\}

Episode context:
- Task description: Reconstruct the target image as SVG code.
- episode\_success: failure
- Number of steps: 2
- Valid pivot\_step range: [0, \allowbreak 1]
- Compact action log:
step 0: action=svg valid\_svg=true remaining\_turns=1 average\_score=0.50
step 1: action=svg valid\_svg=true remaining\_turns=0 average\_score=0.50

\end{promptbox}
\begin{promptbox}{SVG Reconstruction --- Teacher (\armMPR\{\}, privileged scoring)}
You are a precise SVG code generator.

SVG Quick Guide
Goal: Transform the provided image into precise SVG code that replicates the image.

Process:
1. \allowbreak First analyze the image carefully, \allowbreak identifying distinct visual elements
2. \allowbreak Identify colors, \allowbreak dimensions, \allowbreak positions, \allowbreak and relationships between elements
3. \allowbreak Generate accurate SVG code that reproduces the image;\allowbreak  you can use \textless{}path\textgreater{} for better shape

Rewards:
- Overall visual similarity: +5.0
- Structural accuracy: +10.0

Example:
\textless{}redacted\_thinking\textgreater{}I can see the image contains a red circle and a blue rectangle. \allowbreak The circle is positioned at the top-left, \allowbreak while the rectangle is at the bottom-right.\textless{}/redacted\_thinking\textgreater{}
\textless{}answer\textgreater{}\textless{}svg viewBox="0 0 100 100" xmlns="http://www.w3.org/2000/svg"\textgreater{}
  \textless{}circle cx="25" cy="25" r="15" fill="red" /\textgreater{}
  \textless{}rect x="60" y="60" width="30" height="20" fill="blue" /\textgreater{}
\textless{}/svg\textgreater{}\textless{}/answer\textgreater{}

You can take up to 1 action(s) at a time, \allowbreak separated by ,.
You should first give your thought process, \allowbreak and then your answer.
Your response should be in the format of:
\textless{}redacted\_thinking\textgreater{}...\textless{}/redacted\_thinking\textgreater{}\textless{}answer\textgreater{}...\textless{}/answer\textgreater{}
e.g. \allowbreak \textless{}redacted\_thinking\textgreater{}I can see the image contains a red circle and a blue rectangle. \allowbreak The circle is positioned at the top-left, \allowbreak while the rectangle is at the bottom-right.\textless{}/redacted\_thinking\textgreater{}
\textless{}answer\textgreater{}\textless{}svg viewBox="0 0 100 100" xmlns="http://www.w3.org/2000/svg"\textgreater{}
  \textless{}circle cx="25" cy="25" r="15" fill="red" /\textgreater{}
  \textless{}rect x="60" y="60" width="30" height="20" fill="blue" /\textgreater{}
\textless{}/svg\textgreater{}\textless{}/answer\textgreater{}

[Initial Observation]:
Your target image is shown here: \textless{}image\textgreater{}
Please carefully observe the image, \allowbreak and generate SVG code that reproduces it as accurately as possible.

**Episode-Level Skill**
Refer to this episode-level skill when deciding what action to take in the current episode:
[Failure mode: timeout
Refer to this episode-level skill when deciding what action to take: Avoid omitting the main icon strokes. \allowbreak At step 1, \allowbreak redraw the sun disk and rays before submitting.]

Hindsight 3-panel around the predicted failure pivot (not available when selecting the current action):
\textless{}image\textgreater{}

Decide on your SVG code.
\end{promptbox}
\vspace{0.1em}

\begin{figure}[p]
\centering
\includegraphics[width=0.82\textwidth]{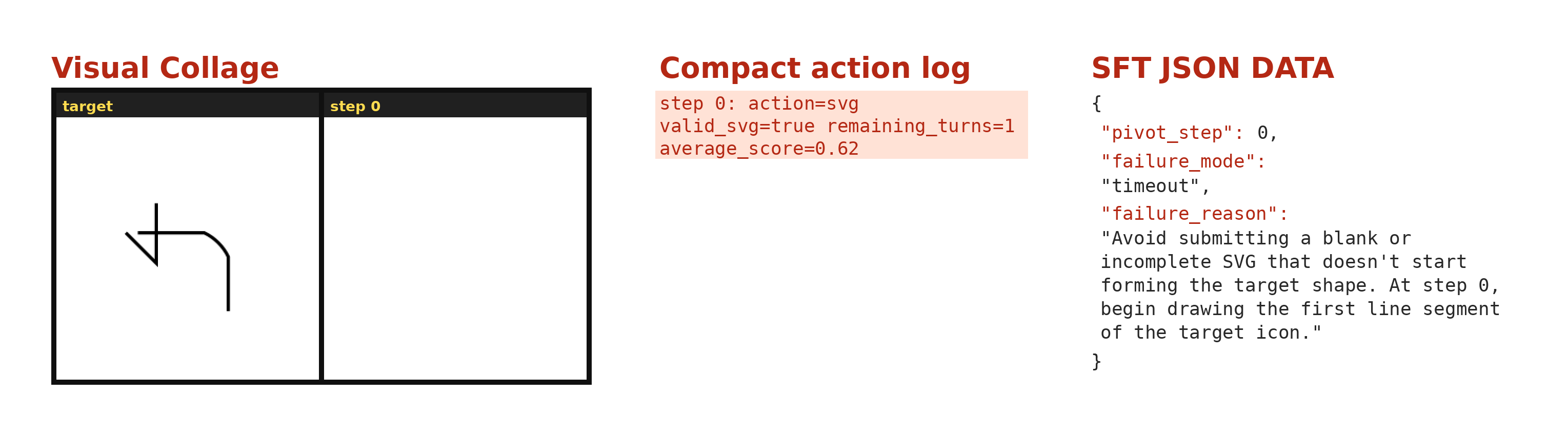}
\smallskip
\includegraphics[width=0.82\textwidth]{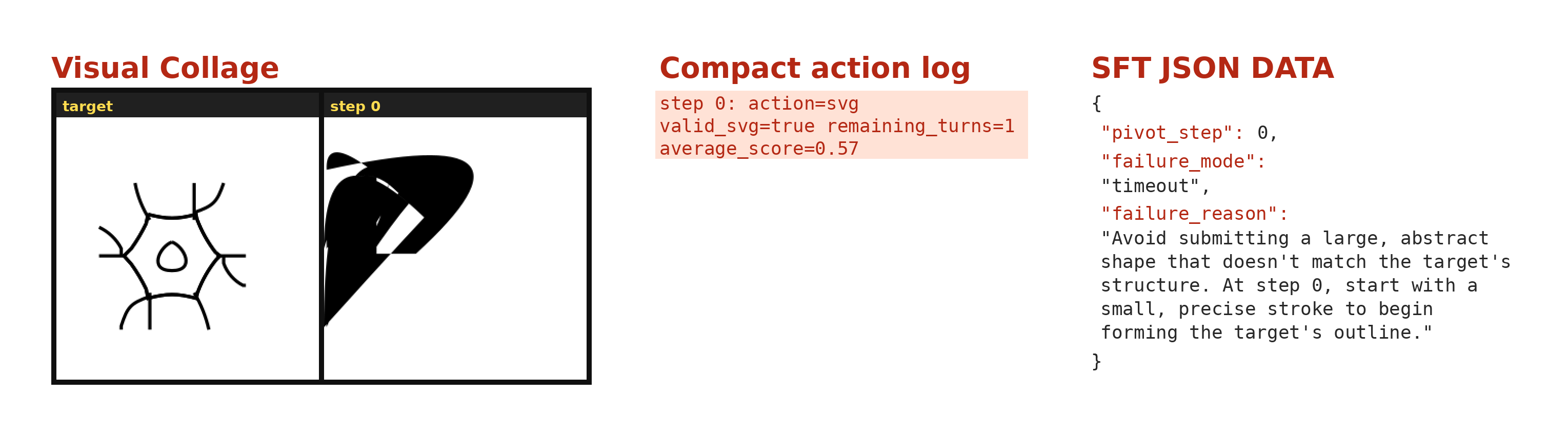}
\smallskip
\includegraphics[width=0.82\textwidth]{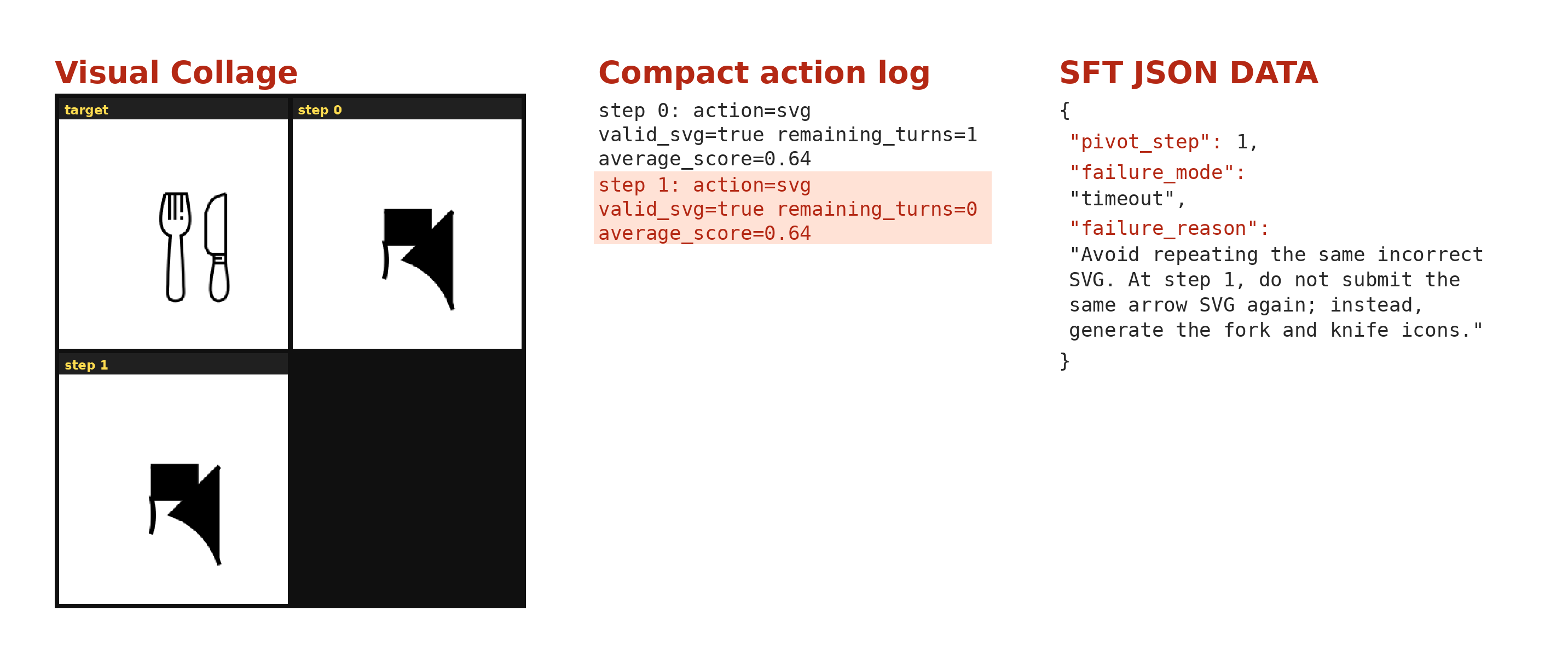}
\smallskip
\includegraphics[width=0.82\textwidth]{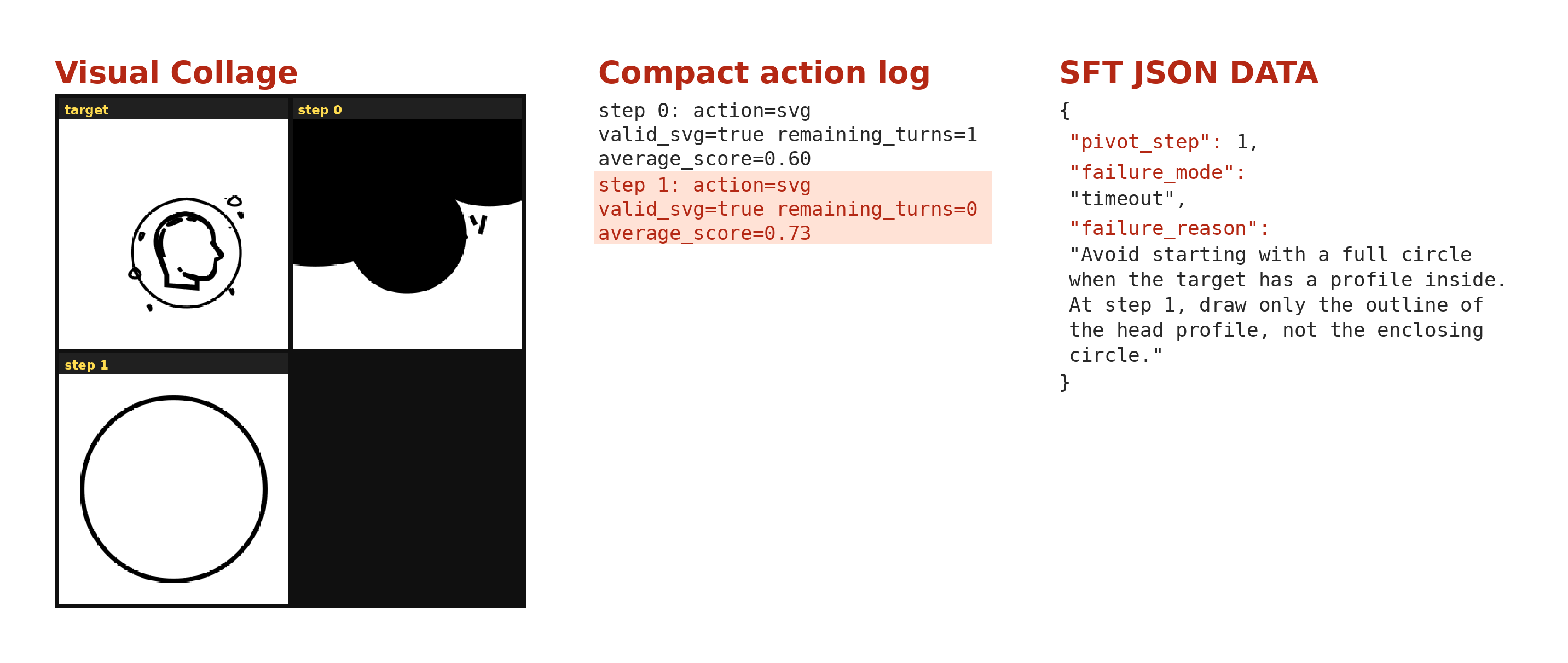}
\caption{SVG. SFT-stage input--output examples for this environment. Columns~1--2 are the visual trajectory collage and compact action log; column~3 is the JSON supervision target.}
\label{fig:app-collages}
\end{figure}

\enlargethispage{2.5\baselineskip}
\noindent
\appblock{Analyzer SFT collages.}
Figure~\ref{fig:app-collages} summarizes SFT-stage input--output examples across environments (continued on subsequent pages as noted).
Each figure is environment-specific: columns~1--2 are the visual trajectory collage and compact action log; column~3 is the JSON supervision target (Navigation logs truncate with ``\ldots'' around the pivot step).

\clearpage
\begin{figure}[p]
\centering
\includegraphics[width=0.82\textwidth]{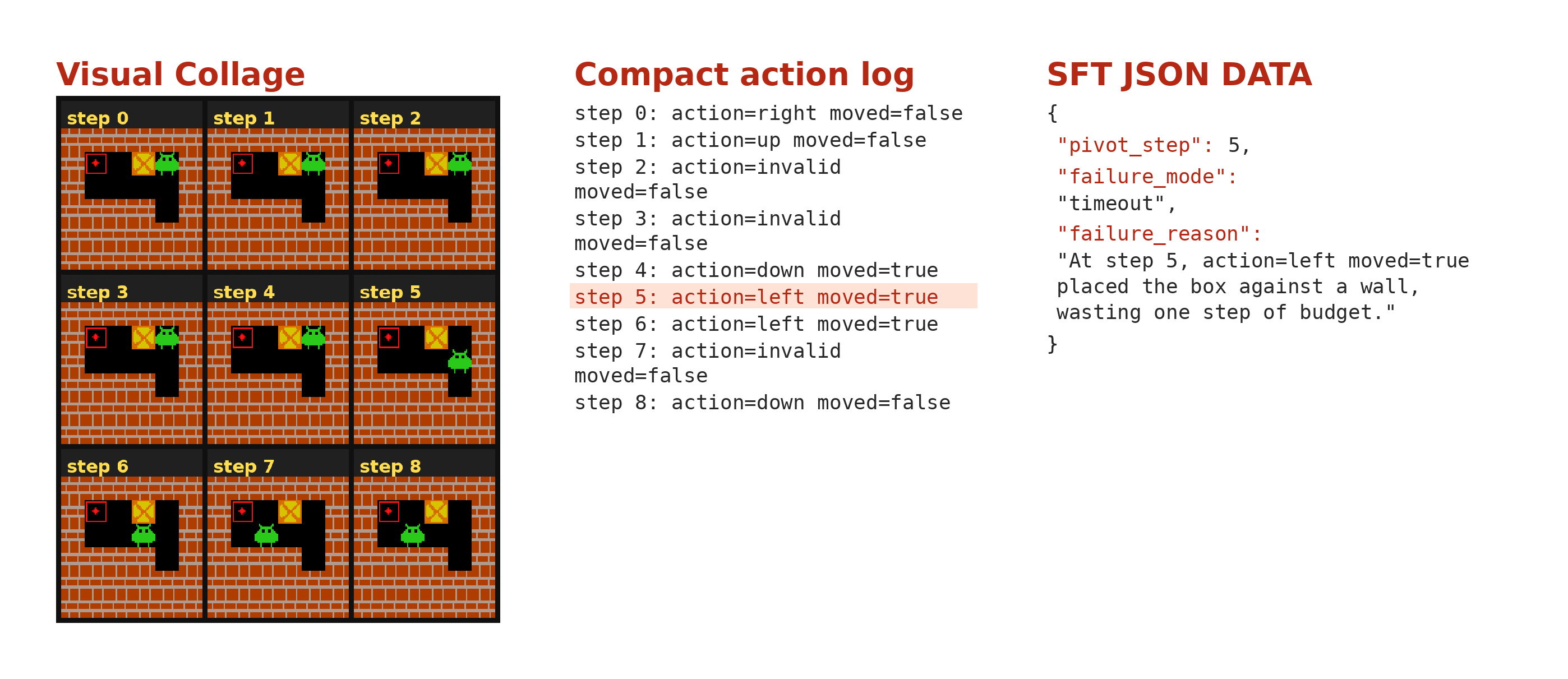}
\smallskip
\includegraphics[width=0.82\textwidth]{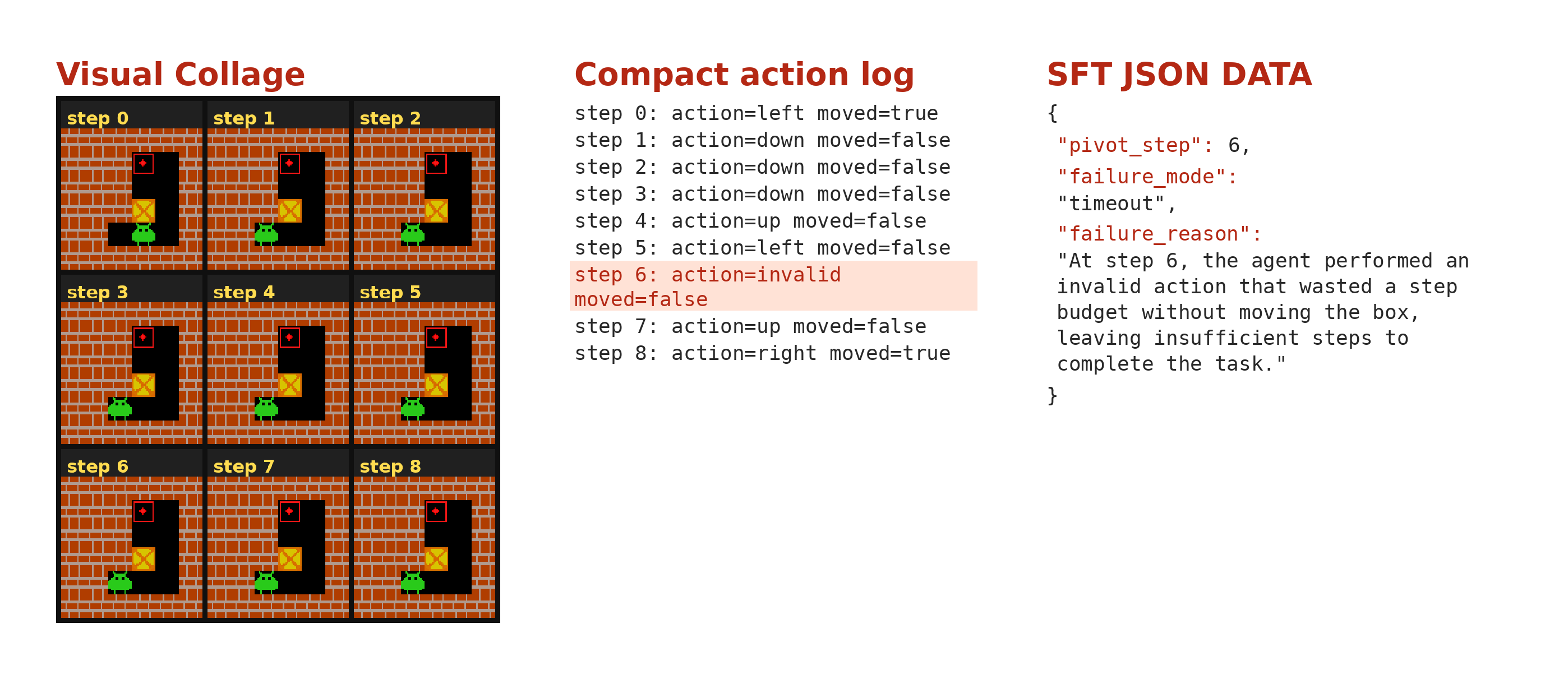}
\smallskip
\includegraphics[width=0.82\textwidth]{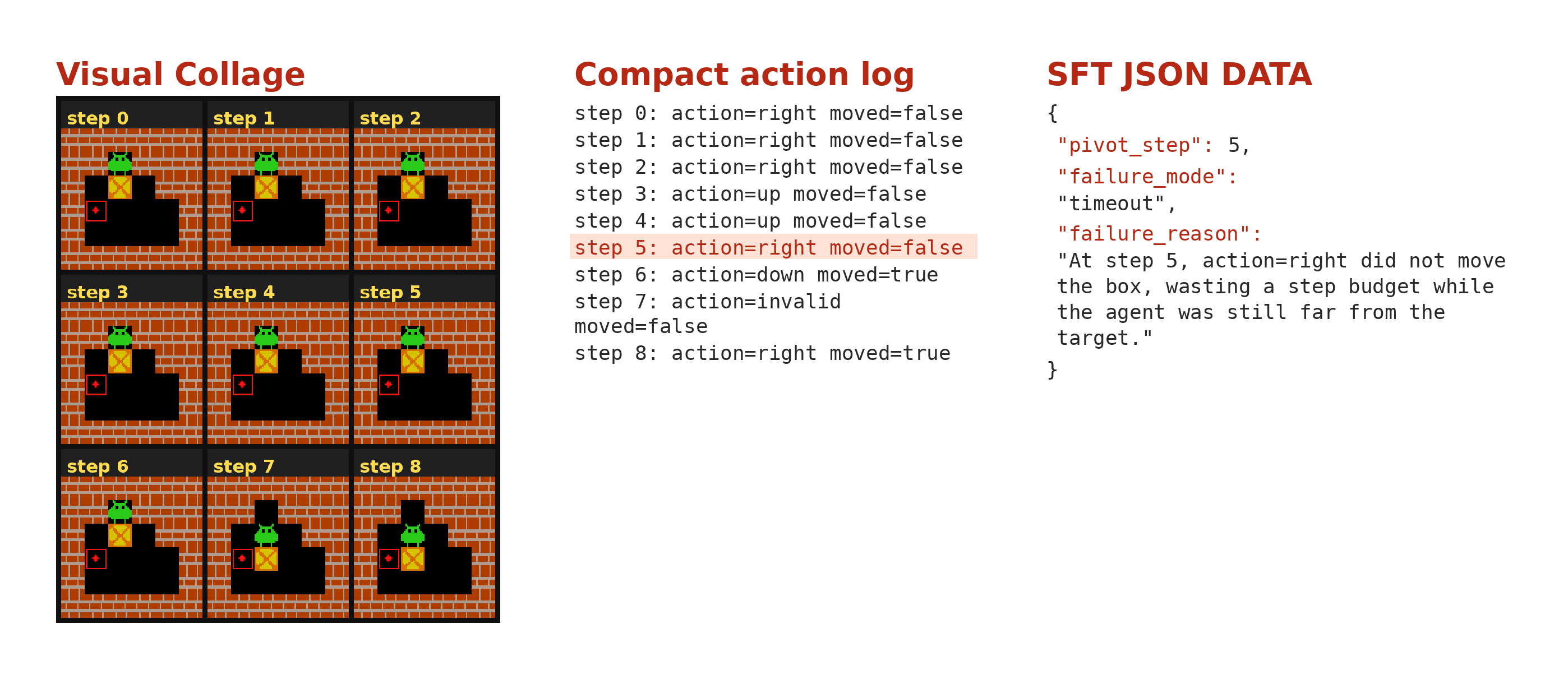}
\smallskip
\includegraphics[width=0.82\textwidth]{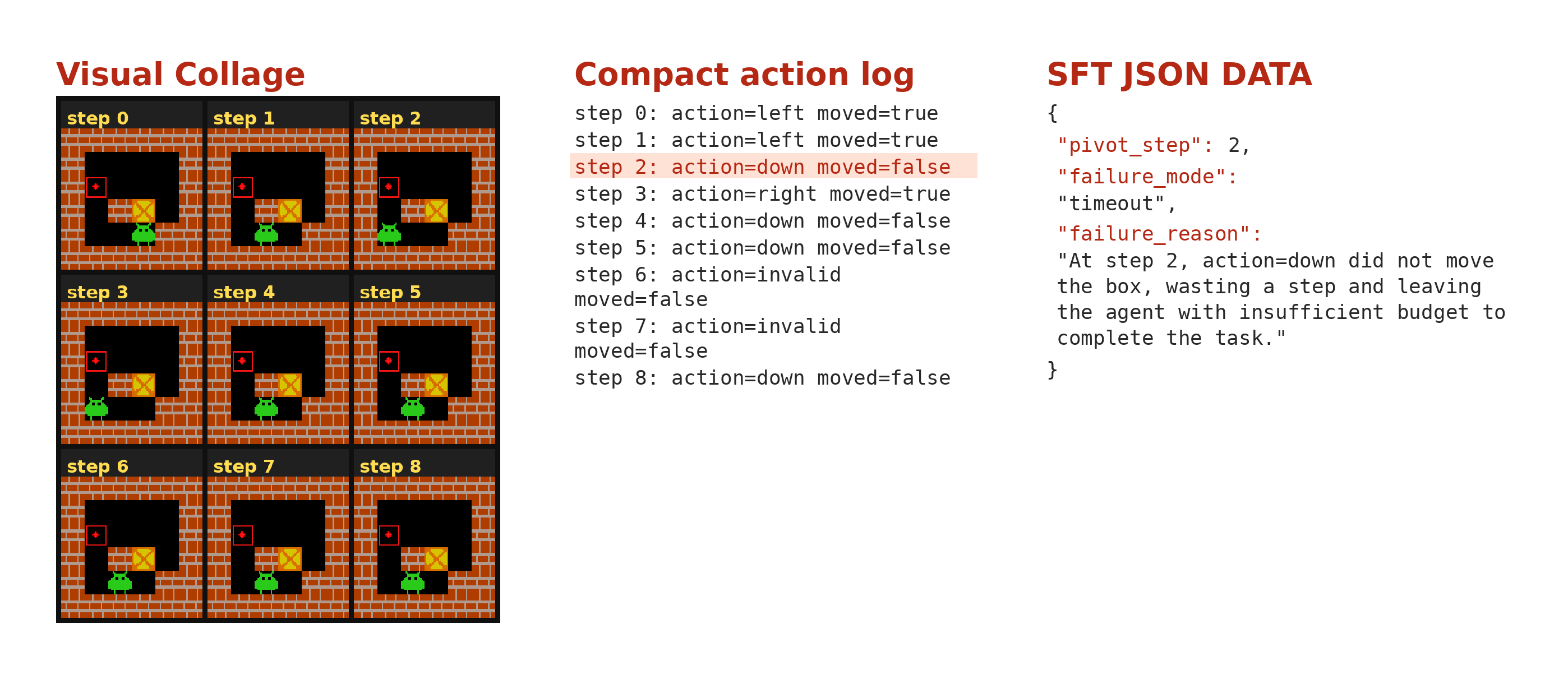}
\caption{Sokoban. SFT-stage input--output examples for this environment. Columns~1--2 are the visual trajectory collage and compact action log; column~3 is the JSON supervision target.}
\label{fig:app-col-sok}
\end{figure}

\begin{figure}[p]
\centering
\includegraphics[width=0.82\textwidth]{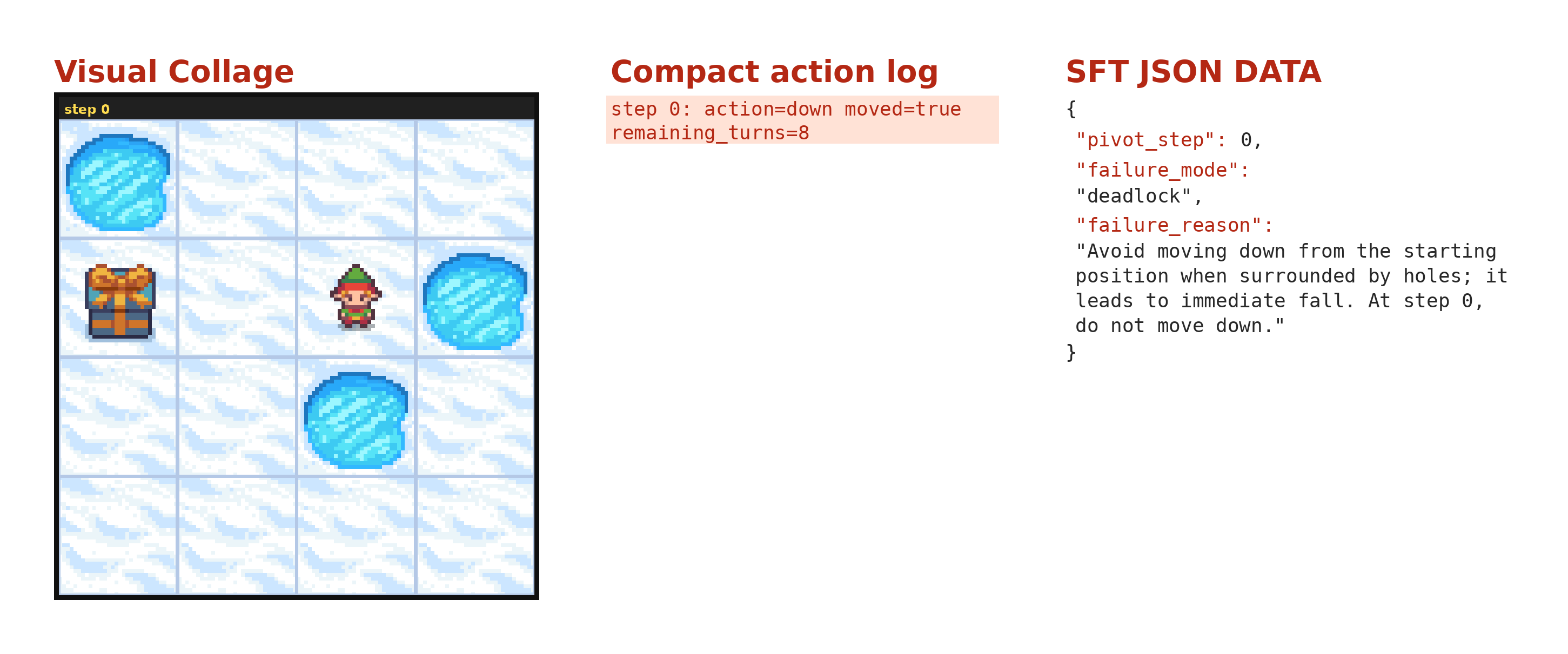}
\smallskip
\includegraphics[width=0.82\textwidth]{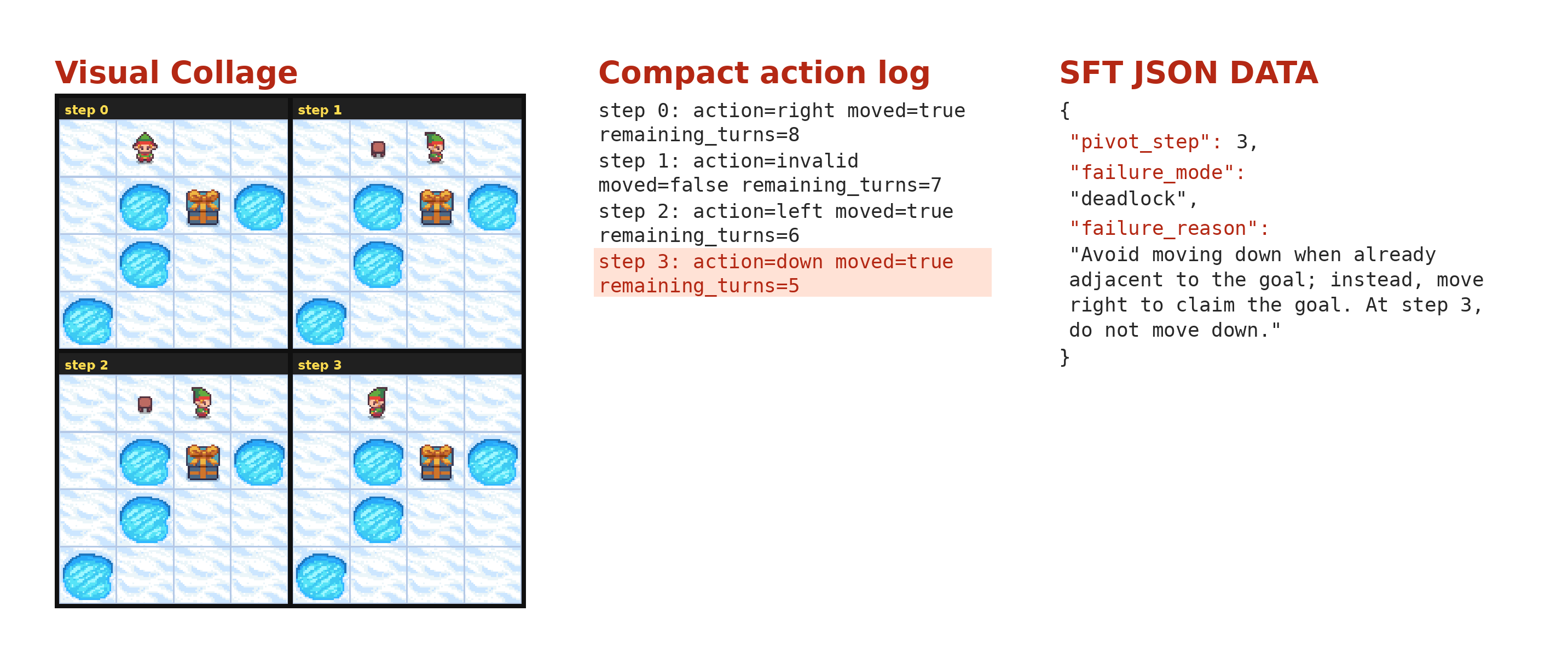}
\smallskip
\includegraphics[width=0.82\textwidth]{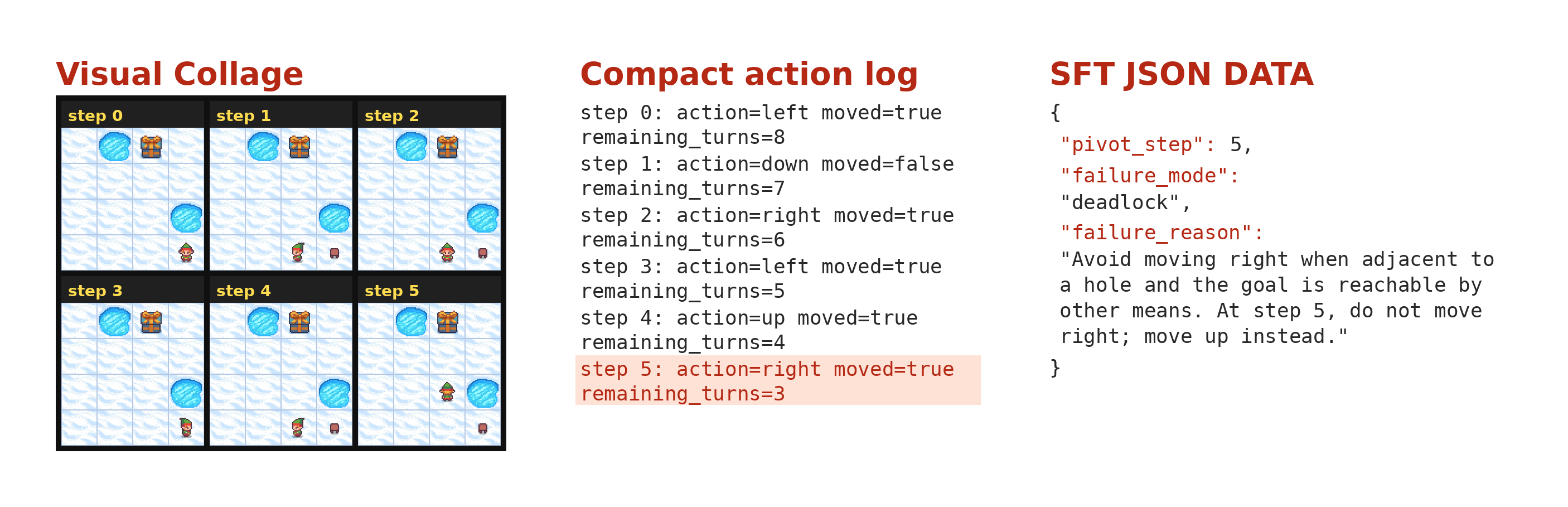}
\smallskip
\includegraphics[width=0.82\textwidth]{pivot_self_opd/frozenlake/collage_09.png}
\caption{FrozenLake. SFT-stage input--output examples for this environment. Columns~1--2 are the visual trajectory collage and compact action log; column~3 is the JSON supervision target.}
\end{figure}

\begin{figure}[p]
\centering
\includegraphics[width=\textwidth]{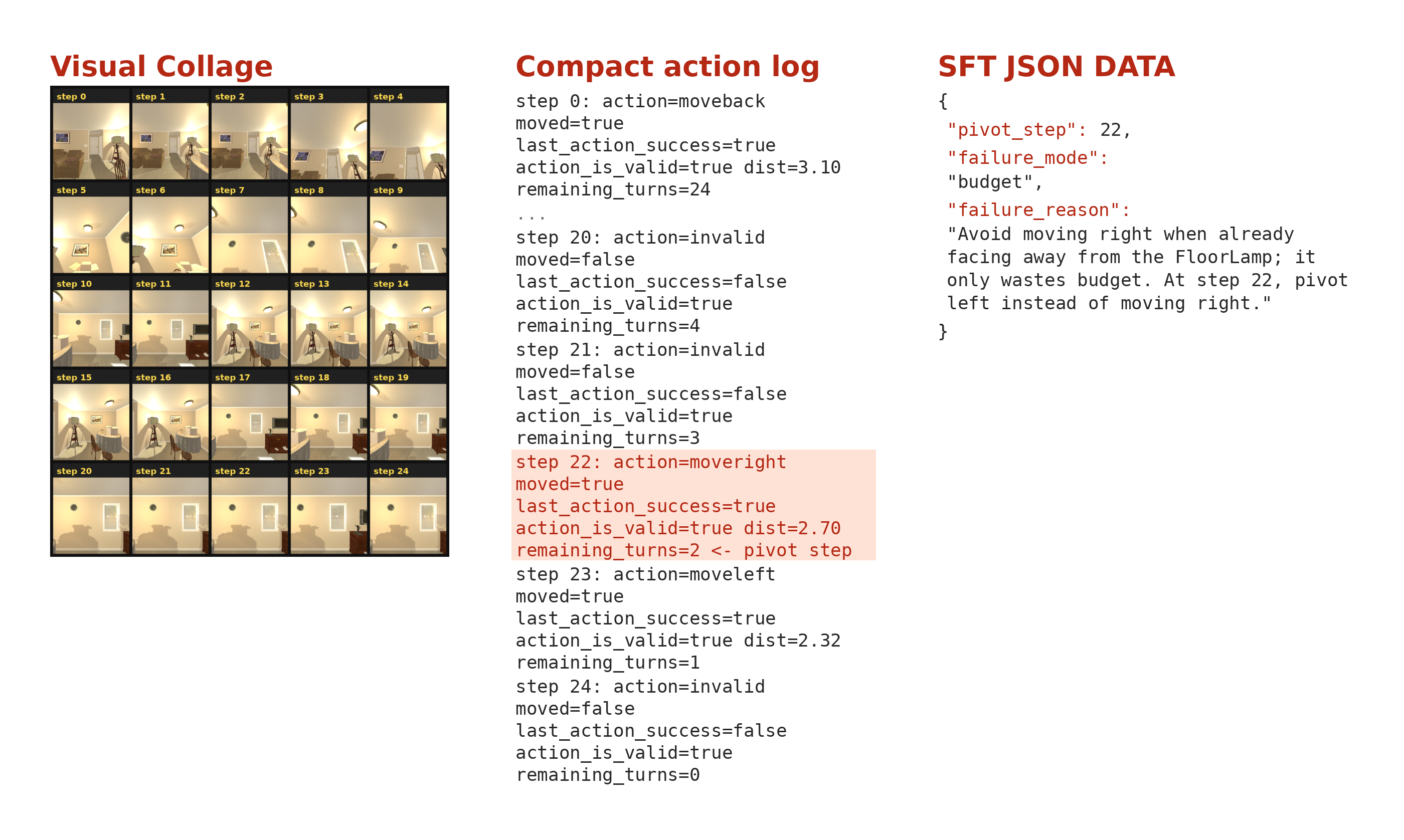}
\medskip
\includegraphics[width=\textwidth]{pivot_self_opd/navigation/collage_03.png}
\end{figure}

\begin{figure}[p]
\addtocounter{figure}{-1}
\ContinuedFloat
\centering
\includegraphics[width=\textwidth]{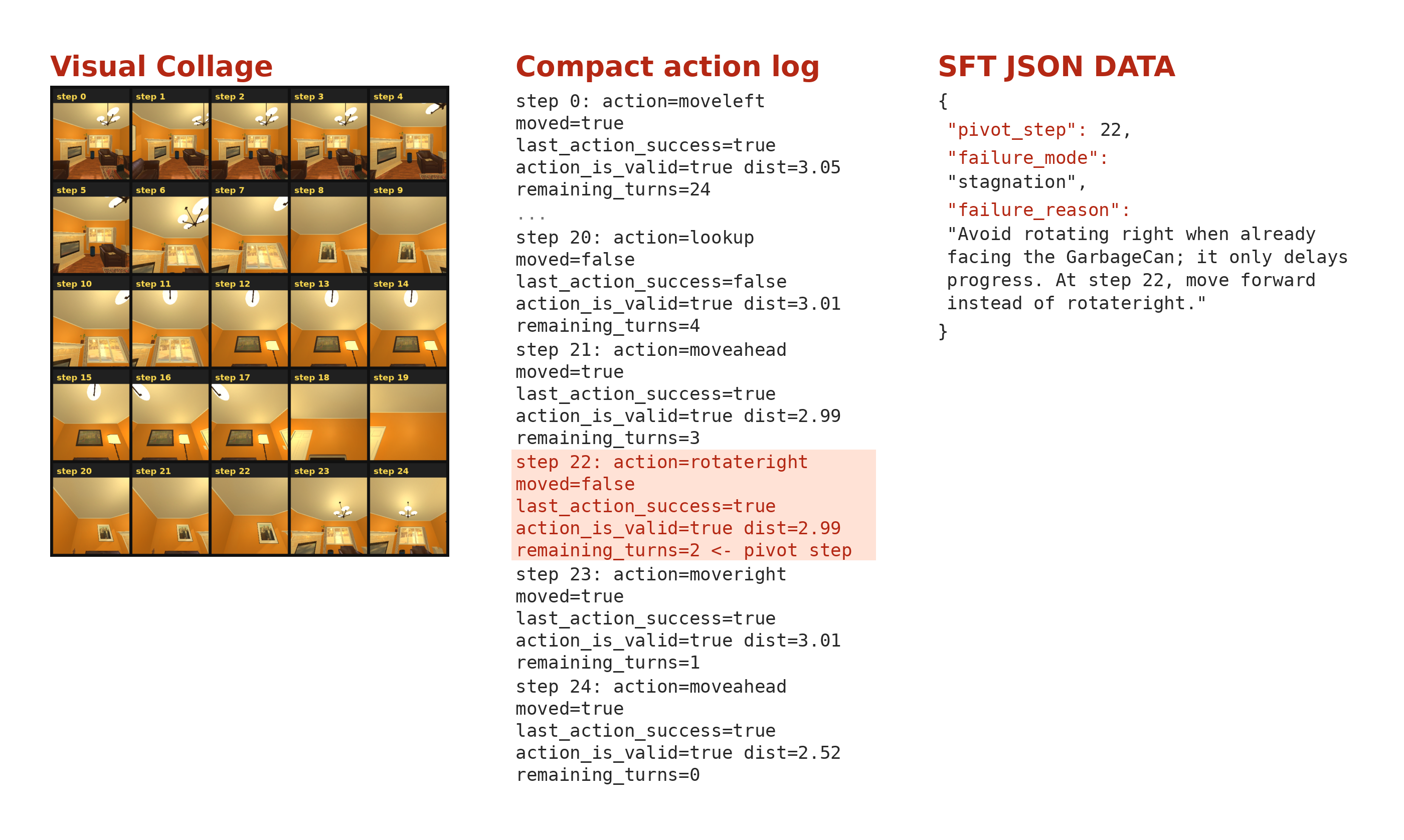}
\medskip
\includegraphics[width=\textwidth]{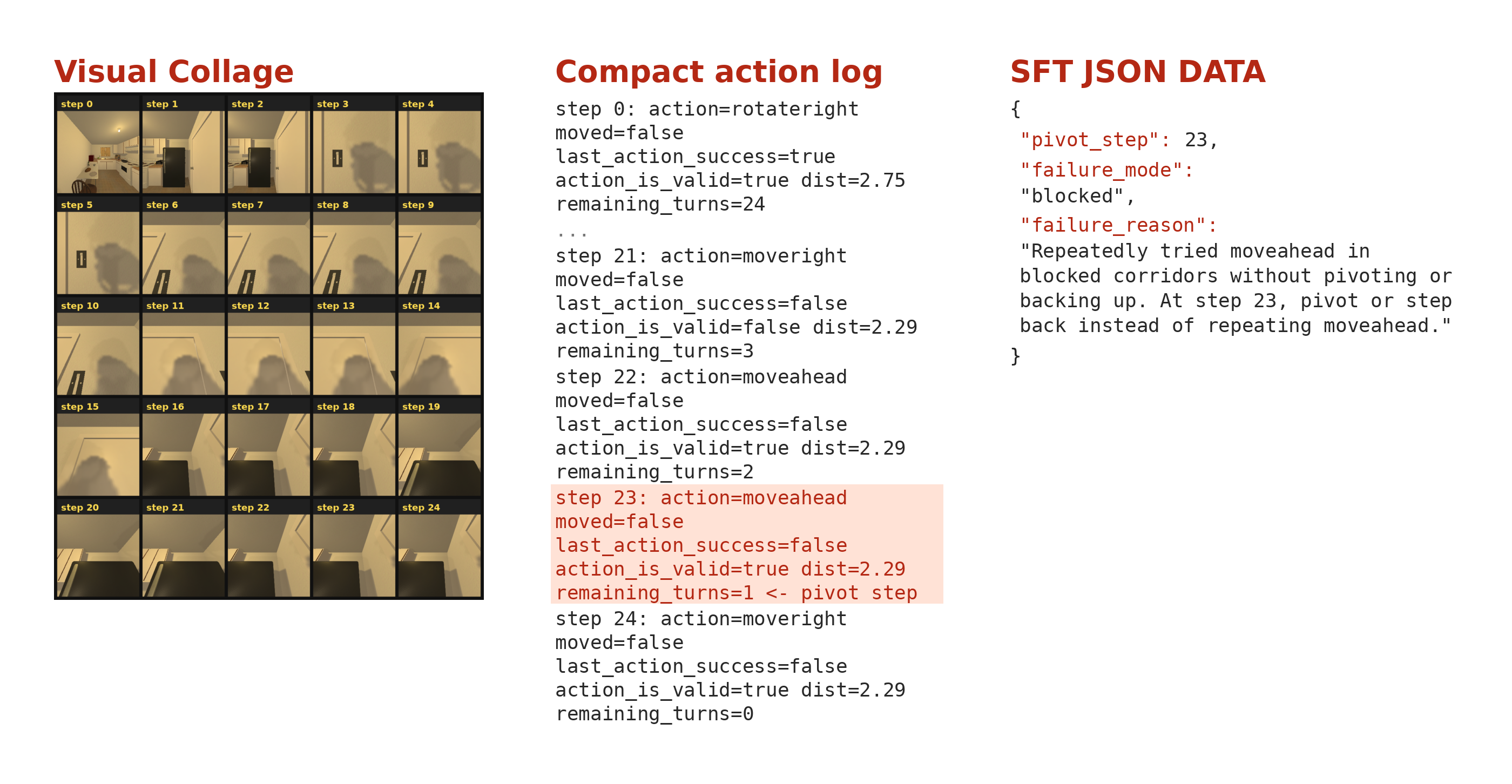}
\caption{Navigation. SFT-stage input--output examples for this environment. Columns~1--2 are the visual trajectory collage and compact action log; column~3 is the JSON supervision target.}
\end{figure}

\begin{figure}[p]
\centering
\includegraphics[width=\textwidth]{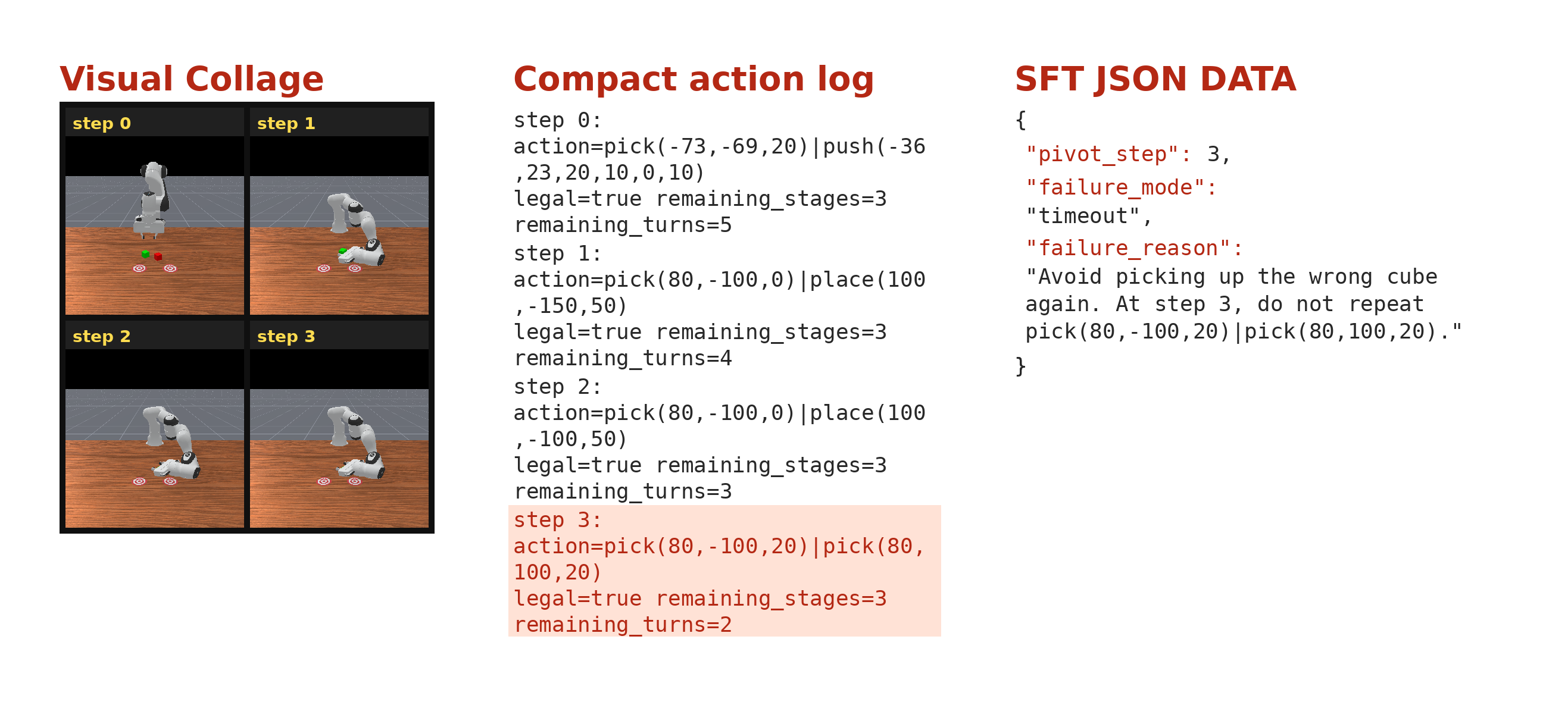}
\smallskip
\includegraphics[width=\textwidth]{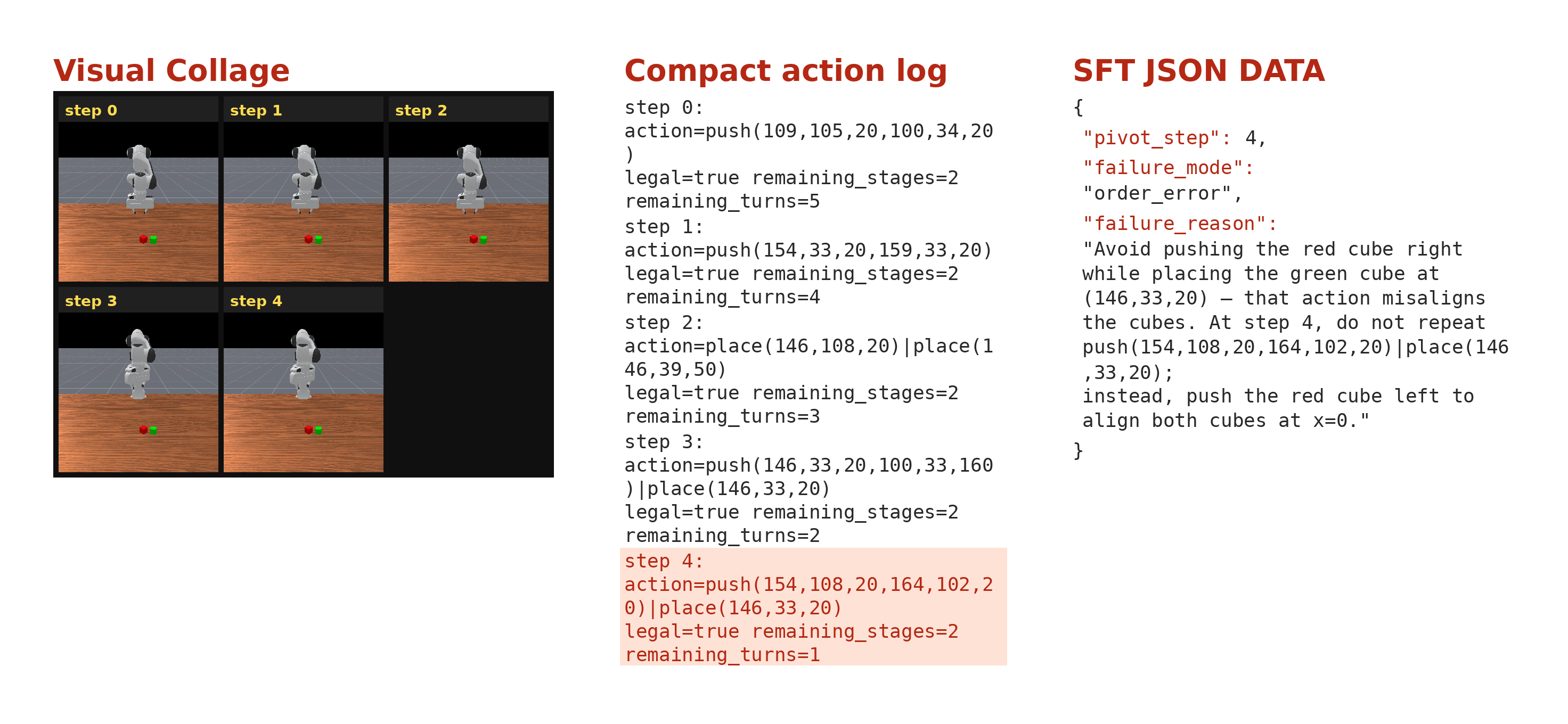}
\smallskip
\includegraphics[width=\textwidth]{pivot_self_opd/primitiveskill/collage_09.png}
\caption{PrimitiveSkill. SFT-stage input--output examples for this environment. Columns~1--2 are the visual trajectory collage and compact action log; column~3 is the JSON supervision target.}
\end{figure}

\clearpage
\section{SDAR Reproduction}
\label{app:sdar}

We reproduce SDAR~\citep{lu2026sdar} on the same five VAGEN tasks, rewards, and validation budget as \ourmodel{}.
Unlike \ourmodel{}, SDAR skips Stage~I Analyzer SFT and never localizes a pivot step: a frozen larger VLM writes one episode-level visual skill and supplies gated reverse-KL targets on all valid action tokens.
The student remains \texttt{Qwen2.5-VL-3B-Instruct} initialized from the raw Instruct checkpoint; because the Qwen2.5 family has no 8B variant, the teacher is \texttt{Qwen2.5-VL-7B-Instruct}.
Tab.~\ref{tab:main} reports scores; the following records roles and implementation flags for audit.

\appblock{Roles.}
Only the 3B student $\pi_\theta$ is trained; a frozen 7B teacher $\pi_\phi$ is queried through an OpenAI-compatible vision API (\texttt{analysis\_backend=openai}).
There is no Analyzer head, failure-mode vocabulary, or pivot panel.
On each on-policy trajectory, $\pi_\phi$ sees every RGB frame as its own image (\texttt{use\_visual\_collage=False}) together with the action transcript, generates a single free-form episode skill $r_\phi$ (\texttt{skill\_mode=episode\_only}, prompt \texttt{seed\_visual}), and re-scores the student's sampled tokens under $\tilde h_t^{\phi}=(h_t,r_\phi)$.
The 7B model is dropped at inference, leaving the same zero-overhead student rollout as \ourmodel{}.

\appblock{Objective.}
Let $\ell_t^{\mathrm{stu}}=\log\pi_\theta(a_t\mid h_t)$ and $\ell_t^{\phi}=\log\pi_\phi(a_t\mid \tilde h_t^{\phi})$, with gated reverse KL
\begin{equation}
\Delta_t^{\phi}
=
\operatorname{sg}\!\bigl[\ell_t^{\phi}-\ell_t^{\mathrm{stu}}\bigr],
\qquad
g_t^{\phi}
=
\sigma\!\bigl(\beta_{\mathrm{sdar}}\Delta_t^{\phi}\bigr),
\label{eq:sdar-gate}
\end{equation}
\begin{equation}
\mathcal{L}_{\mathrm{SDAR}}(\theta)
=
\mathbb{E}\Bigl[
m_{\mathrm{act}}\,
g_t^{\phi}\,
\bigl(
\operatorname{sg}[\ell_t^{\phi}]
-
\ell_t^{\mathrm{stu}}
\bigr)
\Bigr].
\label{eq:sdar-loss}
\end{equation}
The joint loss is the same GRPO-plus-distillation form as Eq.~\eqref{eq:joint-objective},
\begin{equation}
\mathcal{L}(\theta)
=
\mathcal{L}_{\mathrm{GRPO}}(\theta)
+
\lambda_{\mathrm{sdar}}
\mathcal{L}_{\mathrm{SDAR}}(\theta),
\label{eq:sdar-joint}
\end{equation}
with $\phi$ under stop-gradient.
Unlike \ourmodel{}, Eq.~\eqref{eq:sdar-loss} is applied to all valid action tokens rather than failed trajectories only, and the teacher context is a global episode skill rather than a localized visual panel $P_{\hat t}$.

\appblock{Protocol.}
We match the VAGEN data, horizons, sparse rewards, GRPO grouping, and validation budget of Sec.~\ref{sec:setup} and App.~\ref{app:envs}: Sokoban (room $[6,6]$, 1 box, solution length $[1,5]$, $T_{\max}{=}9$), FrozenLake $4{\times}4$ with no slip ($T_{\max}{=}9$), Navigation ($T_{\max}{=}25$), PrimitiveSkill ($T_{\max}{=}6$), and SVG ($T_{\max}{=}2$).
SDAR runs only on Qwen2.5-VL-3B (no Qwen3-VL-2B row in Tab.~\ref{tab:main}).

\begin{table}[t]
\centering
\small
\setlength{\tabcolsep}{4pt}
\caption{SDAR settings (3B student, frozen 7B teacher).}
\label{tab:sdar-hyper}
\begin{tabular}{@{}>{\raggedright\arraybackslash}p{0.38\linewidth}@{}>{\raggedright\arraybackslash}p{0.56\linewidth}@{}}
\toprule
Item & Value \\
\midrule
Student $\pi_\theta$ / teacher $\pi_\phi$ & \texttt{Qwen2.5-VL-3B} / \texttt{7B-Instruct} (frozen) \\
Skill & \texttt{seed\_visual}, \texttt{episode\_only}; all frames, no collage \\
Distillation & all valid tokens; $\lambda_{\mathrm{sdar}}{=}0.01$, $\beta_{\mathrm{sdar}}{=}5$ \\
RL (shared with \ourmodel{}) & $N{=}8$, LR $1{\times}10^{-6}$, KL $0.01$, invalid pen.\ $0.1$ \\
Training & batch $16$/val $128$; $250$ updates (Sok.\ $350$); history $2$ \\
Pivot Analyzer / $P_{\hat t}$ & none \\
\bottomrule
\end{tabular}
\end{table}

\clearpage
\section{Training Curves}
\label{app:curves}
\enlargethispage{1.5\baselineskip}

\begin{center}
\parbox{0.98\linewidth}{\raggedright\noindent
Figs.~\ref{fig:ablation} and~\ref{fig:ablation-2b} show Stage~II validation learning curves for \armGRPO{}, \armM{}, \armMP{}, and \armMPR{} on Qwen2.5-VL-3B and Qwen3-VL-2B; dots mark running-best validation points aligned with Tab.~\ref{tab:main}.}

\nobreak
\includegraphics[width=\linewidth,height=0.285\textheight,keepaspectratio]{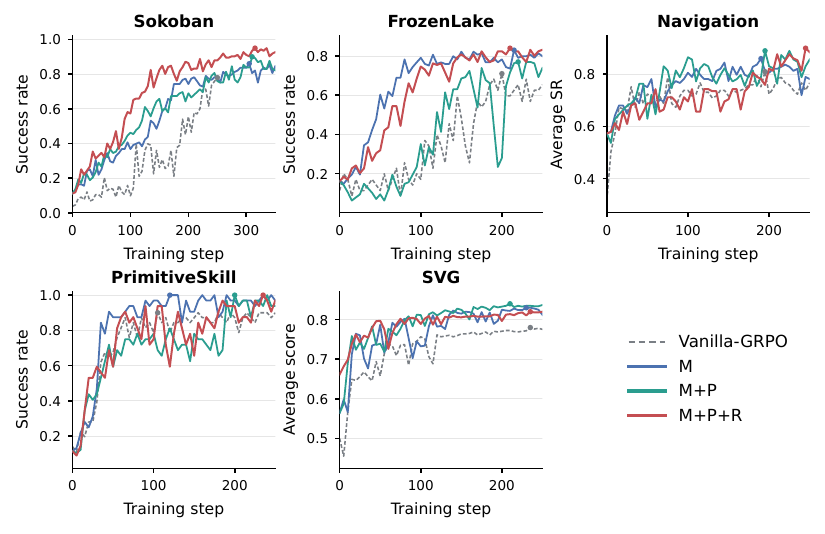}
\captionof{figure}{Training curves for \armGRPO{}, \armM{}, \armMP{}, and \armMPR{} on Qwen2.5-VL-3B.
Dots mark the running-best validation point.}
\label{fig:ablation}

\includegraphics[width=\linewidth,height=0.285\textheight,keepaspectratio]{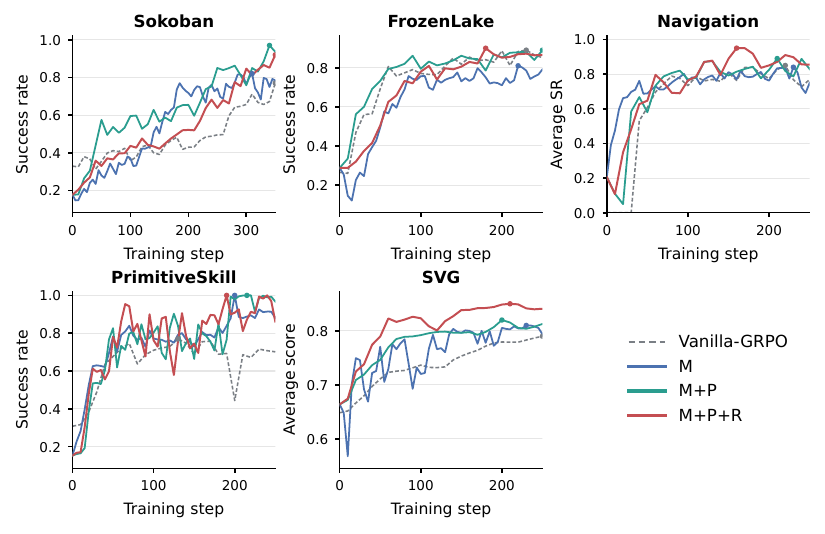}
\captionof{figure}{Training curves for \armGRPO{}, \armM{}, \armMP{}, and \armMPR{} on Qwen3-VL-2B.
Dots mark the running-best point.}
\label{fig:ablation-2b}
\end{center}

\end{document}